\documentclass[12pt]{article}
\usepackage[margin=1in]{geometry}
\usepackage{amsmath, amsfonts, amssymb, amsthm}
\usepackage[utf8]{inputenc}
\usepackage[english]{babel}
\usepackage{hyperref}
\usepackage{graphicx}
\usepackage{lipsum}
\usepackage[round, authoryear, longnamesfirst]{natbib}
\usepackage{blkarray}
\usepackage{float}
\usepackage{pifont}
\usepackage{subcaption}

\begin{document}

\title{Rank Pruning for Dominance Queries in CP-Nets}
\date{}
\maketitle
\vspace{-1cm}
\noindent\begin{tabular}{l c r}
\emph{Authors:} & & \\
\textbf{Kathryn Laing} & \hspace{6cm} & K.Laing@leeds.ac.uk\\
\emph{University of Leeds} & & \\
\textbf{Peter Adam Thwaites} & & P.A.Thwaites@leeds.ac.uk\\
\emph{University of Leeds} & & \\
\textbf{John Paul Gosling} & & J.P.Gosling@leeds.ac.uk\\
\emph{University of Leeds} & & \\
\end{tabular}

\section*{Abstract}

Conditional preference networks (CP-nets) are a graphical representation of a person’s (conditional) preferences over a set of discrete variables. In this paper, we introduce a novel method of quantifying preference for any given outcome based on a CP-net representation of a user’s preferences. We demonstrate that these values are useful for reasoning about user preferences. In particular, they allow us to order (any subset of) the possible outcomes in accordance with the user’s preferences. Further, these values can be used to improve the efficiency of outcome dominance testing. That is, given a pair of outcomes, we can determine which the user prefers more efficiently. Through experimental results, we show that this method is more effective than existing techniques for improving dominance testing efficiency. We show that the above results also hold for CP-nets that express indifference between variable values. \\

\noindent \textbf{Keywords:} CP-Nets, Preference Reasoning, Dominance Testing

\section{Introduction}

Conditional preference networks (CP-nets) as described by \citet{boutJAIR2004} are structures for modelling a person's conditional preferences over a set of discrete variables. Representing and reasoning with a person's preferences is an area of interest in AI with applications in automated decision making \citep{nuneEAAI2015}, recommender systems \citep{ricc2011}, and product configuration \citep{alanIEA-AIE2014}. CP-nets represent preferences in a compact manner that is easily interpreted. Further, they are based upon \emph{ceteris paribus} (all else being equal) preference statements, which are easy to elicit from a non-expert user or client. For example, if a user was being asked about seat preferences on a flight, they might say that they prefer to sit in business class rather than economy, given it is a long-haul flight.  It is implicit that the user is assuming everything else about the seat is the same when making this statement.\\

Using the CP-net model to represent a user's preference, we introduce a novel way of quantifying the user's preference for a single outcome, where an outcome is an assignment of values to all variables of interest. In our example, an outcome might be the seat specification $\langle$\emph{economy class}, \emph{long-haul flight}, \emph{window seat}$\rangle$. This quantification of user preference over outcomes makes it easier to reason about the user's preferences about the outcomes. Many questions of interest in this setting are naturally about preference over the outcomes, in particular, outcome optimization, consistent orderings, and dominance queries \citep{boutJAIR2004, brafCI2004, boutCI2004, goldJAIR2008,sant2016}. Outcome optimization asks which outcome is optimal, possibly given a partial assignment to the variables, which would be of interest when doing product configuration or automated decision making. Consistent orderings are orderings of (some subset of) the outcomes that obey everything we know about the user's preferences. That is, if one outcome ($o$) is known to be preferred to another ($o'$), then $o$ should appear in the ordering before $o'$. A natural application of this is in recommender systems, in particular e-commerce, so that items can be displayed such that those of most interest to the customer appear first. Finally, a dominance query asks, given two outcomes, which is preferred by the user. Being able to answer such queries is critical to automated decision making. \\

Outcome optimisation has been dealt with by \citet{boutJAIR2004}, who provide a method of obtaining the optimal outcome (possibly given a partial variable specification) in linear time in the number of variables. We demonstrate how our quantification of user preference over the outcomes (which are called outcome ranks) can be used to obtain a consistent ordering of (any subset of) the outcomes. \citet{boutJAIR2004} also detail how a consistent ordering can be obtained for all outcomes or any subset. However, we demonstrate that for larger subsets of the outcomes our method is more efficient. The size of the outcome set is at least $2^n$ (where $n$ is the number of variables). Thus, subsets of the outcomes can get very large even for relatively small CP-nets. Furthermore, our method of obtaining consistent orderings can be applied, with the same complexity, in the case of CP-nets with indifference statements; whereas the complexity of the method by \citet{boutJAIR2004} for consistently ordering subsets is unknown in this case, though they conjecture that it is hard.\\

Despite being a natural question, dominance queries are NP-hard problems (even when restricting to binary variables and an acyclic preference structure); see papers by \citet{boutJAIR2004} and \citet{goldJAIR2008} for further results on the complexity of answering dominance queries. \citet{santAAAI2010} introduce a novel approach to answering dominance queries by using model checking. However, their experimental results all utilise binary CP-nets, thus, it is unclear how well this method performs when there are multivalued variables. \citet{sunESWA2017} introduce a different approach; they successively compose the preferences of all variables (in topological order) to form a single preference table. From this table, consistent orderings can be obtained and dominance queries can be answered. However, they also consider only binary CP-nets and so how well these methods handle multivalued CP-nets is unknown. The more standard way of answering dominance queries is to attempt to construct an improving flipping sequence between the two outcomes of interest \citep{boutJAIR2004}, we explain this notion in more detail later on. If the dominance query asks `Is $o$ preferred to $o'$?', this can be visualised as building up a search tree from the root node $o'$, that either eventually reaches $o$ (and so the dominance query is true) or eventually can not expand any further (and so the dominance query is false). There have been several attempts to improve the efficiency of this method by introducing procedures for pruning the branches of this search tree as one constructs it \citep{boutJAIR2004, liAAMS2011}. We show how our outcome ranks can be used to prune this search tree in a different way and thus improve the efficiency of answering dominance queries. Our pruning technique can be combined with any of the existing pruning methods to further improve efficiency. We give an experimental comparison of the performance of our rank pruning with the existing pruning methods that preserve search completeness. These experiments also evaluate the performance of all the possible combinations of the different methods to determine the optimal method for answering dominance queries. The results find that rank pruning is more effective than the existing methods and a valuable addition when considering combinations of methods. \citet{alleJAIR2017} propose improving the efficiency of dominance testing by imposing a bound on the search depth for improving flipping sequences. This bound could be applied to our dominance testing procedure (\S5) in the same way \citet{alleJAIR2017} apply it to the dominance testing procedure given by \citet{liAAMS2011}. However, the depth bound proposed by \citet{alleJAIR2017} has only been experimentally shown to preserve completeness of dominance testing for relatively small binary CP-nets. This has not been proven for binary CP-nets in general and for CP-nets with multivalued variables utilising this bound does not preserve completeness in general.\\

There have been several previous attempts at quantifying a user's preference over outcomes, given a CP-net representation of the user's preferences. \citet{domsIJCAI2003} provide two methods of approximating user preferences in order to obtain an ordering of the outcomes that is consistent with all known preference information. The first method is to construct a penalty function over the outcomes and then order them according to this function. The second method associates each outcome with an $n$-tuple, obtained by evaluating the level of preference for each variable assignment individually. A consistent ordering is then obtained by ordering these vectors lexicographically. However, \citet{domsIJCAI2003} do not discuss how dominance queries about the CP-net might be answered using these approximations. \citet{liAAMS2011} introduce a penalty function over the outcomes very similar to that introduced by \citet{domsIJCAI2003}. They go on to show how these penalty values can be used to prune the search tree of dominance queries analogously to how we use our outcome ranks here. Further, \citet{liTechRep} extend this penalty function so that it is defined for TCP-nets (CP-nets with additional (relative) importance statements) and they claim that it is straightforward to extend their pruning method to dominance testing for TCP-nets, though this is not shown explicitly. We instead illustrate how our definition (and pruning methods) can be generalised to allow the user to express indifference, which is not permitted in Li et al.'s penalty definitions. \citet{boutUAI2001} also look at quantifying preference by introducing an extension of CP-nets that has conditional utilities rather than conditional preferences, these structures are called UCP-nets. One of the aims of combining utilities with the CP-net here is in order to obtain a global utility over the outcomes, so that answering dominance queries would become a simple task of comparing utilities. However, given a UCP-net elicited from a user with, naturally, normalised utilities, their paper focuses on how one can narrow down the global utility possibilities with the aim of choosing an optimal decision rather than with the aim of obtaining a single global utility from which dominance queries can be answered. \citet{mcgeAAAI2002} present a method for obtaining a global utility over the outcomes, given any set of (consistent) \emph{ceteris paribus} preference rules. This utility function obeys the given preference rules but beyond this it cannot claim to be an accurate quantification of user preference.\\

The remainder of this paper is structured as follows. First, \S2 contains all of the required background about CP-nets and their event tree representations. In \S3, we introduce our outcome ranks and show how they can be used to obtain a consistent ordering of (any subset of) the outcomes. Further, we show that this method needs no adaptation for CP-nets with additional plausibility constraints. In \S4, we describe algorithms for calculating the rank of any given outcome. In \S5, we show how our outcome ranks can be used to improve the efficiency of answering dominance queries by pruning the search tree. In \S6, we present our experimental comparison of rank pruning with existing methods of pruning the search tree (and all possible combinations of methods) and analyse the results to determine the best method for answering dominance queries. In \S7, we show how our rank definition can be generalised to allow the user to express indifference. Further, we show that all of our previous results hold for this generalised rank definition and so apply also to CP-nets that have indifference. Finally, \S8 provides a discussion of these results and plans for future work.

\section{Preliminaries}

In this section, we introduce the basics of conditional preference networks (CP-nets) as defined by \citet{boutJAIR2004}. We also show how a CP-net can be represented by an event tree \citep{edwaISR1983}. This alternate representation is important for the construction of our outcomes ranks (\S3.1).

\subsection{CP-Nets}

\textbf{Definition 1: Conditional Preference Network (CP-Net)} \citep{boutJAIR2004}. A \emph{CP-net} $N$, over variables $V$, is a directed graph $G$, with nodes $V$. Each node $X$, is annotated with a \emph{conditional preference table} CPT$(X)$. For any $X\in V$,  $\text{CPT}(X)$ gives, for each possible assignment of values to Pa$(X)$ (the parent variables of $X$ in $G$), the user's order of \emph{ceteris paribus} preference over all possible values $X$ can take (that is, the domain of $X$,~Dom$(X)$). We assume all variables to have discrete domains.\\

The graph $G$ is a preferential dependency graph with the following underlying assumption. For any $X, Y\in V$, $X$ is \emph{preferentially independent} of $Y$ given Pa$(X)$. That is, once~Pa$(X)$ have been assigned values, the user's preference over Dom$(X)$ is fixed and not affected by the value taken by $Y$. To illustrate these ideas, consider the following example.\\

\textbf{Example 1.} Suppose again that we are modelling a user's preference over aeroplane seats. The variables we might take into account, and their respective domains, are as follows.

\begin{tabular}{l l}
$A = \text{Flight Length } \hspace*{0.5cm}$ & $\text{Dom}(A)=\{a:\text{short, } \bar{a}:\text{long-haul}\}$\\
$B = \text{School Term Time } \hspace*{0.5cm}$ & $\text{Dom}(B)=\{b:\text{term, } \bar{b}:\text{holiday}\}$\\
$C = \text{Class } \hspace*{0.5cm}$ & $\text{Dom}(C)=\{c:\text{economy, } \bar{c}:\text{business, } \bar{\bar{c}}:\text{first}\}$\\
$D = \text{Pay Extra for Wi-Fi } \hspace*{0.5cm}$ & $\text{Dom}(D)=\{d:\text{no, } \bar{d}:\text{yes}\}$\\
\end{tabular}\\

One example of a CP-net over these variables is given in Figure 1. The structure of this CP-net shows that the user has a strict preference for short flights over long-haul flights (\emph{ceteris paribus}, that is, given $B,C,D$ take the same values) and for flying in term time over flying in holiday time (\emph{ceteris paribus}, that is, given $A,C,D$ take the same values). These preferences are unaffected by the values taken by any other variable. However, the user's preference for which class they fly in is dependent (conditional) upon the values taken by~$A$ and $B$ (Flight Length and School Term Time). If it is a short flight in term time, then the user prefers economy to business to first class (\emph{ceteris paribus}- given that $D$ takes the same value). However, if it is a short flight in holiday time, then the user prefers business to first to economy class. Once the values of $A$ and $B$ are determined, these preferences over $C$ (Class) are fixed and do not change (regardless of the value taken by $D$), by our preferential independence assumption above. Similarly, the user's preference over $D$ (Pay Extra for Wi-Fi) depends on the value taken by $C$, but these preferences are independent of the values taken by $A$ and $B$.\\

\begin{figure}[h]
\begin{center}
\includegraphics[scale=0.5]{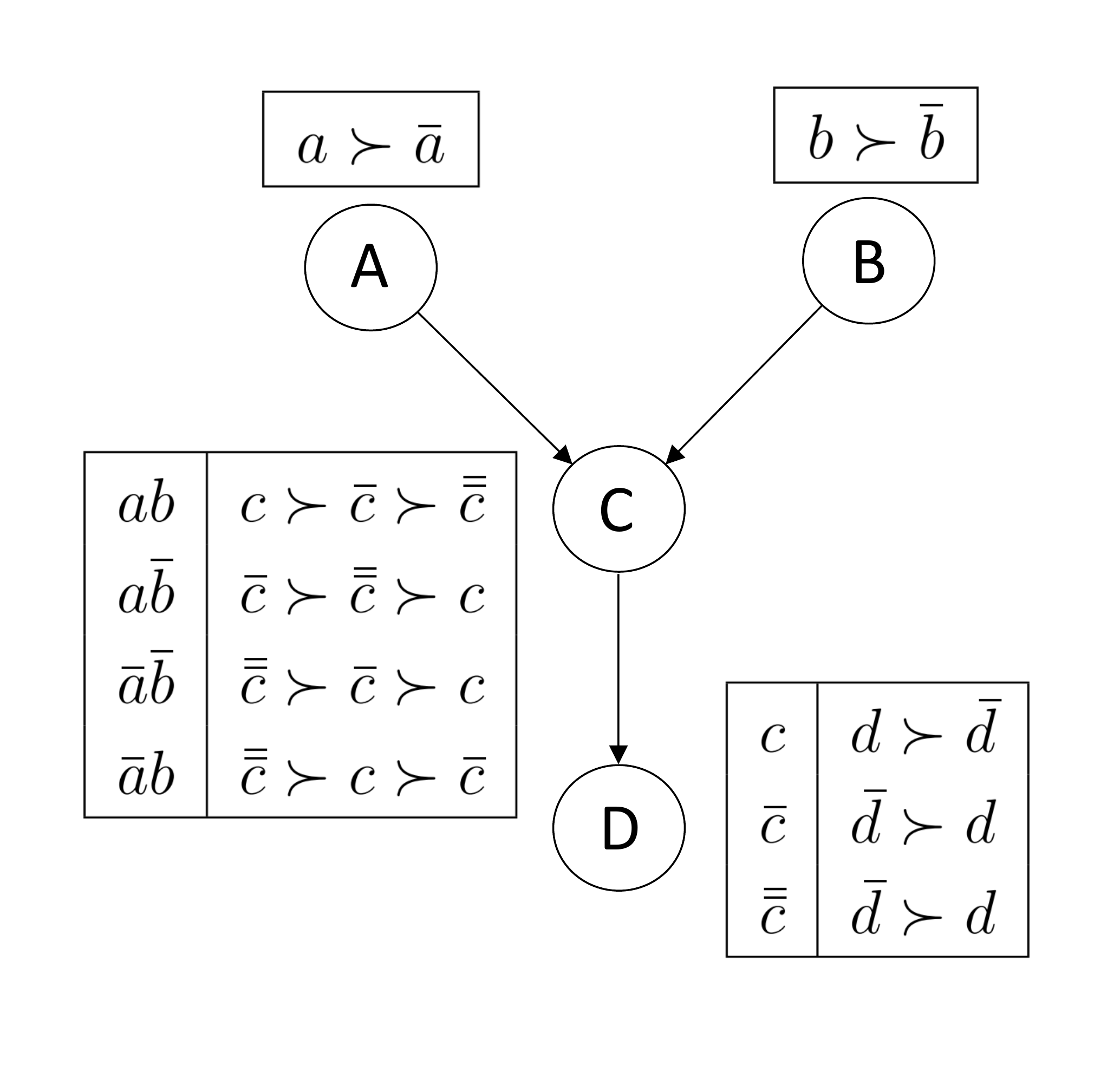}
\end{center}
\caption{CP-Net Example}
\end{figure}

We assume throughout this paper that we are dealing with CP-nets whose directed graph (structure) is acyclic. When we refer to binary CP-nets we mean CP-nets where all variables are binary. For the majority of this paper (except \S7), we assume that every row of the conditional preference tables (CPTs) contains a strict complete order of the appropriate domain. In \S7, we discuss how our results can be generalised to apply to cases where the CPTs may express indifference between values. This is an important extension because indifference is a natural notion that we may reasonably expect people to express when specifying their preferences.\\

Given a CP-net $N$, over variables $V=\{V_1,V_2,...,V_n\}$, an \emph{outcome} $o$, is an $n$-tuple representing an assignment of values to all variables, $o\in$ Dom$(V_1)\times$ Dom$(V_2)\times\cdots\times$ Dom$(V_n)$. Let $\Omega$ denote the set of all outcomes, then $|\Omega|= |\text{Dom}(V_1)|\times |\text{Dom}(V_2)|\times\cdots\times |\text{Dom}(V_n)|$. For our previous example, $a\bar{b}\bar{\bar{c}}\bar{d}$ is an example of an outcome (specifically this is a short flight in holiday time, sitting in first class with Wi-Fi). In total, there are 24 possible outcomes for the CP-net given in Example 1. In general, $|\Omega|\geq 2^n$ with equality only in the case of binary CP-nets.\\

The \emph{preference graph} associated with $N$ is a directed graph $G_N$, with the outcomes $\Omega$ as nodes and edges defined as follows. Let $o,o'\in\Omega$, then there is an edge $o\rightarrow o'$ if and only if $o$ and $o'$ differ on the value assigned to exactly one variable, say $X$, and, in the row of~$\text{CPT}(X)$ corresponding to the assignment of values to Pa$(X)$ in both $o$ and $o'$, the value of $X$ taken in~$o'$ is preferred to that in $o$. As the preference statements in the CPTs of a CP-net are all \emph{ceteris paribus}, they only encode preferences between outcome pairs that differ on exactly one variable. Thus, the edges in the preference graph (and their transitive closure) represent all known preference information about the user. That is, it is an equivalent representation to the CP-net itself. \citep{boutJAIR2004}\\

\textbf{Definition 2: Entailment.} Let $N$ be a CP-net with associated preference graph $G_N$, and two associated outcomes $o$ and $o'$. We say that $N$ \emph{entails} the relation `$o$ is preferred to~$o'$', denoted $N\vDash o\succ o'$, if and only if there is a directed path $o'\leadsto o$ in $G_N$.\\

The entailed relations are all of the user preferences between outcome pairs encoded in the CP-net. A \emph{consistent ordering} for a CP-net is a complete ordering of the outcomes~$\Omega$, that obeys all known (entailed) preference information about the user. Equivalently, a consistent ordering is any ordering of $\Omega$ such that if there is a path $o\leadsto o'$ in the preference graph, then~$o'$ comes before $o$ in the ordering (that is, any topological ordering of the preference graph). Notice that entailed relations hold in all consistent orderings. Further, as we are considering only acyclic CP-nets, there will always be at least one consistent ordering \citep{boutJAIR2004}.\\

\textbf{Remark 1.} The above definitions for entailment and consistent orderings are equivalent but not identical to those given by \citet{boutJAIR2004}. \citet{boutJAIR2004} define orderings that satisfy the CP-net (consistent orderings) to be complete orderings of the outcomes that obey all of the \emph{ceteris paribus} preference statements given in the CPTs. They go on to define entailment as $N\vDash o\succ o'$ if and only if $o$ comes before $o'$ in all consistent orderings. \citet{boutJAIR2004} show their definition of entailment to be equivalent to the above definition. For consistent orderings, it is clear that the two definitions are equivalent as the preference graph is an equivalent representation of the CP-net. Thus, an ordering that respects the CPTs is equivalent to an ordering that respects the preference graph. We use the above definitions for simplicity. As we are using equivalent definitions, all results by \citet{boutJAIR2004} continue to hold.\\

\textbf{Definition 3: Dominance Query} \citep{boutJAIR2004}. Let $N$ be a CP-net, and let $o$ and $o'$ be associated outcomes. A \emph{dominance query} asks whether $N\vDash o\succ o'$ holds.\\

This dominance query holds if and only if there is a directed path $o'\leadsto o$ in the associated preference graph. That is, there is a sequence of outcomes $o'=o_1,o_2,...,o_m=o$ such that~$o_i$ and $o_{i+1}$ differ on the value of exactly one variable and $N\vDash o_{i+1}\succ o_i$. We call this type of outcome sequence an \emph{improving flipping sequence} (IFS). Thus, dominance queries can be reframed as a search for an IFS between the outcomes of interest; this is how we approach dominance queries later on in this paper. \citep{boutJAIR2004}\\

Finally, a short remark on notation. Let $N$ be a CP-net over variables $V$, and let $o$ be an associated outcome. Let $X\in V$ and $Y\subseteq V$. For ease of notation, let $o[X]$ denote the value assigned to $X$ in $o$ and let $o[Y]$ be the $|Y|$-tuple of values assigned to $Y$ in $o$. Further, if $Y=\{Y_1,Y_2,...,Y_k\}$ then let $\text{Dom}(Y)$ denote $\text{Dom}(Y_1)\times\text{Dom}(Y_2)\times\cdots\times\text{Dom}(Y_k)$. That is, $\text{Dom}(Y)$ consists of all $|Y|$-tuples of values that may be assigned to $Y$.

\subsection{Event Tree Representation}
Let $N$ be a CP-net over variables $V$. We have mentioned previously that the associated preference graph is an equivalent representation of this information. Another equivalent way of representing CP-nets is by an event tree \citep{edwaISR1983}. We use this alternate representation to motivate and construct our quantification of user preference in \S3.1. The \emph{event tree representation} of $N$, denoted $T(N)$, can be constructed in three steps.\\

First, put the variables in a topological order according to the CP-net structure,\break$V=\{V_1,...,V_n\}$. That is, Pa$(V_i)\subseteq \{V_1,...,V_{i-1}\}$. For the CP-net given in Example 1, there are two such orderings, $ABCD$ and $BACD$. We use $ABCD$ for simplicity.\\

Second, construct an event tree representing the successive events of $V_1$ taking a value, then $V_2$ taking a value, and so on up to $V_n$. The root node branches into $|\text{Dom}(V_1)|$ possibilities (each branch should be labelled with an associated element of $\text{Dom}(V_1)$). Then, each of these nodes branches into $|\text{Dom}(V_2)|$ possibilities (each labelled with an associated element of $\text{Dom}(V_2)$). And so on until each of $V_1,V_2,...,V_n$ have taken a value. The final tree has $|\Omega|$ root-to-leaf paths, corresponding to the outcomes. Figure 2 gives the event tree representation for the CP-net in Example 1 (ignore the branch weights for the moment).\\

Finally, the branches need to be labelled with the level of preference of the associated variable assignment. Suppose we are labelling the branch $b$, which represents that $X=x$ (for some $X\in V$). By inspecting the unique path from the root to the start of $b$, identify the values assigned to Pa$(X)$. From the appropriate row of $\text{CPT}(X)$, you can identify the position of preference of the choice $X=x$. If $x$ is the best choice under this assignment to~Pa$(X)$, then label $b$ with `$1^{st}$', if it is the second best, then label it `$2^{nd}$', and so on. For our running example, at the first stage, we would label the $A=a$ branch `$1^{st}$' and the $A=\bar{a}$ branch `$2^{nd}$' because of $\text{CPT}(A)$. Similarly, both $B=b$ branches have the label `$1^{st}$' and both $B=\bar{b}$ branches have the label `$2^{nd}$'. Now, consider the top-most instance of the tree branching into the options for $C$ ($c,\bar{c},\bar{\bar{c}}$). At this point $A$ and $B$ have been assigned values~$a$ and $b$ and so we are concerned with this (top) row of $\text{CPT}(C)$. From the CPT, we can see that, given the history of this path, $c$ is preferred to $\bar{c}$ is preferred to $\bar{\bar{c}}$, thus we give the $C=c$ branch the label `$1^{st}$', the $C=\bar{c}$ branch the label `$2^{nd}$', and the $C=\bar{\bar{c}}$ branch the label `$3^{rd}$'. Labelling the rest of the $C$ and $D$ branches is a similar process. However, for the $D$ branches we only need to look at the value previously taken by $C$ to determine which~$\text{CPT}(D)$ row to consult.\\

From the example used above, it is clear that $T(N)$ can become very large even for smaller CP-nets. As mentioned previously, we use this event tree representation to aid the construction of our outcome ranks in \S3.1. However, in \S4 we demonstrate that constructing this tree is not necessary for their calculation, so the exponential size of the trees is not a limitation.\\

\textbf{Remark 2.} This event tree representation, $T(N)$, is equivalent to the original CP-net (recalling that the CP-net consists of both the structure and the CPTs). Clearly, one can construct $T(N)$ from $N$; this process is described above. The key part to this claim is that one can reconstruct $N$ given $T(N)$. A sketch of this process is as follows. From $T(N)$ we can read off the domains of the variables and a topological ordering. Suppose $X_1,X_2,...,X_n$ is this topological ordering, then $\text{Pa}(X_i)\subseteq\{X_1,...,X_{i-1}\}$. If $Y\in\{X_1,...,X_{i-1}\}$, then $Y\in\text{Pa}(X_i)$ if and only if there are two direct paths $p$ and $p'$, of length $(i-1)$ that begin at the root of $T(N)$ and have the following property. By definition, $p$ and $p'$ assign values to $\{X_1,...,X_{i-1}\}$ only. We must have that $p$ and $p'$ differ only on the value assigned to $Y$ and that the preference order over $X_i$ is different for these two assignments. That is, the labelling of the branches corresponding to the different values of $X_i$ that come directly after path $p$ is different to those after $p'$. Identifying the parent sets determines the structure of~$N$ and it only remains to construct the CPTs. Given the parents of a variable and their respective domains, we know already the set of all parent assignments that corresponds to the rows of the CPTs. Given a row of $\text{CPT}(X_i)$, that is, an assignment of values to $\text{Pa}(X_i)$, say $u_i$, we need to recover the corresponding preference order over $X_i$. This can be done by finding any path $p$ of length $(i-1)$ that begins at the root of $T(N)$ and assigns $\text{Pa}(X_i)$ the values in $u_i$. The relevant preference order over $X_i$ can be read off from the labels on the branches corresponding to the different values of $X_i$ that come directly after $p$; the value of~$X_i$ assigned the label `$1^{st}$' is first in the preference order (most preferred), the value assigned the label~`$2^{nd}$' is second and so on. We have now completely constructed the CPTs and, thus, $N$.

\section{Outcome Ranks}
Given a CP-net representing the user's preferences, our aim is to quantify the user's preference for each outcome; we will call this value an outcome rank. These values should induce a consistent ordering of the outcomes. In most cases, CP-nets do not fully specify the user's preferences over the outcomes. Rather, there are usually several orderings of the outcomes that could match the user's preference (consistent orderings). Furthermore, given a basic CP-net and no further information, we are unable to judge any consistent ordering to be more likely than another to be the user's true preference ordering. Thus, if you wish to order the outcomes according to user preference, then you can do no better than to find \emph{any} consistent ordering.\\

\subsection{Calculating Outcome Ranks}
In this section, we introduce our outcome ranks (which successfully induce a consistent ordering of the outcomes). These are obtained using the event tree representation discussed in \S2.2. Specifically, we first weight the edges of the event tree representation and then read off the rank of an outcome from this weighted tree. The ranking we construct reflects user preference so more preferred outcomes have higher scores.\\

To motivate our weighting convention for the edges of $T(N)$, we must look at what determines the user's level of preference for an outcome $o$. The position of preference of the values taken by the individual variables, according to the CPTs, needs to be taken into account. However, according to the semantics of CP-nets, ancestor variables in the CP-net structure are more important to the user than their descendants \citep{boutJAIR2004}. Thus, if variable $A$ is an ancestor of variable $B$, then when quantifying user preference over outcomes, we must have a larger penalty for  a decrease of preference for $A$ than for a decrease of preference for $B$. Therefore, the position of variables in the CP-net structure will need to be taken into account in determining the user's level of preference for an outcome.\\

As we are allowing our CP-net variables to be multivalued, we must also take into account how domain size affects user preference. By the semantics of CP-nets, domain size should be independent of the importance of a variable. Suppose we have variables $X$ and~$Y$ such that $Y$ is a descendant of $X$ in the CP-net. Then any decrease of preference in $X$ should dominate any decrease of preference in $Y$, regardless of their domain sizes. Thus, our quantification of preference must also have this property.\\

Motivated by these restrictions imposed by the CP-net semantics, we have created the following weighting for the branches of the event tree representation of a CP-net. Let $N$ be a CP-net over variables $V=\{X_1,..,X_n\}$ and assume that this ordering of the variables is a topological ordering with respect to the structure of $N$. Now, consider the event tree representation of $N$, $T(N)$. Let $e$ be the edge of $T(N)$ that indicates variable $X_i$ takes value~$x_i$ given $X_1,...,X_{i-1}$ take values $x_1,...,x_{i-1}$. Use $p$ to denote the directed path from the root of $T(N)$ to the start of $e$, that dictates in turn that $X_1=x_1$, $X_2=x_2,...,X_{i-1}=x_{i-1}$. Let $u_i\in \text{Dom}(\text{Pa}(X_i))$ be the assignment of values to the parents of $X_i$ dictated by $p$. We attach the following weight to $e$:
\begin{equation} \label{TreeWeight}
\Bigg(\prod_{Y\in \text{Anc}(X_i)}\frac{1}{n_Y}\Bigg) (d_{X_i}+1) \frac{n_{X_i}-k+1}{n_{X_i}},
\end{equation}
which uses the following notation:
\begin{itemize}
\item $\text{Anc}(X_i)$ is the set of variables $Y\in V$ such that there is a directed path $Y\leadsto X_i$ in the structure of $N$ (these are referred to as the \emph{ancestors} of $X_i$),

\item $n_{X_i} := |\text{Dom}(X_i)|$,

\item $d_{X_i}$ is the number of distinct directed paths of any length in the structure of $N$ that originate at ${X_i}$ (the number of \emph{descendent paths} of $X_i$),

\item $k$ is the position of preference of the choice of $X_i=x_i$ given Pa$(X_i)=u_i$. So, if $X_i=x_i$ is the best choice for the user, then $k=1$, if it is the second best choice, then $k=2$, and so on. If it is the worst possible choice for $X_i$, then $k=|\text{Dom}(X_i)|$.
\end{itemize}

We refer to the leftmost product term in \eqref{TreeWeight} as the \emph{ancestral factor} of $X_i$, $AF_{X_i}$. This factor scales the weight down by the size of $X_i$'s ancestors' domains. The purpose of this is so that any decrease in preference of an ancestor will dominate a decrease in preference of~$X_i$, regardless of the size of the ancestor's domain relative to $|\text{Dom}(X_i)|$.\\

Consider the central term of \eqref{TreeWeight}, $(d_{X_i}+1)$. If $X$ is an ancestor of $Y$, then $d_X>d_Y$. An ancestor variable is more important to the user than its descendent variables, this term allocates these more important variables more weight. In particular, this term ensures that reductions in preference of an ancestor variable have larger penalties than reductions in preference of a descendant.\\

We refer to the rightmost product term in \eqref{TreeWeight} as the \emph{preference position} of the choice $X_i=x_i$ given Pa$(X_i)=u_i$, denoted $P_P\{X_i=x_i\hspace{0.1cm}|\hspace{0.1cm}\text{Pa}(X_i)=u_i\}$. This is a value in $\{ 1/n_{X_i}, 2/n_{X_i},...,(n_{X_i}-1)/n_{X_i}, 1\}$. This is simply a factor on the (0,1] scale indicating to what degree the user prefers this choice of value for $X_i$. This naturally impacts the user's preference for the overall outcome. This factor gets larger for more preferred values with the best value assigned preference position 1.\\

Notice that the preference position factor decreases in equal increments. Due to a lack of information provided by the CP-net, we cannot justify a more complex increment when quantitatively representing the user's preferences over $\text{Dom}(X_i)$. Consider a variable $A$ with $\text{Dom}(A)=\{a_1, a_2, a_3\}$ and CPT $a_1\succ a_2\succ a_3$. This could mean that, to the user, $a_2$ is \emph{slightly} worse than $a_1$, but $a_3$ is \emph{much} worse than $a_2$. Alternatively, it could be that $a_2$ is \emph{much} worse than $a_1$, but $a_3$ is only \emph{slightly} worse than $a_2$. We cannot determine which of these is the case due to lack of information, and so we assume that preference decreases in equal increments each time. In this situation, our preference positions would be $1, \frac{2}{3}$, and $\frac{1}{3}$ for $a_1, a_2,$ and $a_3$ respectively. \\

We refer to the event tree representation of $N$ weighted using the above convention as the \emph{weighted tree representation} of $N$, $W(N)$.\\

\textbf{Example 2.} We now return to the CP-net $N$, from Example 1 and corresponding event tree $T(N)$, given in \S2.2. Simple examination of the CP-net structure and CPTs gives us the following information:
\begin{equation}
\nonumber
\text{Anc}(A)=\varnothing, \hspace{0.2cm} \text{Anc}(B)=\varnothing, \hspace{0.2cm} \text{Anc}(C)=\{A, B\}, \hspace{0.2cm} \text{Anc}(D)=\{A, B, C\},
\end{equation}
\begin{equation}
\nonumber
n_A=2, \hspace{0.2cm} n_B=2, \hspace{0.2cm} n_C=3, \hspace{0.2cm} n_D=2,
\end{equation}
\begin{equation}
\nonumber
d_A=2, \hspace{0.2cm} d_B=2, \hspace{0.2cm} d_C=1, \hspace{0.2cm} d_D=0.
\end{equation}
From the $n_X$ values and the ancestor sets, we can calculate the ancestral factor of each variable:
\begin{equation}
\nonumber
AF_A=1, \hspace{0.2cm} AF_B=1,
\end{equation}
\begin{equation}
\nonumber
AF_C=\frac{1}{2}\times \frac{1}{2} = \frac{1}{4},
\end{equation}
\begin{equation}
\nonumber
AF_D=\frac{1}{2}\times \frac{1}{2}\times \frac{1}{3} = \frac{1}{12}.
\end{equation}
We can now use these values and the CPTs to directly calculate the edge weights and thus construct the weighted tree representation of this example. $W(N)$ is given in Figure 2 with the preference positions given in bold.\\

\begin{figure}
\includegraphics[scale=0.52]{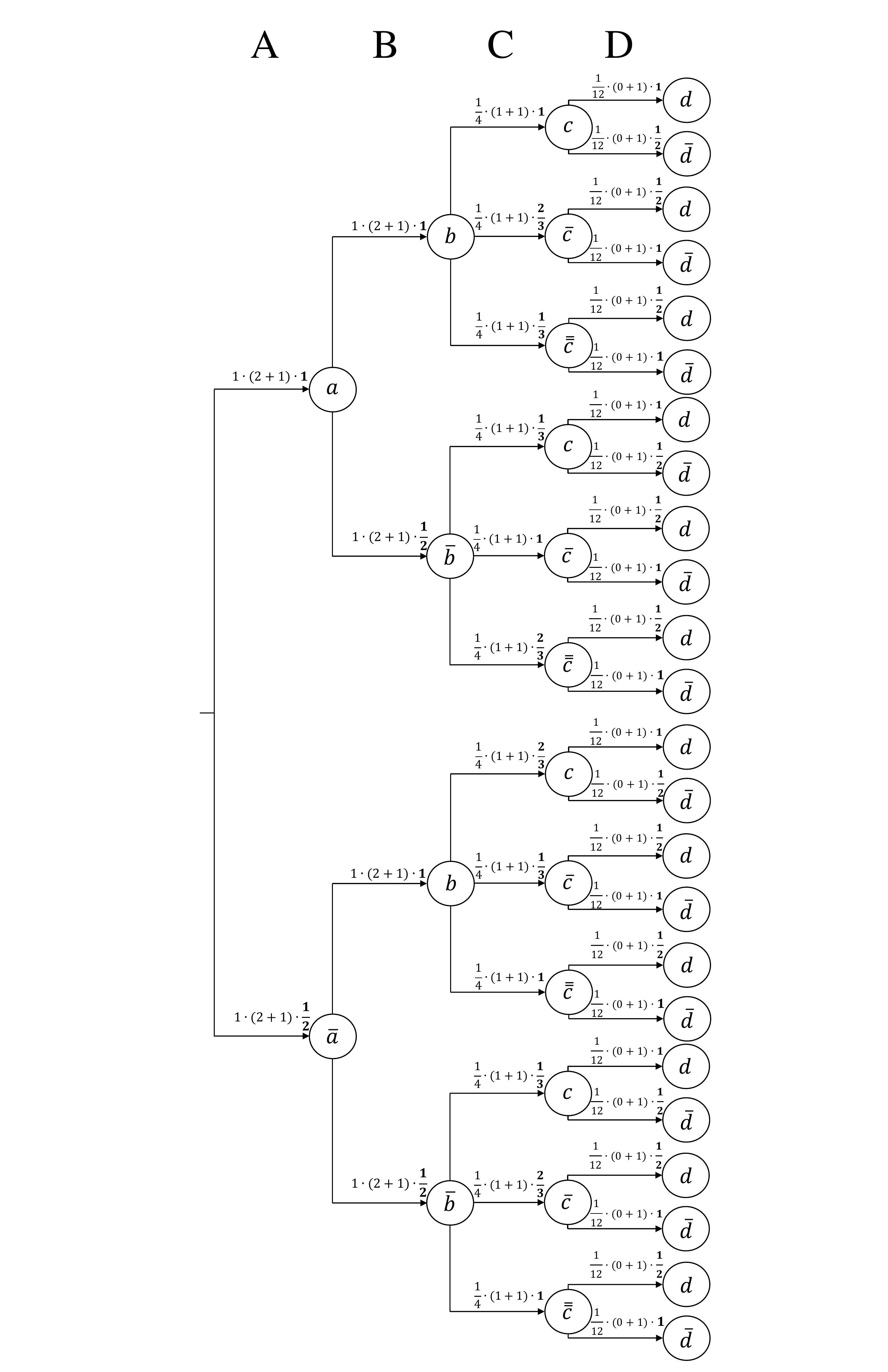}
\caption{Weighted Event Tree Example}
\end{figure}

By examining the weighted tree for this example, it can be seen that the weights attached to any two edges indicating the value taken by the same variable differ only on the preference position (the bolded number). Consider the set of edges leaving any node in the tree. By the definition of preference position, those edges indicating that the next variable takes a more preferred value will have larger weights. Thus, we can recover $T(N)$ given $W(N)$. As $T(N)$ is an equivalent representation to $N$, this shows that $W(N)$ is also an equivalent representation to $N$. Recall that $N$ is both the CP-net structure and the CPTs.\\

For ease of notation we shall, from this point on, simplify the notation for the weighted tree representation of $N$ from $W(N)$ to $W$ without ambiguity.\\

Now that we can construct the weighted tree representation of any given CP-net, we use this structure to define our quantitative measure of preference for any outcome.\\

\textbf{Definition 4: Rank.} Given a CP-net $N$, and an associated outcome $o$, we define the \emph{rank} of $o$, $r(o)$, to be the sum of the weights on the edges of the root-to-leaf path of $W$ that corresponds to $o$.\\

\textbf{Example 3.} Continuing on from Example 2, we calculate the ranks of several outcomes directly from $W$:
\begin{equation}
\nonumber
\begin{split}
r(\bar{a}b\bar{\bar{c}}\bar{d})=& \bigg[1\cdot(2+1)\cdot\frac{1}{2}\bigg]+ \bigg[1\cdot(2+1)\cdot 1 \bigg]+ \bigg[\frac{1}{4}\cdot (1+1)\cdot 1 \bigg] + \bigg[\frac{1}{12}\cdot (0+1)\cdot 1\bigg]\\
=& \hspace{0.1cm} \frac{61}{12},\\
 & \\
r(ab\bar{c}\bar{d}) =& \bigg[1\cdot(2+1)\cdot 1 \bigg]+ \bigg[1\cdot(2+1)\cdot 1 \bigg]+ \bigg[\frac{1}{4}\cdot (1+1)\cdot \frac{2}{3} \bigg] + \bigg[\frac{1}{12}\cdot (0+1)\cdot 1 \bigg]\\
=& \frac{77}{12},\\
 & \\
r(\bar{a}\bar{b}cd) =& \bigg[1\cdot(2+1)\cdot\frac{1}{2}\bigg]+ \bigg[1\cdot(2+1)\cdot \frac{1}{2} \bigg]+ \bigg[\frac{1}{4}\cdot (1+1)\cdot \frac{1}{3} \bigg] + \bigg[\frac{1}{12}\cdot (0+1)\cdot 1\bigg]\\
=& \frac{39}{12}.
\end{split}
\end{equation}

Recall that our aim was to assign higher values to more preferred outcomes. Thus, the relative sizes of these ranks are as we would expect as we can derive the following sequences of preference directly from the CPTs:
\begin{equation}
\nonumber 
ab\bar{c}\bar{d}\succ ab\bar{\bar{c}}\bar{d}\succ \bar{a}b\bar{\bar{c}}\bar{d}
\end{equation}
\begin{equation}
\nonumber
\bar{a}b\bar{\bar{c}}\bar{d}\succ \bar{a}b\bar{\bar{c}}d\succ \bar{a}bcd \succ \bar{a}\bar{b}cd
\end{equation}
Thus, we have $N\models ab\bar{c}\bar{d} \succ \bar{a}b\bar{\bar{c}}\bar{d} \succ \bar{a}\bar{b}cd$ and $r(ab\bar{c}\bar{d}) > r(\bar{a}b\bar{\bar{c}}\bar{d}) > r(\bar{a}\bar{b}cd)$.\\

\subsection{Obtaining Consistent Orderings from Outcome Ranks}
In this section we demonstrate how our outcome ranks can be used to obtain consistent orderings. This can be applied to the whole outcome set, in order to get a complete consistent ordering for the CP-net, or to any subset of the outcomes. Further, this method can be directly applied to CP-nets with additional plausibility constraints.\\

There are several methods of obtaining a consistent ordering in the existing literature, some of which we outline below. As discussed in \S1, \citet{domsIJCAI2003} use their approximations to obtain consistent orderings. The penalty function defined by \citet{liAAMS2011} could also be used to obtain a consistent ordering in this manner, although this is not mentioned in their paper. The utility function defined by \citet{mcgeAAAI2002} would induce a consistent ordering, if the utility was based upon CP-net preferences. \citet{sunESWA2017} utilise their complete preference table in order to obtain consistent orderings. \citet{boutJAIR2004} take yet another approach which is to construct a lexicographic ordering of the outcomes as follows. Let $N$ be a CP-net with variables $\{X_1,...,X_n\}$, listed such that a variable's parents come before the variable itself. Suppose we have two outcomes~$o_1$ and $o_2$, that have the same values for $X_1,...,X_k$ but differ on the value of $X_{k+1}$, say $o_1[X_{k+1}]=x_{k+1}$ and $o_2[X_{k+1}]=x'_{k+1}$. If, given the assignment of values to Pa$(X_{k+1})$ in both $o_1$ and $o_2$,~$\text{CPT}(X_{k+1})$ dictates that $x_{k+1}\succ x'_{k+1}$, then $o_1$ comes before $o_2$ in this ordering.\\

As we have constructed our outcome ranks to reflect user preference, they obey all entailed relations, as we wanted. Thus, our ranks induce a consistent ordering of the outcomes,~$\succ^*$. This $\succ^*$ is obtained simply by ordering the outcomes according to their rank, with outcomes with higher ranks considered to be more preferred. Proof of these claims is given below.\\

\textbf{Theorem 1.} Given a CP-net $N$, for any outcomes $o$ and $o'$, we have that $N\models o\succ o' \Rightarrow r(o)>r(o')$.\\

Proof given in Appendix A.\\

This tells us that if the CP-net dictates that the user prefers $o$ to $o'$, then $r(o)>r(o')$, that is, $o\succ^* o'$. In fact, we can say more than $r(o)>r(o')$; we can find a lower bound for the rank difference, $r(o)-r(o')$. Details of this lower bound are given in \S5.\\

\textbf{Corollary 1.}  Given a CP-net $N$, and two distinct associated outcomes $o$ and $o'$, $r(o)=r(o') \Rightarrow N\nvDash o\succ o'\wedge N\nvDash o'\succ o$ (we say that $o$ and $o'$ are \emph{incomparable}).\\

\noindent\textbf{Proof:} Theorem 1 tells us that for any two outcomes $o_1$ and $o_2$, $N\models o_1\succ o_2 \Rightarrow r(o_1)>~r(o_2)$, or equivalently $r(o_1)\leq r(o_2) \Rightarrow N\nvDash o_1\succ o_2$. Using this equivalent result gives us the following:
\begin{equation}
\nonumber 
r(o)=r(o') \Rightarrow (r(o)\leq r(o'))\wedge (r(o')\leq r(o)) \Rightarrow N\nvDash o\succ o'\wedge N\nvDash o'\succ o.
\end{equation}
$\blacksquare$\\

\textbf{Corollary 2.} Let $N$ be a CP-net. Let $\succ^*$ be the ordering of the outcomes of $N$ induced by the outcome ranks. Then $\succ^*$ is a consistent ordering of the outcomes with respect to $N$.\\

\noindent\textbf{Proof:} In order to show that $\succ^*$ is a consistent ordering, we need to show that, for any two outcomes $o_1$ and $o_2$, $N\vDash o_1\succ o_2 \implies o_1\succ^* o_2$. Theorem 1 shows that $N\vDash o_1\succ o_2 \implies r(o_1) > r(o_2)$. By definition of $\succ^*$, $r(o_1) > r(o_2) \implies o_1\succ^* o_2$. Thus, we have $N\vDash o_1\succ o_2 \implies o_1\succ^* o_2$ and so can conclude that $\succ^*$ is a consistent ordering of the outcomes. $\blacksquare$\\

We cannot guarantee $\succ^*$ is a strict order. There is a possibility that two distinct outcomes~$o$ and $o'$ could be assigned equal rank. However, Corollary 1 shows that this can only occur when  we do not know which the user prefers. If we want a strict ordering of the outcomes, then it is enough to force any outcomes with equal ranks into an arbitrary order. Any strict ordering of the outcomes obtained from $\succ^*$ in this manner is a consistent ordering of the outcomes as we have only altered the order of incomparable outcomes.\\

We have now introduced a novel method of quantifying user preference and obtaining a consistent outcome ordering given any (possibly multivalued) acyclic CP-net. Further, we can ensure that this is a strict ordering of the outcomes. From now on, when we refer to the outcome ordering induced by outcome ranks, we are referring to a strict ordering.\\

\textbf{Remark 3.} Going from a CP-net to a consistent ordering gives the impression of losing a great deal of information, especially as there are likely to be many consistent orderings and we have constructed one that is no better than any other. Moreover, the process of forcing our ordering to be strict arbitrarily discards several possible orderings. However, we have found that, given this consistent ordering, we can answer ordering and dominance queries directly, without needing to consult the CP-net. Further, we can use these ranks to improve the efficiency of answering dominance queries. We can also determine whether $o\succ^* o'$ is entailed by the CP-net ($N\models o\succ o'$) or constructed ($N\nvDash o\succ o'\wedge N\nvDash o'\succ o$), and update~$\succ^*$ given new (consistent) preference information, both without consulting the CP-net. In fact, despite constructing a consistent ordering somewhat arbitrarily, we have not lost any information at all. In this paper, we focus on demonstrating how these ranks can be used to improve the efficiency of answering dominance queries. The other results discussed above are explored in a forthcoming paper.\\

\textbf{Example 4} For the CP-net in Example 1, the ordering of the outcomes induced by the ranks is as follows:
\begin{equation}
\nonumber
\begin{split}
abcd\succ^* abc\bar{d} \succ^* ab\bar{c}\bar{d} \succ^* ab\bar{c}d
& \succ^* ab\bar{\bar{c}}\bar{d} \succ^* ab\bar{\bar{c}}d \succ^* 
a\bar{b}\bar{c}\bar{d} = \bar{a}b\bar{\bar{c}}\bar{d} \succ^* \\
a\bar{b}\bar{c}d = \bar{a}b\bar{\bar{c}}d \succ^* a\bar{b}\bar{\bar{c}}\bar{d} = \bar{a}bcd  & \succ^* 
a\bar{b}\bar{\bar{c}}d = \bar{a}bc\bar{d} \succ^* 
a\bar{b}cd = \bar{a}b\bar{c}\bar{d} \succ^*\\
 a\bar{b}c\bar{d} =
\bar{a}b\bar{c}d \succ^*  \bar{a}\bar{b}\bar{\bar{c}}\bar{d} \succ^* \bar{a}\bar{b}\bar{\bar{c}}d  & \succ^* \bar{a}\bar{b}\bar{c}\bar{d} \succ^* \bar{a}\bar{b}\bar{c}d \succ^* \bar{a}\bar{b}cd \succ^* \bar{a}\bar{b}c\bar{d}.\\
\end{split}
\end{equation}
We can obtain a strict ordering of the outcomes simply by replacing each $=$ with a $\succ^*$.\\

Our method of obtaining a consistent ordering using outcome ranks has the advantage of how easily it can be adapted to find a consistent ordering of any subset of the outcomes. Let~$N$ be a CP-net over variables $V$ and let $O$ be some subset of the outcomes, $O\subseteq\Omega$. Suppose we wish to put these outcomes $O$, in an order that agrees with everything the CP-net tells us about the user's preference. That is, we wish to find a strict order over $O$, $\succ_O$, such that for any two outcomes $o_1,o_2\in O$, we have that $N\vDash o_1\succ o_2 \implies  o_1\succ_O o_2$. To motivate the consistent ordering of subsets, consider an online shopping website displaying its products and suppose the seller wishes to promote a certain range of items; the seller would want exactly these items to appear on the first page. Putting these selected items into an order such that those items of more interest to the client are higher up, is an example of why we might want to put a specified subset of outcomes into a consistent order.\\

A consistent ordering of $O\subseteq\Omega$ can be obtained in exactly the same way we obtained a consistent ordering for $N$. For each $o\in O$, calculate the rank of $o$, $r(o)$, and then order $O$ according to rank value. To get a strict consistent ordering of $O$, force outcomes of equal rank into an arbitrary order. Call this strict ordering of $O$ $\succ_O$. We can see that $\succ_O$ is a consistent ordering of $O$ by using exactly the same reasoning we used to show that $\succ^*$ is a consistent ordering. Another way to obtain $\succ_O$ is to construct $\succ^*$ and then restrict the ordering to $O$ but this is less efficient.\\

In \S4, we present an algorithm that can calculate $r(o)$ for any outcome in time $O(|V|^4)$. Thus, a consistent ordering for a subset of size $k$ can be obtained as described above in $O(|V|^4k +k^2)$ time. \citet{boutJAIR2004} also proposed a solution to the problem of obtaining a consistent ordering for any subset of the outcomes. They proposed finding a consistent ordering of $O$ by repeatedly answering ordering queries (an \emph{ordering query} essentially asks, given two outcomes, find a consistent ordering of the two). Using this method, a consistent ordering for a subset of size $k$ can be obtained in $O(|V|k^2)$ time (as their method of answering ordering queries has complexity $O(|V|)$). Thus, for larger subsets of the outcomes, our method becomes more efficient. This is because every ordering query has complexity~$O(|V|)$, whereas, in our method, once the ranks are calculated the problem is reduced to a simple sorting task.  Note that the number of outcomes is at least $2^{|V|}$ (with equality only in the case of binary CP-nets), so subsets of the outcomes can get very large even for relatively small CP-nets.\\

A particularly interesting application of being able to consistently order any subset of the outcomes is finding a consistent ordering for CP-nets that have additional plausibility constraints. That is, a CP-net such that a specified proper subset of the outcomes, say $P\subsetneq \Omega$, are possible and the remainder are considered impossible. In reality, this kind of asymmetry in a CP-net system is commonplace. Consider, for example, an airline where there are no flights between specified dates and destinations with available business class seats, this would then be an impossible outcome.\\

\textbf{Lemma 1.} Given a CP-net $N$, and the further constraint that the only outcomes that are possible are those contained in $P\subset\Omega$, call the CP-net with these added constraints $N_C$. Let~$\succ_P$ be any strict ordering over $P$ such that, for all $o,o'\in P$, we have $N\vDash o\succ o' \implies o\succ_P o'$. Then $\succ_P$ is a consistent ordering for $N_C$.\\

\noindent\textbf{Proof:} In order to show $\succ_P$ to be a consistent ordering for $N_C$, it is enough to show that $N_C\vDash o\succ o'\implies o\succ _P o'$. We know that $N\vDash o\succ o' \implies o\succ_P o'$ holds so it will be sufficient to prove that $N_C \vDash o\succ o' \implies N\vDash o\succ o'$ holds. Recall that a CP-net entails the relation $o\succ o'$ if and only if there is a path $o'\leadsto o$ in the preference graph. Let $G_N$ be the preference graph for $N$ and let $G_{N_C}$ be the preference graph for $N_C$. Then $G_{N_C}$ is the induced subgraph of $G_N$ on outcomes $P$. Thus, if there exists a $o'\leadsto o$ path in $G_{N_C}$, then this will be a path (improving flipping sequence) in $G_N$ that exclusively uses outcomes in $P$. Therefore, there is a path $o'\leadsto o$ in $G_N$ and so we have that $N_C \vDash o\succ o' \implies N\vDash o\succ o'$ holds. $\blacksquare$\\

By Lemma 1, every consistent ordering (with respect to $N$) of the subset $P\subset\Omega$ is a consistent ordering for $N_C$. Thus, being able to obtain a consistent ordering of any subset of outcomes means we can also obtain a consistent ordering for any constrained CP-net.\\

In the case of CP-nets with additional plausibility constraints, any consistent ordering restricted to $P$ will be a consistent ordering for $N_C$. To obtain a consistent ordering of $P$ using outcome ranks you do not have to construct the full consistent ordering. In fact, you only need to calculate the edge weights for $W$ for edges that are on root-to-leaf paths corresponding to some $o\in P$. Thus, depending on the severity of the plausibility constraints, this could cut down calculations significantly.\\

\textbf{Example 5.} Consider the CP-net given in Example 1 with the following constraints. 
\begin{equation}
\nonumber
C=\{\neg\bar{a}, \neg(b\wedge c),  \neg(\bar{b}\wedge \bar{c}), \neg(\bar{b}\wedge\bar{\bar{c}}\wedge\bar{d})\}.
\end{equation}
In order to construct a consistent ordering for $N_C$, we only need to consider the restricted~$W$ seen in Figure 3 (edge weights are calculated exactly the same way as in Figure 2).\\

\begin{figure}[h]
\begin{center}
\includegraphics[scale=0.5]{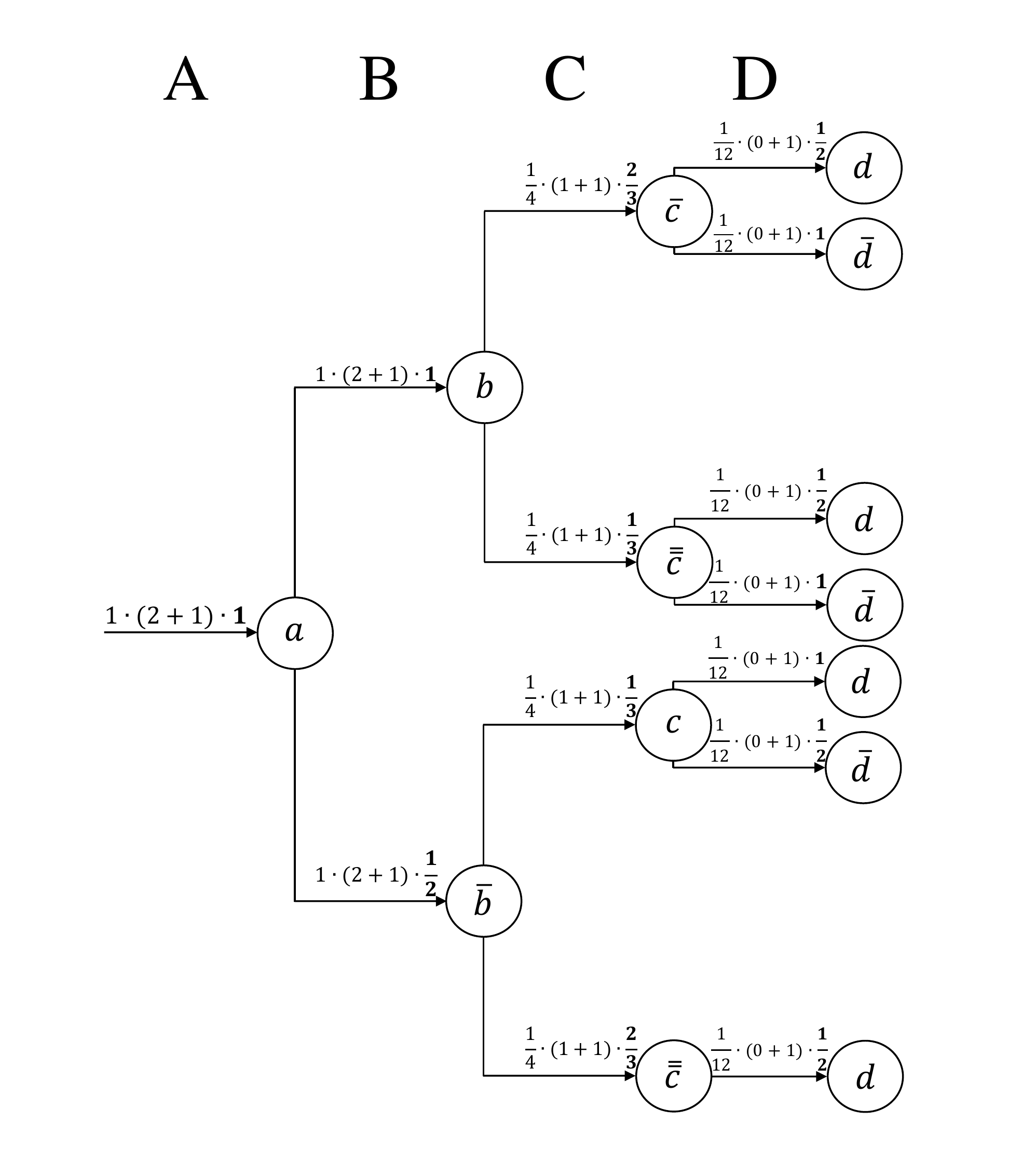}
\end{center}
\caption{Constrained CP-Net Example}
\end{figure}

From this much smaller tree, we calculate ranks as usual and order the possible outcomes~($P$) by their rank:
\begin{equation}
\nonumber
ab\bar{c}\bar{d}\succ_P ab\bar{c}d\succ_P ab\bar{\bar{c}}\bar{d}\succ_P ab\bar{\bar{c}}d\succ_P a\bar{b}\bar{\bar{c}}d\succ_P a\bar{b}cd\succ_P a\bar{b}c\bar{d}.
\end{equation}
This is a consistent ordering of $P$ for $N_C$. It can be seen by comparing $\succ_P$ to $\succ^*$ (given in Example 4) that $\succ_P$ is the restriction of $\succ^*$ to $P$.\\

We have now introduced a novel quantification of user preference from a given CP-net. We have shown that these ranks successfully reflect all entailed relations and how they can be used to obtain a consistent ordering of the outcomes. Further, we have shown that this method can be directly applied to obtain a consistent ordering of any subset of the outcomes or any CP-net with additional plausibility constraints.

\section{Rank Calculation Algorithms}

The outcome ranks defined in \S3.1 (Definition 4) are time consuming to calculate by hand even for fairly small CP-net examples. In this section, we present an algorithm for calculating the rank of any outcome. In the previous section, we used the event tree representation of CP-nets in both constructing our rank definition and in calculating example ranks. However, in this section, we show that ranks can be calculated directly from a CP-net input. Further, we can calculate the rank of any outcome in $O(|V|^4)$ time, where $|V|$ is the number of variables in the CP-net.\\

Algorithm 1 takes a CP-net and an outcome as inputs and outputs the rank of the given outcome. Recall, the rank of an outcome $o$, is the sum of the weights on the root to leaf path of~$W$ corresponding to $o$. Algorithm 1 calls two other algorithms. Algorithm 2 takes a variable $X$, and outputs the set of its ancestors in the CP-net, Anc$(X)=\{Y\hspace{0.1cm}|\hspace{0.1cm}\exists \text{ a directed path } Y\leadsto X \text{ in } N\}$. Algorithm 3 takes a variable $X$, and calculates the number of descendent paths of this variable in the CP-net, $d_X$. Algorithms 2 and 3 are given in Appendix B.3.\\

For the remainder of this section, suppose we have a CP-net $N$, over a set of variables $V=\{X_1,...,X_n\}$, which are in  a topological order with respect to the structure of $N$. We assume that $N$ is input to Algorithm 1 as a pair, $N=(A,CPT)$, where $A$ is the \emph{adjacency matrix} for the structure of $N$. That is, $A$ is a $|V|\times |V|$ matrix such that $A_{i,j}=1$ if there is an edge $X_i\rightarrow X_j$ in $N$, and $A_{i,j}=0$ otherwise. The second entry in the CP-net pair,~$CPT$, is the set of CPTs associated with $N$. We assume this to be input in a particular format, which is given in Appendix B.1 with an illustrative example. From this $CPT$ input, we can extract $|\text{Dom}(X_i)|$ for any $1\leq i\leq n$. To keep Algorithm 1 as readable as possible, we assume that, given $i$, we can obtain $|\text{Dom}(X_i)|$, rather than putting the details of how this is achieved (these details are given in Appendix B.1). We also leave the details of the format for input outcomes to Appendix B.1.\\

\newpage
\hrule
\vspace*{0.1cm}
\textbf{Algorithm 1: Rank Calculation Algorithm}
\vspace*{0.1cm}
\hrule
\vspace*{0.1cm}
\textbf{Inputs:} $N=(A,CPT)$ - CP-Net

\hspace{1.55cm} $\textbf{o}$ - Outcome
\vspace*{0.1cm}
\hrule
\vspace*{0.2cm}
\textbf{1} $r(\textbf{o}):=0$

\textbf{2} \textbf{for} $i$ in $\{1,2,...,|V|\}$ \hspace{2cm} \#Looping through the set of variables

\textbf{3} \hspace*{0.5cm} $Anc:=ancestor(i,A)$ \hspace{2cm}\#\emph{ancestor} function calls Algorithm 2

\hspace{7.05cm}\#$Anc$: set of ancestors of the current variable ($i$)

\textbf{4} \hspace*{0.5cm} $AF:=\prod_{Y\in Anc}\frac{1}{|\text{Dom}(Y)|}$

\textbf{5} \hspace*{0.5cm} $d:=DP(i, A)$ \hspace{2cm}\#\emph{DP} function calls Algorithm 3

\hspace{5.65cm}\#$d$: number of descendent paths of the current variable

\textbf{6} \hspace*{0.5cm} $Pa:=\{j\hspace{0.1cm}|\hspace{0.1cm} A_{j,i}=1\}$ \hspace{2cm}\#Set of parents of the current variable

\textbf{7} \hspace*{0.5cm} $\textbf{u}:=\textbf{o}[Pa]$ \hspace{2cm}\#Values taken by the parents of $i$ in outcome $\textbf{o}$

\textbf{8} \hspace*{0.5cm} $\textbf{order}:=CPT[i][\textbf{u}]$ \hspace{2cm}\#Preference order over $i$ given that $Pa=\textbf{u}$

\textbf{9} \hspace*{0.5cm} $k:=\textbf{order}[\textbf{o}[i]]$ \hspace{2cm}\#$\textbf{o}[i]$: value taken by $i$ in outcome $\textbf{o}$

\hspace{5.8cm}\#$k$: position of preference of $\textbf{o}[i]$ in the previous order

\textbf{10} \hspace*{0.25cm} $P_P:=\frac{|\text{Dom}(X_i)|-k+1}{|\text{Dom}(X_i)|}$

\textbf{11} \hspace*{0.25cm} $r(\textbf{o})= r(\textbf{o}) + AF\cdot (d+1)\cdot P_P$

\textbf{12} \textbf{return} $r(\textbf{o})$
\vspace*{0.1cm}
\hrule
\vspace*{0.5cm}

Algorithm 1 takes the CP-net $N$, and some outcome $o$, and outputs the rank of this outcome $r(o)$. It calculates $r(o)$ by setting the value of $r(o)$ to 0 (step \textbf{1}) and successively adding the edge weights of the root to leaf path in $W$ that corresponds to $o$ (steps \textbf{2-11}). A more detailed explanation of how Algorithm 1 works and why it is correct can be found in Appendix B.2.\\

We have used $W$ here and in Appendix B.2 to help explain what Algorithm 1 is doing and to show why it is correct. However, notice that the algorithm itself does not utilise $W$ at any point and instead works directly with the CP-net to obtain the rank. This shows that, whilst the event tree representation was useful in motivating and explaining our ranking system, constructing the tree is not a necessary step in calculating the rankings. This is reassuring as it is clear from the relatively small CP-net given in Example 1, that $W$ quickly becomes large.\\

For a CP-net $N$, with $n$ variables, Algorithm 2 and Algorithm 3 both have complexity~$O(n^3)$ and Algorithm 1 has complexity $O(n^4)$. Thus, for any associated outcome $o$, we can compute $r(o)$ in $O(n^4)$ time; that is, finding the rank of an outcome is tractable.\\

\textbf{Remark 4.} We could use Algorithm 1 to produce a consistent ordering, given a CP-net~$N$, as shown by Corollary 2. This is done by using Algorithm 1 to calculate the rank of each outcome, and then sorting these outcomes into rank-order. However, to obtain a consistent ordering in this manner, we are applying Algorithm 1 $|\Omega|$ many times, making the time complexity in terms of $|\Omega|$. As $|\Omega|\geq 2^n$ (with equality only in the case of binary CP-nets) this is not a tractable method. This is unsurprising as putting $|\Omega|$ objects into an order will always have time complexity in terms of $|\Omega|$ (intractable). Our aim is to use these ranks (algorithms) to improve the efficiency of dominance testing, which, as shown in \S5, does not require a consistent ordering. Thus, we are not concerned by this lack of tractability.

\section{Rank Pruning for Dominance Queries}

In \S3, we constructed an outcome rank that reflects all entailed relations, that is, $N\vDash o_1\succ o_2$ $\implies r(o_1)>r(o_2)$. In this section, we demonstrate how these ranks can be used to improve the efficiency of dominance testing. We first show that this statement is not all we can say about the difference in ranks of $o_1$ and $o_2$. We can also identify a lower bound on the difference in rank values as detailed below.\\

\textbf{Definition 5: Least Rank Improvement.} Let $N$ be a CP-net over variables $V$. For any $X\in V$, we define the \emph{least rank improvement} of $X$, denoted $L(X)$, as
\begin{equation}
\nonumber
L(X)= AF_X(d_X+1)\frac{1}{n_X} - \sum_{Y\in \text{Ch}(X)}AF_Y(d_Y+1)\frac{n_Y-1}{n_Y},
\end{equation}
where, for any $X\in V$, $n_X=|\text{Dom}(X)|$ and Ch$(X)=\{Y\in V\hspace{0.1cm}|\hspace{0.1cm} X\in \text{Pa}(Y)\}$. We call Ch$(X)$ the \emph{children} of $X$.\\

This value $L(X)$, is interpreted as the least possible increase in rank that can result from flipping $X$ to a more preferred value. That is, $L(X)$ corresponds to the rank increase of the improving $X$ flip $\alpha\rightarrow \beta$ ($L(X)=r(\beta)-r(\alpha)$), where $X$ only increases in preference by one preference position and every $Y\in \text{Ch}(X)$ goes from being the most preferred value to the least preferred value. Note that for all other variables $Z$, the value taken by $Z$ and its associated preference position must be identical in $\alpha$ and $\beta$. As $\beta$ must be preferred to $\alpha$, we would expect $L(X)$ to be a strictly positive value. This is shown to hold by the following Lemma.\\

\textbf{Lemma 2.} Let $N$ be a CP-net over variables $V$. For any $X\in V$, $L(X)>0$.\\

Proof in Appendix A.\\

Least rank improvement terms can be used to find a lower bound on the difference in rank implied by entailment. That is, given $N\vDash o_1\succ o_2$, Theorem 1 tells us that $r(o_1)>r(o_2)$, however, using these $L(X)$ terms we can find a lower bound for $r(o_1)-r(o_2)$.\\

\textbf{Corollary 3.} Let $N$ be a CP-net over variables $V$. Let $o_1$ and $o_2$ be associated outcomes and $D=\{X\in V\hspace{0.1cm}|\hspace{0.1cm} o_1[X]\neq o_2[X]\}$. Then,
\begin{equation}
\nonumber
N\vDash o_1\succ o_2 \implies r(o_1)-r(o_2)\geq \sum_{X\in D}L(X)>0.
\end{equation}

Proof in Appendix A.\\

\textbf{Definition 6: Least (Entailed) Rank Difference.} Let $N$ be a CP-net over variables~$V$, and let $o_1,o_2$ be associated outcomes. Let $D=\{X\in V\hspace{0.1cm}|\hspace{0.1cm} o_1[X]\neq o_2[X]\}$. The \emph{least (entailed) rank difference} of $o_1$ and $o_2$, denoted $L_D(o_1,o_2)$, is defined as follows:
\begin{equation}
\nonumber
L_D(o_1,o_2) = \sum_{X\in D}L(X).
\end{equation}

We now illustrate how Corollary 3 can be used to improve the efficiency of answering dominance queries. A \emph{dominance query} \citep{boutJAIR2004} asks whether the user prefers one outcome to another. That is, given a CP-net $N$, and two associated outcomes $o$ and~$o'$, `Does $N\vDash o\succ o'$ hold?' is a dominance query. If $N\vDash o\succ o'$, then the user prefers $o$ to~$o'$ and so $o$ comes before $o'$ in all consistent orderings. As dominance queries require us to consider all consistent orderings (unlike previously since we have been concerned only with finding an arbitrary consistent ordering), they are very complex to answer \citep{boutJAIR2004, goldJAIR2008}. This is because, to answer the dominance query $N\vDash o\succ o'$?, you need to prove either that $o$ comes before $o'$ in every consistent ordering or, alternatively, that there exists a consistent ordering where $o'$ comes before $o$. Unless one is lucky enough to construct a consistent ordering where $o'$ comes before $o$, this cannot be answered by considering a single arbitrary consistent ordering.\\

Suppose we have a CP-net $N$, and we wish to answer the dominance query $N\vDash o\succ o'$?. There are three possibilities, either $N\vDash o\succ o'$, $N\vDash o'\succ o$, or $N\nvDash o\succ o' \wedge N\nvDash o'\succ o$. We can get at least halfway to answering this dominance query by calculating the ranks of~$o$ and~$o'$ and their least rank difference. As shown in Corollary 3, if $r(o')+L_D(o,o')>r(o)$, then $N\nvDash o\succ o'$ and the answer to the dominance query is no. If $r(o)\geq r(o')+L_D(o,o')$, then, by Theorem 1 and Lemma 2, $N\nvDash o'\succ o$ and so it remains to determine whether $N\vDash o\succ o'$ or $N\nvDash o\succ o' \wedge N\nvDash o'\succ o$.\\

Recall that $N\vDash o\succ o'$ if and only if there is a directed path $o'\leadsto o$ in the preference graph of~$N$. This directed path corresponds to a sequence of outcomes $o'=o_1, o_2,..., o_m=o$, such that $o_i$ and $o_{i+1}$ differ on the value of exactly one variable and $N\vDash o_{i+1}\succ o_i$. We call this an \emph{improving flipping sequence} (IFS) from $o'$ to $o$. Therefore, a dominance query can be reframed as a search for an IFS in the preference graph of $N$. \citep{boutJAIR2004}\\

There have been several techniques introduced to improve the efficiency of searching for a flipping sequence \citep{boutJAIR2004, liAAMS2011, alleJAIR2017}. We propose using our outcome ranks to impose an upper bound on such searches, in order to improve efficiency. Ideally, our upper bound would be implemented alongside other methods of improving search efficiency. We evaluate the performance of our method in combination with different techniques in \S6. However, here we illustrate how this upper bound works with a basic search method.\\

Returning to the dominance query $N\vDash o\succ o'$?, suppose we have already confirmed that $r(o)\geq r(o') + L_D(o,o')$. We can answer this dominance query by determining whether or not there exists an IFS from $o'$ to $o$. Note that if $o'=o_1, o_2,..., o_n=o$ is such an IFS, then, by Corollary 3, $o_i$ must satisfy $r(o)\geq r(o_i) + L_D(o,o_i)$; this is what enforces an upper bound on the search.  The method for determining whether such a sequence exists is as follows:\\

For any outcome $o^*$, define $F(o^*):= \{o\hspace{0.1cm}|\hspace{0.1cm}o^*\rightarrow o \text{ is an improving flip}\}$. That is, $F(o^*)$ is the set of outcomes $o$, that differ from $o^*$ on exactly one variable and $N\vDash o\succ o^*$. This set can be evaluated by inspecting the appropriate rows of the CPTs of $N$. First, evaluate $F(o')$, this is all outcomes that can be reached from $o'$ in one improving flip. If $o\in F(o')$, then clearly there is an $o'\leadsto o$ IFS and the answer to the dominance query is yes, $N\vDash o\succ o'$. If $o\not\in F(o')$, then we cannot reach $o$ from $o'$ in one improving flip and the next step is to determine whether it can be reached in two improving flips. However, before looking at all outcomes that can be reached from $F(o')$ in a further improving flip, there may be some search directions that can already be dismissed using our upper bound. For each $o^*\in F(o')$ evaluate $r(o^*)+L_D(o,o^*)$. Any outcome $o^*$ such that $r(o^*)+L_D(o,o^*)> r(o)$ is not on an $o'\leadsto o$ IFS, so it is unnecessary to evaluate what outcomes can be reached by improving flips from $o^*$. Let $\text{Flip}_1 = \{ o^*\in F(o')\hspace{0.1cm}|\hspace{0.1cm} r(o^*)+L_D(o^*,o)\leq r(o)\}$.\\

Let $F(\text{Flip}_1)=\bigcup_{o^*\in \text{Flip}_1}F(o^*)$. If $o\in F(\text{Flip}_1)$, then $o$ can be reached from $o'$ in two improving flips. That is, there is a length two IFS from $o'$ to $o$, so the answer to the dominance query is yes, $N\vDash o\succ o'$. If not, then we move on to looking at whether $o$ can be reached in three improving flips. Again, we may be able to eliminate certain search directions by removing any outcomes $o^*$, such that $r(o^*)+L_D(o,o^*) > r(o)$ before we continue to search. Let $\text{Flip}_2=\{o^*\in F(\text{Flip}_1)\hspace{0.1cm}|\hspace{0.1cm} r(o^*)+L_D(o,o^*)\leq r(o)\}$.\\

We continue to repeat this process until either $o$ is reached, so the answer to the dominance query is yes ($N\vDash o\succ o'$), or we reach some $\text{Flip}_i=\varnothing$, in which case the answer to the dominance query is no ($N\nvDash o\succ o'$). The upper bound means that we can stop considering an improving flipping sequence as soon we reach an outcome $o^*$, such that $r(o^*)+L_D(o,o^*)$ exceeds $r(o)$, rather than pursuing all unsuccessful paths until they reach the optimal outcome (where all IFS terminate). Visualising a consistent ordering induced by the ranks as a list of outcomes, we know that $o$ is above $o'$ and an IFS invariably moves up the list. This upper bound restricts the search area to the $o\rightarrow o'$ segment of this list, as searching stops as soon as you reach any outcome above $o$. The maximum possible number of steps to this search process is equal to the length of the $o\rightarrow o'$ list segment, so it will always terminate in finite time. Note that if we reach an outcome $o^*$, in the search that we have previously considered, then we can dismiss it as we have already considered all possible IFSs that can emanate from $o^*$.\\

\textbf{Example 6.} We now use the CP-net given in Example 1 to illustrate our method of answering dominance queries.\\

Does $N\vDash \bar{a}b\bar{\bar{c}}d\succ \bar{a}b\bar{c}\bar{d}$ hold? First, we evaluate the ranks of the two outcomes, which can be done by consulting $W$, given in Figure 2, or using Algorithm 1:
\begin{equation}
\nonumber
r(\bar{a}b\bar{\bar{c}}d)=\frac{121}{24}, \hspace{0.5cm} r(\bar{a}b\bar{c}\bar{d})=\frac{114}{24}.
\end{equation}
We must also calculate $L_D(\bar{a}b\bar{\bar{c}}d,\bar{a}b\bar{c}\bar{d})$. We calculate $L(X)$ for all $X\in V$:
\begin{equation}
\nonumber
L(A)=\frac{7}{6}, \hspace{0.25cm} L(B)=\frac{7}{6}, \hspace{0.25cm} L(C)=\frac{1}{8}, \hspace{0.25cm} L(D)=\frac{1}{24}.
\end{equation}
Then, we calculate $L_D(\bar{a}b\bar{\bar{c}}d,\bar{a}b\bar{c}\bar{d})$:
\begin{equation}
\nonumber
L_D(\bar{a}b\bar{\bar{c}}d,\bar{a}b\bar{c}\bar{d}) = \sum_{X\in\{C,D\}}L(X) = \frac{1}{6}.
\end{equation}
As $r(\bar{a}b\bar{\bar{c}}d)>r(\bar{a}b\bar{c}\bar{d})+ L_D(\bar{a}b\bar{\bar{c}}d,\bar{a}b\bar{c}\bar{d})$, to answer the dominance query we will need to determine whether there exists an IFS from $\bar{a}b\bar{c}\bar{d}$ to $\bar{a}b\bar{\bar{c}}d$.\\

The first step is to evaluate $F(\bar{a}b\bar{c}\bar{d})$. From the CPTs, we can see that only $A$ and $C$ can be changed into a more preferred position. So we have $F(\bar{a}b\bar{c}\bar{d})=\{ ab\bar{c}\bar{d}, \bar{a}bc\bar{d}, \bar{a}b\bar{\bar{c}}\bar{d}\}.$ As $\bar{a}b\bar{\bar{c}}d\not\in F(\bar{a}bcd)$, we cannot reach $\bar{a}b\bar{\bar{c}}d$ from $\bar{a}b\bar{c}\bar{d}$ in one improving flip. We now calculate $r(o)+L_D(\bar{a}b\bar{\bar{c}}d,o)$ for each $o\in F(\bar{a}b\bar{c}\bar{d})$. Again, we use $W$ or Algorithm 1 to calculate the ranks and we can use the $L(X)$ values calculated above to find the $L_D$ terms.
\begin{equation}
\nonumber
r(ab\bar{c}\bar{d}) + L_D(\bar{a}b\bar{\bar{c}}d,ab\bar{c}\bar{d}) = \frac{154}{24} + \bigg(\frac{7}{6}+\frac{1}{8}+\frac{1}{24}\bigg)=\frac{186}{24},
\end{equation}
\begin{equation}
\nonumber
r(\bar{a}bc\bar{d}) + L_D(\bar{a}b\bar{\bar{c}}d,\bar{a}bc\bar{d}) = \frac{117}{24} + \bigg(\frac{1}{8}+\frac{1}{24}\bigg)=\frac{121}{24}, \hspace{0.1cm} r(\bar{a}b\bar{\bar{c}}\bar{d}) + L_D(\bar{a}b\bar{\bar{c}}d,\bar{a}b\bar{\bar{c}}\bar{d}) = \frac{122}{24} + \bigg(\frac{1}{24}\bigg)=\frac{123}{24}.
\end{equation}
As $ab\bar{c}\bar{d}$ and $\bar{a}b\bar{\bar{c}}\bar{d}$ both satisfy $r(o)+L_D(\bar{a}b\bar{\bar{c}}d,o)>r(\bar{a}b\bar{\bar{c}}d)$, we do not need to pursue these search directions further (as they will not lie on an IFS from $\bar{a}b\bar{c}\bar{d}$ to $\bar{a}b\bar{\bar{c}}d$). Thus, we have $\text{Flip}_1=\{\bar{a}bc\bar{d}\}$.\\

Next, we look at which outcomes can be reached from $\text{Flip}_1$ in a single improving flip. This is in order to see whether $\bar{a}b\bar{\bar{c}}d$ can be reached from $\bar{a}b\bar{c}\bar{d}$ in two improving flips. In this case, $F(\text{Flip}_1)= F(\bar{a}bc\bar{d})$. By inspecting the CPTs we find $ F(\text{Flip}_1)= F(\bar{a}bc\bar{d})=\{abc\bar{d}, \bar{a}b\bar{\bar{c}}\bar{d}, \bar{a}bcd\}$. As $\bar{a}b\bar{\bar{c}}d\not\in F(\text{Flip}_1)$, we cannot reach $\bar{a}b\bar{\bar{c}}d$ from $\bar{a}b\bar{c}\bar{d}$ in two improving flips. Evaluate the ranks and $L_D$ terms of the outcomes in $F(\text{Flip}_1)$:
\begin{equation}
\nonumber
r(abc\bar{d})+L_D(\bar{a}b\bar{\bar{c}}d,abc\bar{d}) = \frac{157}{24} + \bigg(\frac{7}{6}+\frac{1}{8}+\frac{1}{24}\bigg)=\frac{189}{24}, 
\end{equation}
\begin{equation}
\nonumber
r(\bar{a}b\bar{\bar{c}}\bar{d})+L_D(\bar{a}b\bar{\bar{c}}d,\bar{a}b\bar{\bar{c}}\bar{d}) = \frac{122}{24}+\bigg(\frac{1}{24}\bigg)=\frac{123}{24}, \hspace{0.1cm} r(\bar{a}bcd)+L_D(\bar{a}b\bar{\bar{c}}d,\bar{a}bcd)= \frac{118}{24}+\bigg(\frac{1}{8}\bigg)=\frac{121}{24}.
\end{equation}
 As $abc\bar{d}$ and $\bar{a}b\bar{\bar{c}}\bar{d}$ both satisfy $r(o)+L_D(\bar{a}b\bar{\bar{c}}d,o)>r(\bar{a}b\bar{\bar{c}}d)$, we can stop searching in these directions. Alternatively, for $\bar{a}b\bar{\bar{c}}\bar{d}$, we could have dismissed this outcome immediately as we have considered it previously. This leaves us with $\text{Flip}_2=\{\bar{a}bcd\}$.\\

To see if we can reach $\bar{a}b\bar{\bar{c}}d$ from $\bar{a}b\bar{c}\bar{d}$ in three improving flips, we now evaluate $F(\text{Flip}_2)$: $F(\text{Flip}_2)=F(\bar{a}bcd)=\{abcd, \bar{a}b\bar{\bar{c}}d\}$.\\

We have $\bar{a}b\bar{\bar{c}}d\in F(\text{Flip}_2)$, thus $\bar{a}b\bar{\bar{c}}d$ can be reached from $\bar{a}b\bar{c}\bar{d}$ in three improving flips. That is, there is an IFS from $\bar{a}b\bar{c}\bar{d}$ to $\bar{a}b\bar{\bar{c}}d$ of length three, and so the answer to our dominance query is yes, $N\vDash \bar{a}b\bar{\bar{c}}d\succ \bar{a}b\bar{c}\bar{d}$ holds. A helpful way of visualising this method is the search tree given in Figure 4.\\

\begin{figure}[h]
\centering
\includegraphics[scale=0.4]{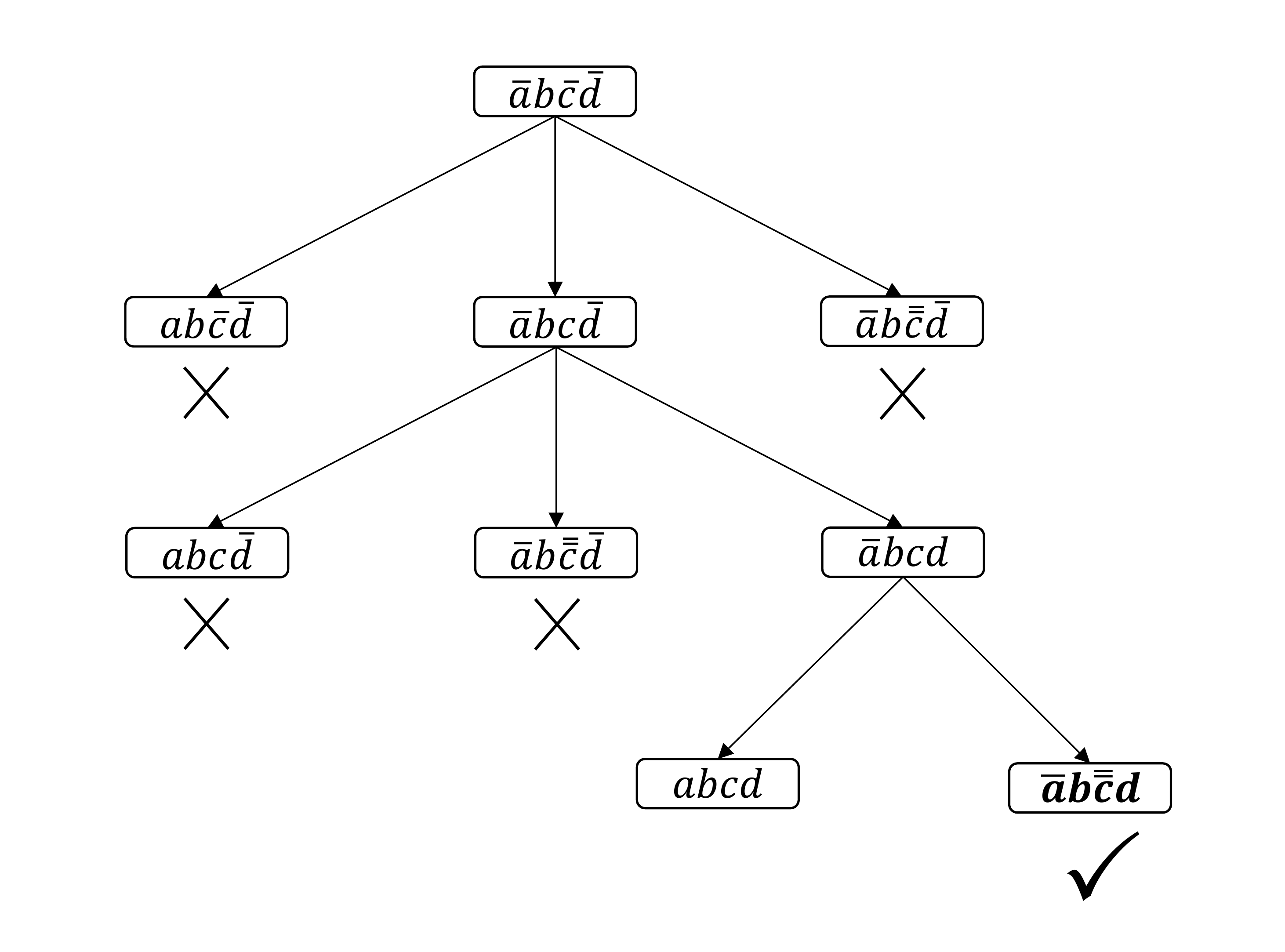}
\caption{Dominance Query Search Tree Example}
\end{figure}

The method we have described in this section uses a basic search tree method for finding an IFS \citep{boutJAIR2004}. Rank values and $L_D$ terms are used to prune certain branches as we construct the tree in order to improve search efficiency. Suffix fixing and least variable flipping, introduced by \citet{boutJAIR2004}, as well as the penalty-based pruning introduced by \citet{liAAMS2011} can all be viewed as methods of pruning this search tree. Ideally, our rank-based pruning method would be implemented alongside some of these other pruning methods for a more efficient dominance testing process. We give an experimental comparison of the performance of our rank pruning with some of these different pruning techniques and their combinations in \S6 in order to evaluate the most efficient pruning schema for dominance testing. \citet{alleJAIR2017} introduce the idea of only searching for an IFS to a certain depth. This could clearly be applied to our dominance testing procedure, one would simply stop searching once the specified depth was reached. They find experimentally that for relatively small binary CP-nets, the longest possible IFS has length $\left \lfloor{\frac{1}{4}(|V|+1)^2}\right \rfloor $. However, it is not proven that this holds for all binary CP-nets and it does not hold for multivalued CP-nets in general. Thus, if such a depth bound was incorporated with our dominance testing procedure, it would lose completeness.

\section{Experimental Evaluation of Pruning Measures}
In \S5 we showed how our outcome ranks can be used to make dominance testing more efficient by pruning the search tree. In this section, we evaluate the performance of our rank pruning, in comparison with the existing pruning methods. We also examine the performance of all possible combinations of these methods, in order to determine the most effective pruning schema for dominance testing. We first give the details of our experiments, then analyse the performance results of the different dominance testing methods. These results show our rank pruning to be the best of the individual methods, and the most important to include when considering combinations of techniques.\\

Before describing the experiments, we must formalise the notions of dominance query search trees and the pruning these trees. In \S5, we mentioned that our method of answering dominance queries, using ranks to eliminate certain directions, could be viewed as building up a search tree and using outcome ranks to prune this tree as it is constructed. We explain this idea more explicitly below.\\

Given the dominance query $N\vDash o\succ o'$, we want to determine whether $o$ is reachable from $o'$ in the preference graph, $G_N$. We do this by building the \emph{dominance query search tree} $G(o')$, until either $o$ is reached (and so the dominance query is true) or it cannot be constructed further (and so the dominance query is false). This search tree is constructed as follows. Start with $o'$ as the root of the tree. Select some leaf $\bar{o}$, and for every improving flip,~$o^*$, of $\bar{o}$ that is not already in $G(o')$ add the edge $\bar{o}\rightarrow o^*$ to the tree. We now say that $\bar{o}$ has been considered. Repeat this process until either $o$ is reached (the dominance query is true) or all leaves have been considered (the dominance query is false). This method successfully answers the dominance query because, when $G(o')$ is fully constructed, we have $\bar{o}\in G(o')$ if and only if $\bar{o}$ is reachable from $o'$ in $G_N$.\\

We can use outcome ranks to prune this tree as it is constructed (without affecting the completeness of the search) as follows. When considering leaf $\bar{o}$, any improving flip, $o^*$, satisfying $r(o^*) + L_D(o,o^*)>r(o)$ can be pruned from the tree. That is, $o^*$ does not need to be added to the tree because, by Corollary 3, we know that searching in this direction will not lead to $o$. This is fundamentally the same as the process described in \S5 for answering dominance queries with outcome ranks. However, in \S5 we considered \emph{all} not-pruned leaves of minimal depth (in the search tree) at once, for ease of explanation. Here, we consider one leaf at a time and do not specify how the leaf should be selected. We discuss the choice of leaf prioritisation in \S5.1.

\subsection{Experiment}
There are many existing methods to improve dominance testing efficiency. We have chosen to compare rank pruning to the other methods for pruning the search tree that preserve search completeness. This means that we are comparing our rank pruning to penalty pruning by \citet{liAAMS2011} and suffix fixing by \citet{boutJAIR2004}. We have excluded from our comparisons, least variable flipping by \citet{boutJAIR2004} and the depth bound on flipping sequences proposed by \citet{alleJAIR2017}, as they do not preserve completeness. We also do not consider the model checking method introduced by \citet{santAAAI2010}, the composition of preference tables introduced by \citet{sunESWA2017}, or the CP-net preprocessing method, forward pruning, by \citet{boutJAIR2004}.\\

\emph{Suffix fixing} \citep{boutJAIR2004} prunes the dominance query search tree as follows. Suppose we are answering the dominance query $N\vDash o\succ o'$ by constructing $G(o')$. Let~$N$ be a CP-net over variables $V$ and suppose $\{X_1,X_2,...,X_n\}$ is a topological ordering of $V$. The~$k^{th}$ \emph{suffix} of any outcome $o^*$ is $o^*[X_k,X_{k+1},...,X_n]$. As we construct $G(o')$, when considering a leaf $\bar{o}$, that has the same $k^{th}$ suffix as $o$, any improving flips of $\bar{o}$ that do not have the same $k^{th}$ suffix are pruned. This pruning condition preserves search completeness as \citet{boutJAIR2004} proved the following. If $o$ and $o'$ have the same $k^{th}$ suffix and $N\vDash o\succ o'$, then there exists an improving flipping sequence $o'=o_1,o_2,...,o_m=o$, such that every $o_i$ has the same $k^{th}$ suffix as $o$ and $o'$.\\

\emph{Penalty pruning} by \citet{liAAMS2011} is based upon their penalty function for outcomes. This penalty function is similar to our rank values in that it quantifies user preference; outcomes more preferred by the user have smaller penalty values. The formula for these penalties is
\begin{equation}
\nonumber
pen(o)=\sum_{X\in V}w_Xp_X^o
\end{equation}
where $w_X$ is the importance weight
\begin{equation}
\nonumber
w_X=\sum_{Y\in\text{Ch}(X)}w_Y(|\text{Dom}(Y)|-1).
\end{equation} 
The $p_X^o$ term is the degree of penalty of $X$ with respect to $o$. That is, if $o[X]$ is the most preferred value of $X$, given $\text{Pa}(X)=o[\text{Pa}(X)]$, then $p^o_X=0$. If $o[X]$ is the second most preferred value of $X$, then $p^o_X=1$ and so on. If $o[X]$ is the least preferred value of $X$, then $p^o_X=~\text{Dom}(X)-1$.\\

Suppose again that we wish to answer the dominance query $N\vDash o\succ o'$ by constructing $G(o')$, this time using penalty values to prune the tree. First we define the following evaluation function
\begin{equation}
\nonumber
f(o^*)= pen(o^*)-pen(o)-\text{HD}(o^*,o),
\end{equation}
where HD is \emph{Hamming distance}, HD$(o_1,o_2)=|\{X|o_1[X]\neq o_2[X]\}|$. \citet{liAAMS2011} have shown that if there is an IFS $o'=o_1,o_2,...,o_m=o$, then $f(o_i)\geq 0$ for all $i$. Thus, when constructing $G(o')$, any improving flips with $f<0$ can be pruned.\\

This penalty-based pruning was originally presented by \citet{liAAMS2011} in combination with suffix fixing. We treat penalty pruning separately here, in order to see more clearly which pruning methods are most effective, both individually and in different combinations.\\

It is simple to combine any of the three pruning measures we are considering. Suppose we wish to answer the dominance query $N\vDash o\succ o'$, utilising the combination of a set of pruning measures, $\Gamma$. We build $G(o')$ as usual. When considering the outcome $\bar{o}$, let $F(\bar{o})$ denote the set of all improving flips of $\bar{o}$, as in \S5. As usual, we prune any elements of $F(\bar{o})$ that are already present in $G(o')$. Then, for each pruning measure $\gamma\in\Gamma$, in turn, we prune all elements remaining in $F(\bar{o})$ that satisfy the pruning condition of $\gamma$. Any improving flips that have not been pruned from $F(\bar{o})$ are added to $G(o')$ in the normal manner. We continue until $o$ is reached, that is, the dominance query is true, or the pruned $G(o')$ is complete (that is, all leaves have been considered), and thus the dominance query is false.\\

In our experiment, we evaluated the performance of each pruning measure individually, all pairwise combinations, and all three methods combined. Thus, we are comparing the performance of seven different pruning schemas. However, as mentioned previously, in order for these methods to be fully defined, we must declare how we select the next leaf for consideration when constructing $G(o')$. Different methods of leaf prioritisation have been suggested previously by \citet{boutJAIR2004} and \citet{liAAMS2011} and one could similarly propose a prioritisation heuristic based on rank values, but no analysis has been done on the effect of this choice. In our experiments, each pruning method utilised the prioritisation heuristic that was optimal for that method (given that it did not require additional calculations). Full details of the prioritisation methods we considered and those utilised, as well as the performances of each pruning method when used in conjunction with all possible prioritisation techniques are available online at \texttt{www.github.com/KathrynLaing/DQ-Pruning}.\\
	
We measured the performance of the dominance testing functions in two ways. First, we looked at \emph{outcomes traversed}, this is the number of outcomes added to the search tree before an answer to the dominance query can be determined. This is similar to the measure used by \citet{liAAMS2011} in their pruning method comparisons (where they compared penalty pruning combined with suffix fixing to suffix fixing and least variable flipping). Outcomes traversed provides us with a theoretical measure of how effective the different methods are at pruning the search tree. It reflects the number of steps the different algorithms have to go through before the queries can be answered, thus showing how efficient the different methods are in a theoretical sense. This measure has the advantage of being independent of the specific code used and the order in which pruning conditions in combinations are considered.\\

Note that it is possible for the number of outcomes traversed to be zero, that is, the dominance query may be answered without starting to construct a search tree. This can happen in three different ways for the dominance query $N\vDash o\succ o$; first, if $o=o'$, then this is trivially false. Second, if (one of) the pruning measure(s) used is penalty pruning, then, if $f(o')<0$, we can determine the dominance query to be false \citep{liAAMS2011}. Finally, if (one of) the pruning measure(s) used is rank pruning, then, if $r(o)-r(o')< L_D(o,o')$, we can determine the dominance query to be false, by Corollary 3. As these conditions are all assessed before starting to construct the search tree, they result in zero outcomes traversed. In Appendix C we look at the proportion of queries that the different functions immediately determine to be false (that is, those queries that have zero outcomes traversed). By evaluating this proportion in comparison to the total proportion of false queries, we can see how accurately these initial conditions predict the query outcome.\\

Our second measure of performance is the time elapsed (in seconds) while the function answers the query. Whilst this measure is dependent upon the exact code used, we have tried to keep the code for the different functions as uniform as possible, so that differences in performance are due to the methods rather than the code. From time elapsed we can identify which method will be the most efficient in practice. By looking at both performance measures, we can see the tradeoff between how effective a method is theoretically and the time cost due to the complexity of implementing that method. Ultimately, we will see if the theoretical benefit is worth the cost in complexity by looking at the time elapsed plots .\\

The experiments we ran to evaluate performance were as follows. For given $n$ (number of CP-net variables) and $d_U$ (maximum domain size of the variables) values, 100 CP-nets were randomly generated. Each of these CP-nets has an acyclic structure over $n$ variables, each variable has a domain size of at most $d_U$, and all parent-child relations are valid (that is, if there is an edge $X\rightarrow Y$ in the CP-net structure, it is possible to change the preference over~$Y$ by altering the value of $X$ only). For each CP-net, 10 dominance queries were randomly generated. Each of these 1000 dominance queries was answered by all seven dominance testing functions and the outcomes traversed and time elapsed was recorded. The average of these values over the 1000 queries are the values plotted for $(n,d_U)$ in the following plots.\\

This experiment was run in the binary case, $d_U=2$, for $n=3-10$. For the multivalued variable case, we allowed domain size to be up to five. We ran the experiments in this case ($d_U=5$) for $n=3-8$.\\

We have made further details of the above experiment available at the online repository \texttt{www.github.com/KathrynLaing/DQ-Pruning}. This includes the random CP-net generator code and a description of how this generator works, as well as the code for the different dominance testing functions. We have also uploaded the raw results of the experiment to this repository.

\subsection{Results}
We have four sets of data (binary CP-nets with outcome traversed and time elapsed data, and similarly for multivalued CP-nets) for each of the 7 functions. These four data sets are given in Figures 5-8. In each of Figures 5-8, Figure (a) shows the performance of the three pruning measures when used individually. To keep this plot legible, a logarithmic scale is used. Figure (b) shows the performance of all seven combinations of the three pruning measures.\\

In each figure, the $\pm$SE (standard error) interval of rank pruning is illustrated by a shaded region. The standard error intervals depict where we expect the true mean performance of the function to lie. The uncertainty represented by these intervals comes from the fact that the complexity of a dominance query, regardless of the pruning technique used, is dependent upon both the CP-net and the outcomes of interest; CP-nets with denser structures, or more convoluted preference graphs, are more likely to produce dominance queries that take longer to answer. Once a CP-net has been chosen, the position of the outcomes of interest within the preference graph further impacts how difficult the dominance query is to answer. As our CP-nets and queries were randomly generated, it is unsurprising that each function shows variation in performance. However, as all functions were tested on the same set of dominance queries, our results should accurately portray their relative performance on average.\\

In the multivalued case, the domain sizes were allowed to vary between two and five. Larger domain sizes will produce harder dominance queries in general, so we would expect this extra uncertainty to result in further variation within the results. Moreover, CP-nets with larger domain sizes have larger preference graphs, so there will also be more variation in dominance queries of the same CP-net. Hence, in the multivalued case, we expect more uncertainty in the average performance of the functions.\\

\begin{figure}
\begin{subfigure}[b]{\textwidth}
\begin{center}
\includegraphics[width=0.8\textwidth , height =10cm]{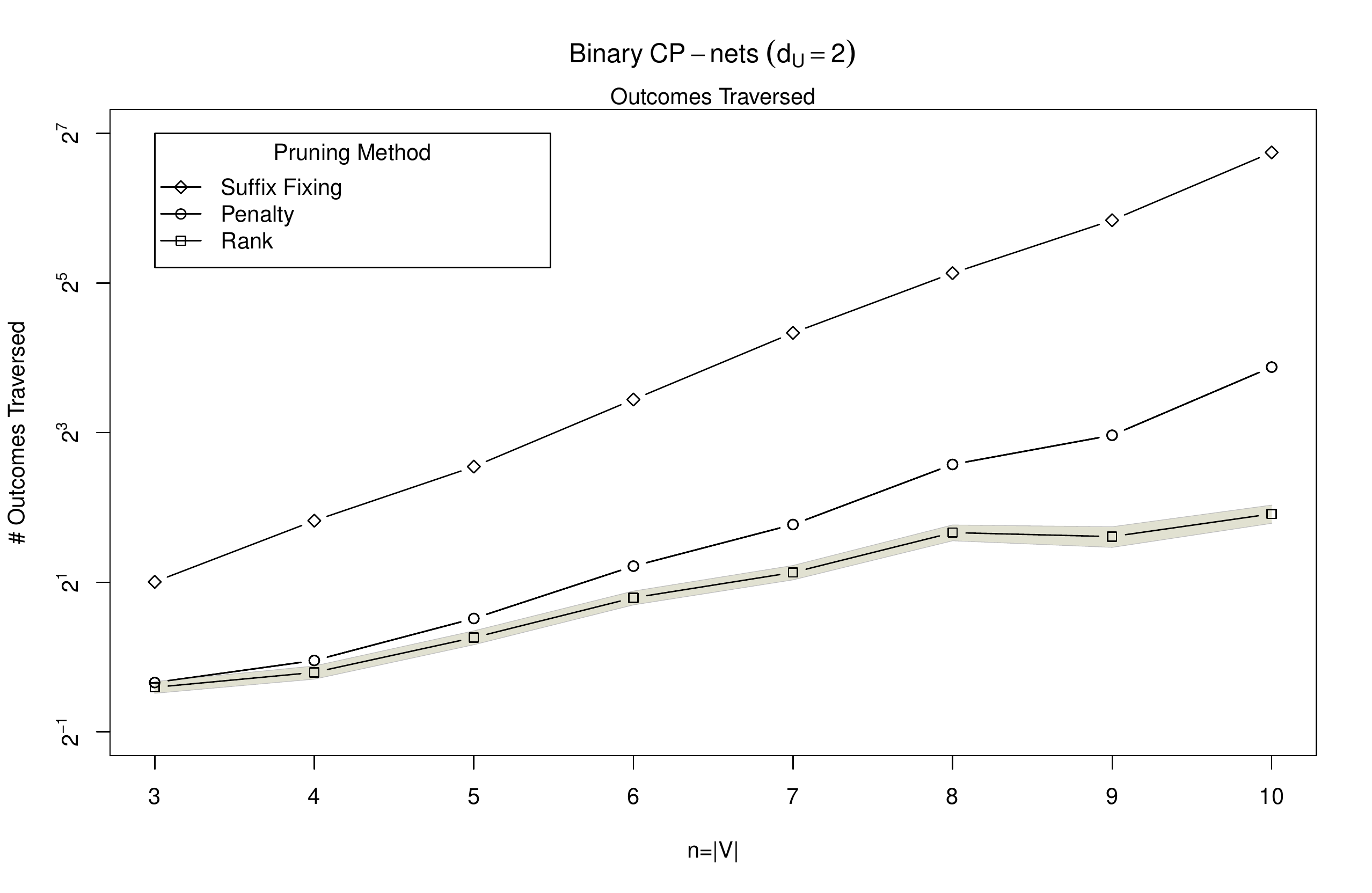}
\end{center}
\caption{Individual Pruning Measures}
\end{subfigure}

\begin{subfigure}[b]{\textwidth}
\begin{center}
\includegraphics[width=0.8\textwidth , height=10cm]{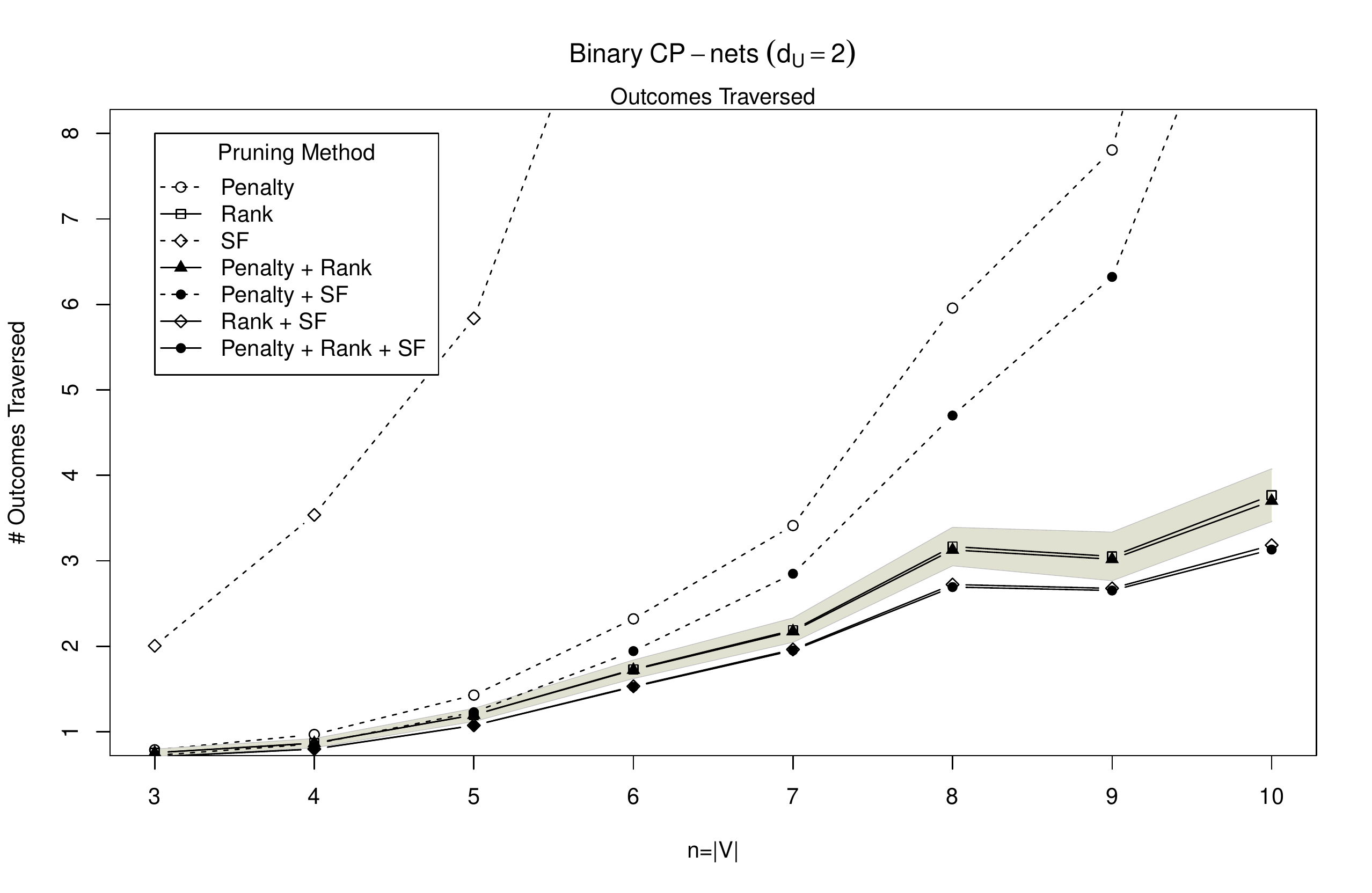}
\end{center}
\caption{All Pruning Measure Combinations}
\end{subfigure}
\caption{Binary CP-Nets - Outcomes Traversed Results}
\end{figure}

\begin{figure}
\begin{subfigure}[b]{\textwidth}
\begin{center}
\includegraphics[width=0.8\textwidth , height =10cm]{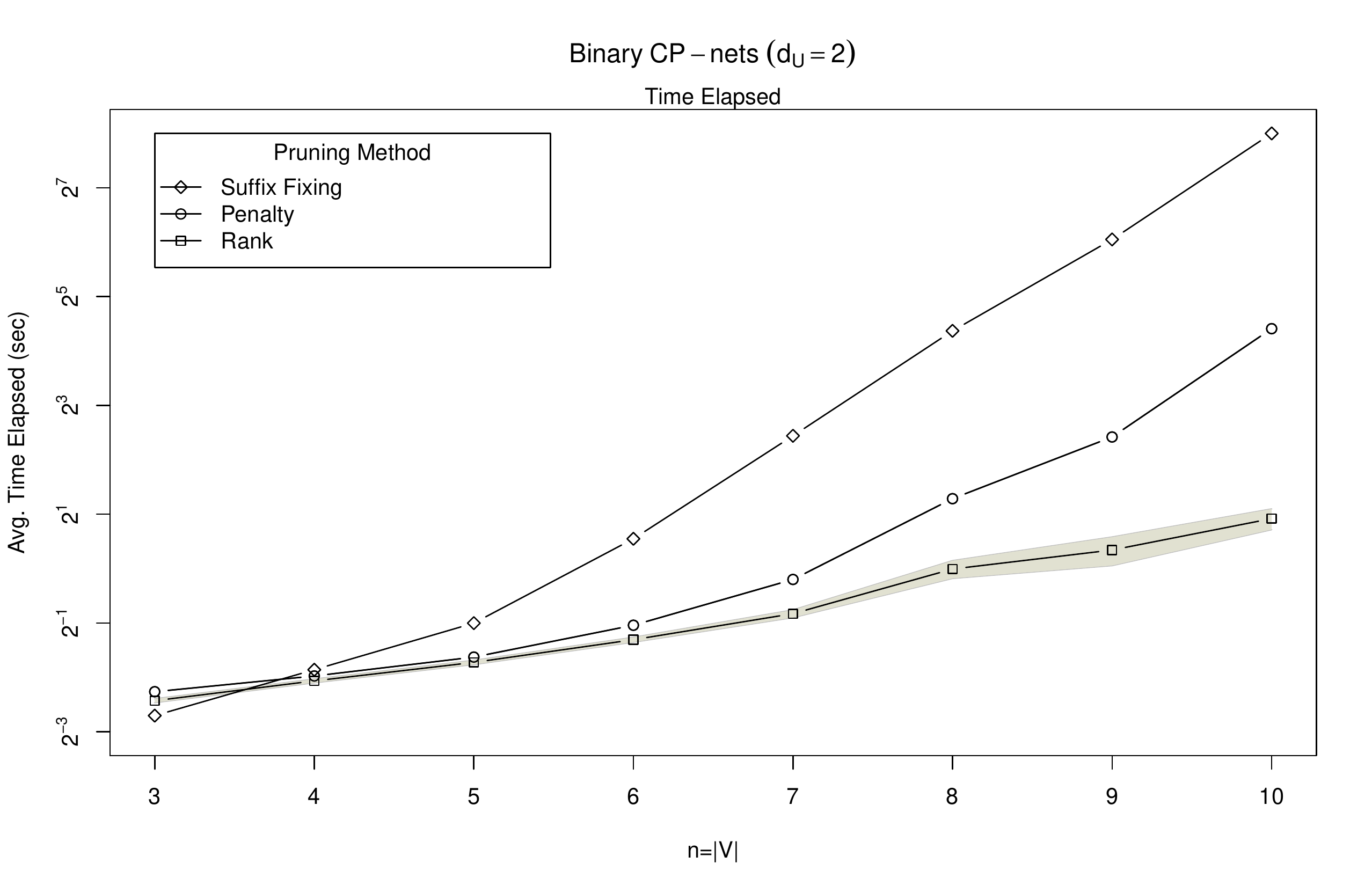}
\end{center}
\caption{Individual Pruning Measures}
\end{subfigure}

\begin{subfigure}[b]{\textwidth}
\begin{center}
\includegraphics[width=0.8\textwidth , height=10cm]{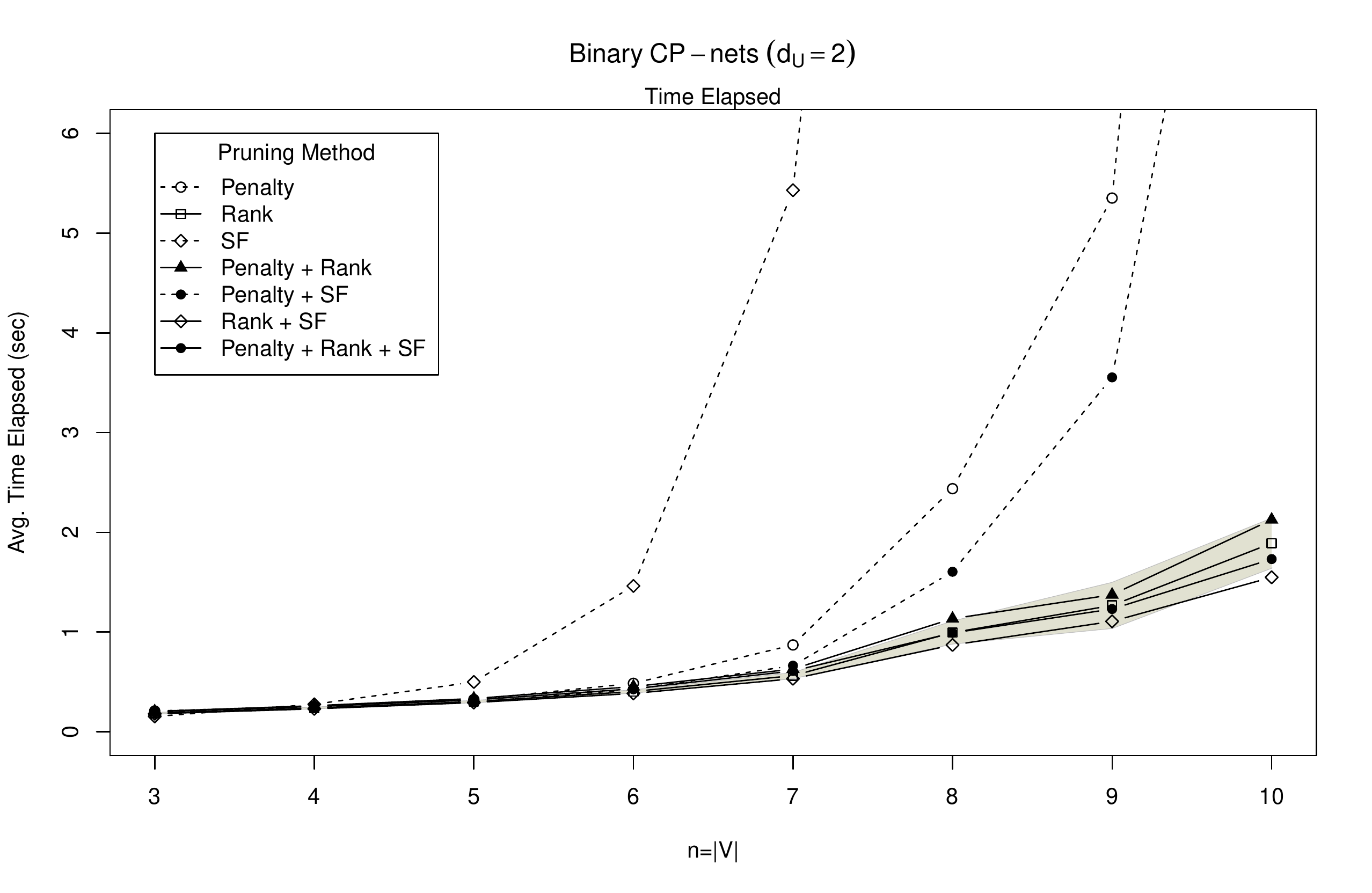}
\end{center}
\caption{All Pruning Measure Combinations}
\end{subfigure}
\caption{Binary CP-Nets - Time Elapsed Results}
\end{figure}

\begin{figure}
\begin{subfigure}[b]{\textwidth}
\begin{center}
\includegraphics[width=0.8\textwidth , height =10cm]{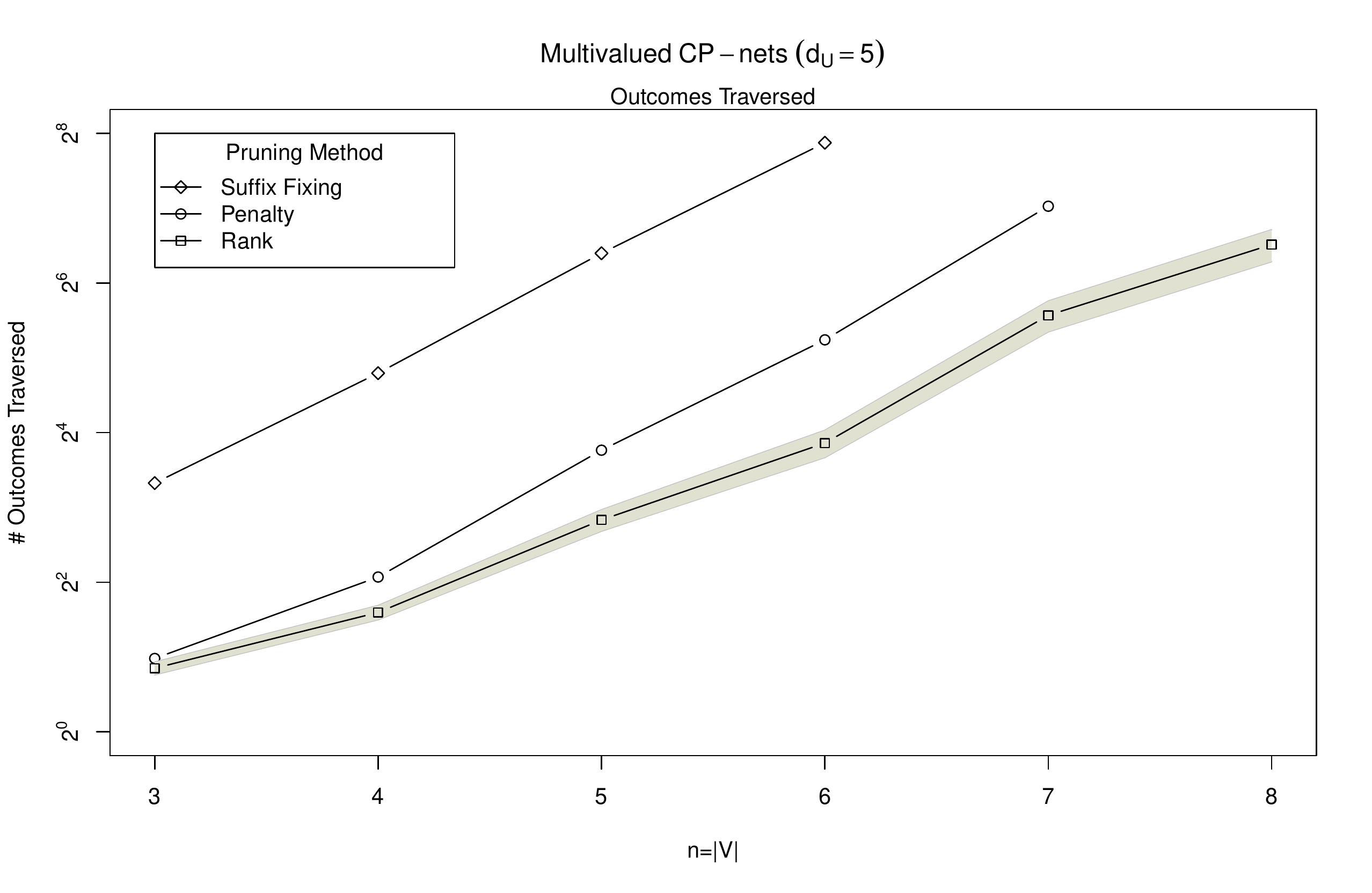}
\end{center}
\caption{Individual Pruning Measures}
\end{subfigure}

\begin{subfigure}[b]{\textwidth}
\begin{center}
\includegraphics[width=0.8\textwidth , height=10cm]{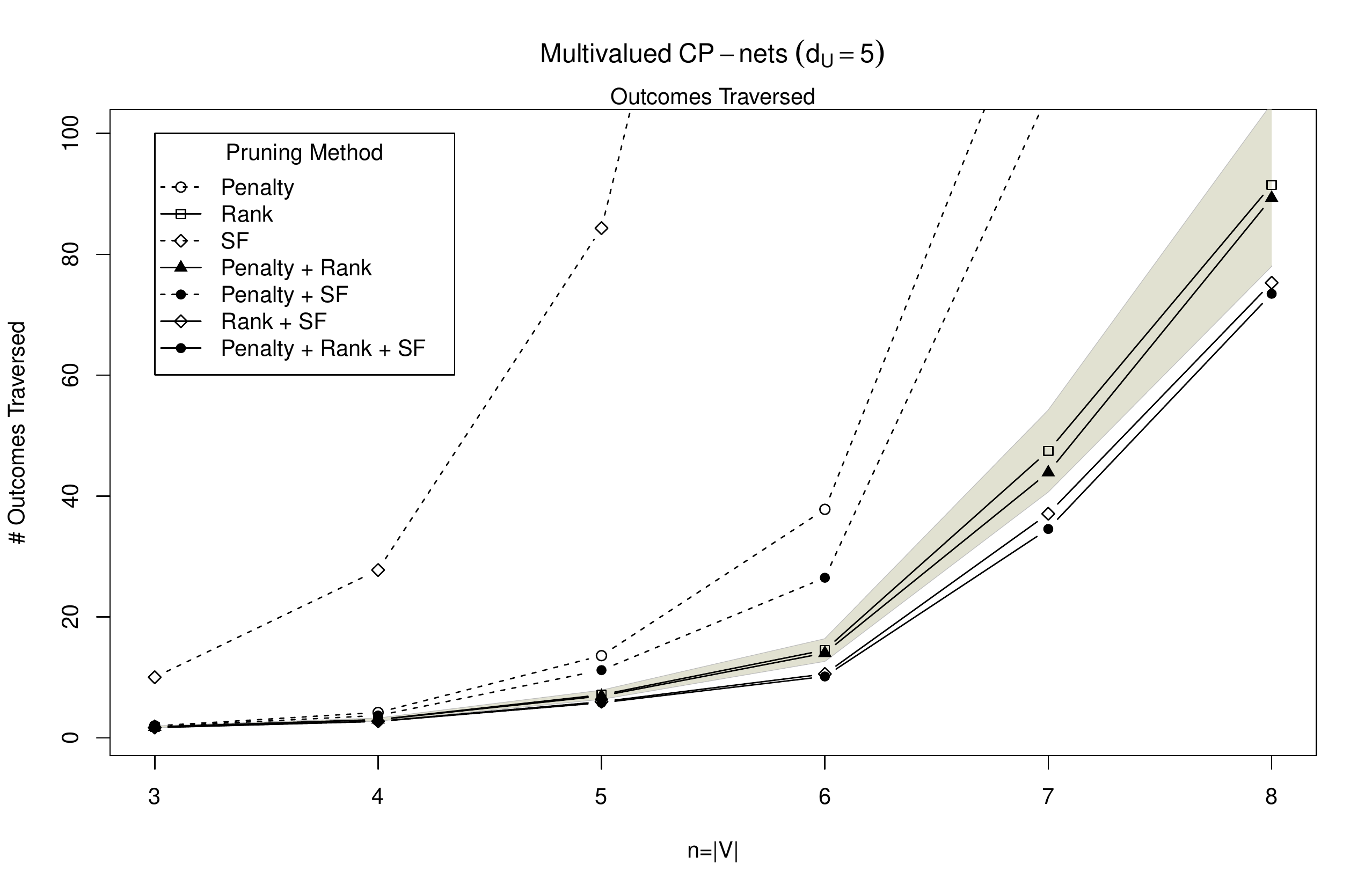}
\end{center}
\caption{All Pruning Measure Combinations}
\end{subfigure}
\caption{Multivalued CP-Nets - Outcomes Traversed Results}
\end{figure}

\begin{figure}
\begin{subfigure}[b]{\textwidth}
\begin{center}
\includegraphics[width=0.8\textwidth , height =10cm]{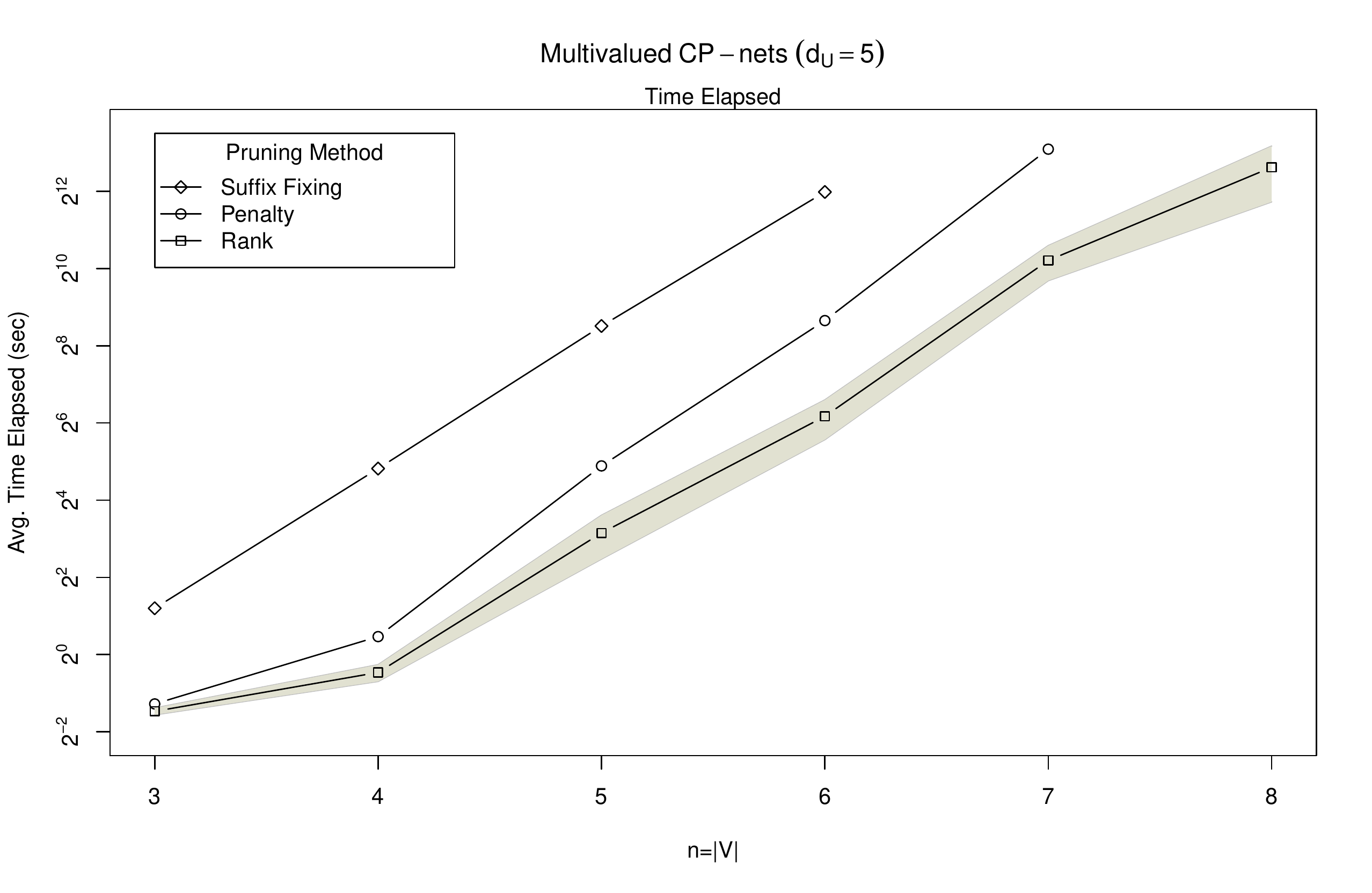}
\end{center}
\caption{Individual Pruning Measures}
\end{subfigure}

\begin{subfigure}[b]{\textwidth}
\begin{center}
\includegraphics[width=0.8\textwidth , height=10cm]{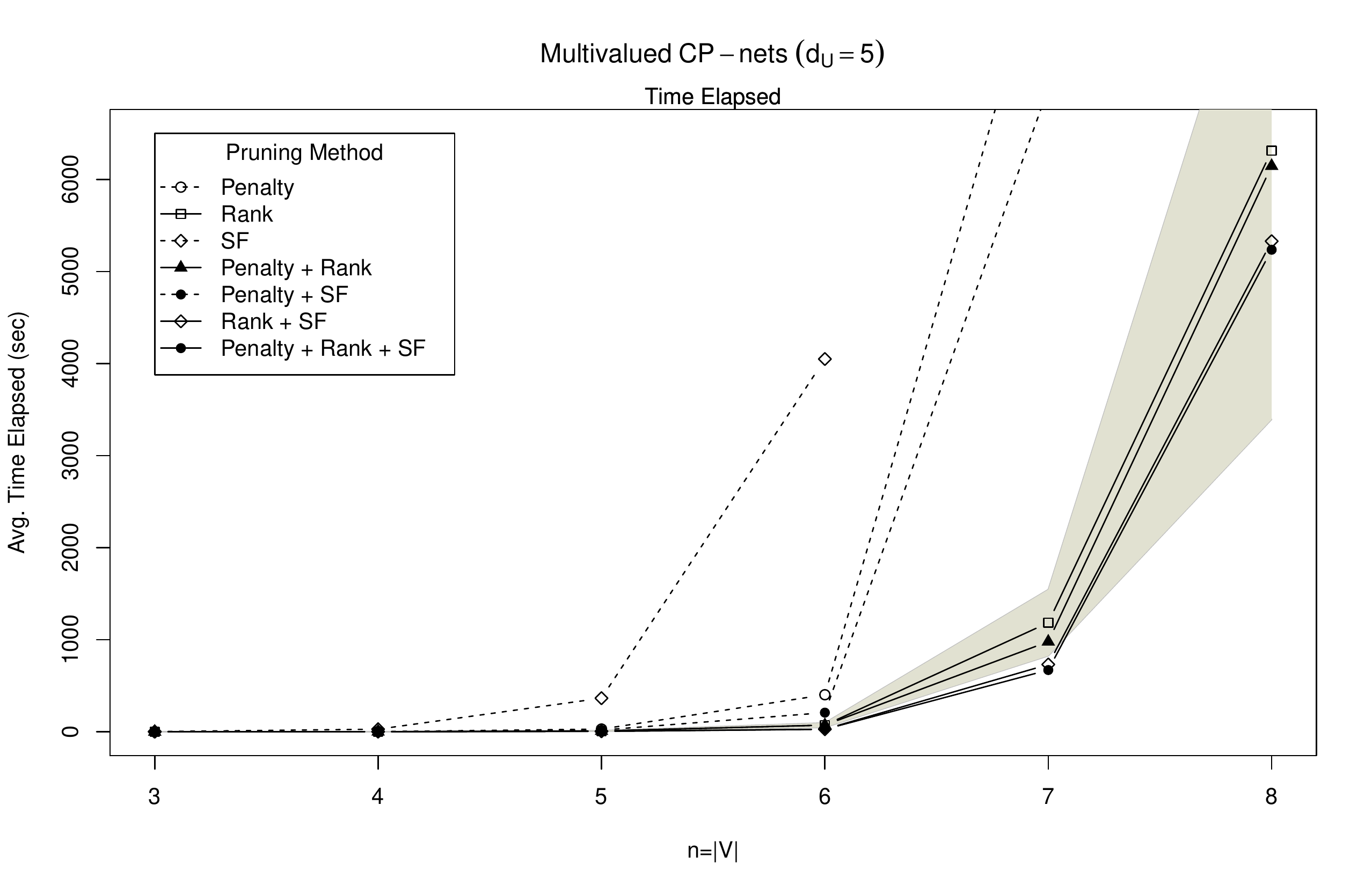}
\end{center}
\caption{All Pruning Measure Combinations}
\end{subfigure}
\caption{Multivalued CP-Nets - Time Elapsed Results}
\end{figure}

From Figures 5(b) and 7(b), we can see that adding extra pruning conditions always improves the theoretical performance of a method (that is, it results in less outcomes traversed on average). This shows that all three pruning measures are distinct, and that no pruning measure is subsumed by any other. Further, this shows us that each technique prunes branches that are not affected by either of the other two methods. It is not obvious from the way in which they are formulated that the three pruning measures are distinct in this manner. Moreover, this is not confirmed by any comparisons in existing literature. From this result, it is unsurprising that the best performing function, in the theoretical sense, is that which uses all three pruning measures.\\

However, finding the best pruning schema is not as simple as applying as many pruning conditions as possible. As we are aware that additional pruning methods come at the cost of additional complexity, we would naturally question whether these improvements are large enough to warrant the additional cost. Looking at the time elapsed results (Figures 6(b) and~8(b)), we can see that some of these `improvements' actually increase the average time taken, so the theoretical benefit is not worth the complexity cost. In particular, in the binary case, we find that a pairwise combination is actually faster than using all three pruning methods.\\

Consider the Figure (a) plots, which show the performance of the three pruning methods used individually. It is clear in all four cases that rank pruning is the most effective and most efficient method of the three by a large margin. Further, the performance of rank pruning (outcomes traversed or time elapsed) shows a much slower rate of growth than the others as the number of variables ($n$) increases, particularly in the binary variable case. Thus, if we wanted to pick a single pruning method, rank pruning is plainly the best choice.\\

Now consider the Figure (b) plots, these show the performance of all possible combinations of the different pruning methods. In all four of these figures, you can see a clear distinction in performance between the functions represented by dashed lines and those represented by solid lines. The functions represented by solid lines perform much better and show a slower rate of growth with $n$ than the functions represented by the dashed lines. These solid lines are exactly those functions that include rank pruning in their combination. Hence, we can see a clear distinction in performance between those functions that do and do not apply rank pruning. Thus, we may conclude that rank pruning is a necessary ingredient for a good pruning schema.\\

In Figure (b), the shaded area shows the standard error interval for rank pruning, the best performing of the individual pruning measures. Thus, only functions that lie below this area may be considered significantly better than using rank pruning alone. In both Figures~5(b) and 6(b), adding penalty pruning to rank pruning makes little improvement to the average number of outcomes traversed. This suggests that there are few branches pruned by penalty pruning that are not already pruned by rank pruning. Thus, it is unsurprising that adding penalty pruning results in only minor improvement in the time elapsed cases also. In fact, in the binary case (Figure 6(b)), adding penalty pruning increases the average time elapsed. This is because the additional complexity of checking the penalty condition outweighs the theoretical benefit.\\

The combination of rank pruning with suffix fixing and the combination of all three measures both perform significantly better than rank pruning alone, in terms of outcomes traversed (Figures 5(b) and 7(b)). This is probably due to less overlap in the branches pruned by rank pruning and suffix fixing. The two functions show very similar performances in terms of outcomes traversed (both in the binary and multivalued cases), though the function using all three methods does slightly better in this theoretical case, as expected. In terms of time elapsed (Figures 6(b) and 8(b)), these functions again perform better than rank pruning alone. However, they are not significantly faster than rank pruning, probably due to the associated cost of implementing the additional pruning measures. In the binary case, the combination of rank pruning and suffix fixing outperforms the combination of all three pruning methods, as the slight theoretical improvement provided by penalty pruning is not worth the associated complexity cost. In the multivalued case however, the combination of all three methods is more efficient on average, though its performance remains close to that of rank pruning and suffix fixing combined. We conjecture that, for larger values of~$n$, the combination of rank pruning and suffix fixing is likely to become more efficient than using all three methods. This is motivated by the fact that the average size of the search tree is rapidly increasing, and thus so is the number of times the penalty condition must be checked (in the case of using all three methods). Whereas the number of additional branches pruned by this check, that is, the theoretical improvement of adding penalty pruning (to rank pruning and suffix fixing combined) remains small.\\

From the above results, we have seen that our rank pruning is the most efficient of the individual methods considered. Further, from the clear distinction between functions that do and do not utilise rank pruning, we can see that rank pruning constitutes a valuable contribution to the existing methods when we allow combinations. Considering all the possible combinations of our pruning methods, the above results suggest that the most efficient combination for dominance testing in the binary case is rank pruning and suffix fixing. In the multivalued case, the most efficient method is to use all three pruning measures.

\section{CP-Nets with Indifference}
In this section, we give a more general form of our outcome rank formula that allows for indifference statements within the CP-net's CPTs. These more general ranks still reflect all entailed relations and, therefore, allow all of our previous methods and results to be applied to CP-nets that express indifference as we show below.\\

We do not assume here that the preference ordering over Dom$(X)$, given the values taken by Pa$(X)$, is a strict ordering \citep{boutJAIR2004}. For example, consider the CP-net given in Example 1, we would now permit $\text{CPT}(C)$ to express that, if it is a short flight in term time, then the user prefers to fly economy but is indifferent between first and business class. This would make the entry of $\text{CPT}(C)$ that corresponds to $AB=ab$ be $c\succ \bar{c}\sim\bar{\bar{c}}$. This kind of \emph{ceteris paribus} indifference statement is natural and likely to be commonplace when looking at real world systems \citep{alleAAC2013}. Thus, being able to deal with indifference expands the applications of our results. Further, if one were comfortable modelling unknown preferences as indifference, our results could be applied to partially specified CP-nets also.\\

\citet{boutJAIR2004} show that the presence of such indifference allows CP-nets with acyclic structures may be inconsistent (that is, have no consistent ordering). However, this can be avoided if one assumes that switching between indifferent assignments of values to parent variables should have no effect on the order of preference over any children \citep{boutJAIR2004}. Therefore, we assume here that all CP-nets with indifference statements obey this condition.\\

Recall from \S3.1 that the rank of an outcome $o$, was the sum of weights attached to each variable assignment (of $o$). These weights were constructed to approximate the utility of each variable choice in $o$. If $o[X]=x$ and $o[\text{Pa}(X)]=u$, then the weight attached to the assignment of $X$ is:
\begin{equation}
\nonumber
AF_X (d_X+1) P_P\{X=x\hspace{0.1cm}|\hspace{0.1cm}\text{Pa}(X)=u\}
\end{equation}

The justification for the presence of each of these factors remains valid for CP-nets with indifference statements. Thus, we do not need to create a new weighting convention, we simply need to generalize this formula so that it is defined in the cases of indifference. The~$AF_{X_i}$ and $d_{X_i}$ terms depend only on the CP-net structure, not on the CPTs, and thus can remain as they were defined previously. The $P_P\{X=x\hspace{0.1cm}|\hspace{0.1cm}\text{Pa}(X)=u\}$ factor, as defined in \S3.1, needs to be redefined to allow for the possibility of indifference statements.\\

Recall that $P_P\{X=x\hspace{0.1cm}|\hspace{0.1cm}\text{Pa}(X)=u\}$ is a factor on the (0,1] scale indicating to what degree the user prefers this choice of value for $X$ (given $\text{Pa}(X)=u$). We redefine\break$P_P\{X=x\hspace{0.1cm}|\hspace{0.1cm}\text{Pa}(X)=u\}$ more generally as follows. Suppose the row of $\text{CPT}(X)$ that corresponds to $\text{Pa}(X)=u$ has $\ell$ indifferences, then 
\begin{equation}
\nonumber
P_P\{X=x\hspace{0.1cm}|\hspace{0.1cm}\text{Pa}(X)=u\}=\frac{(n_{X}-\ell)-k+1}{n_{X}-\ell}.
\end{equation}
Where $k$ is the position of preference of the choice of $X=x$ given $\text{Pa}(X)=u$. Note that we consider all values of $X$ to which the user is pairwise indifferent to be in the same preference position, that is, there are $n_X-\ell$ possible positions of preference ($1,2,...,n_X-\ell$). Here, $k=1$ if $x$ is (one of) the most preferred value(s) $X$ can take, $k=2$ if $x$ is (one of) the value(s) of $X$ in the $2^{nd}$ most preferred position, and so on.\\

\textbf{Example 7.} Let $N$ be a CP-net over variables $V$. Let $X\in V$ be some variable with the following partial CPT:
\begin{center}
\begin{tabular}{|c|c|}
\hline
$\text{Pa}(X)=u$ & $x_1\succ x_2\sim x_3 \sim x_4 \succ x_5 \succ x_6 \sim x_7\succ x_8$\\
\hline
\end{tabular}
\end{center}
Then, using the generalised $P_P$ definition, we have the following $P_P$ values.
\begin{equation}
\nonumber
\begin{split}
P_P\{X=x_1\hspace{0.1cm}|\hspace{0.1cm}\text{Pa}(X)=u\}= \frac{(8-3)-1+1}{8-3}=\frac{5}{5}, \hspace{0.1cm} & P_P\{X=x_2\hspace{0.1cm}|\hspace{0.1cm}\text{Pa}(X)=u\}= \frac{(8-3)-2+1}{8-3}=\frac{4}{5}, \\
\end{split}
\end{equation}
\vspace{-0.8cm}
\begin{align*}
P_P\{X=x_3\hspace{0.1cm}|\hspace{0.1cm}\text{Pa}(X)=u\}=\frac{4}{5}, \hspace{0.1cm} & P_P\{X=x_4\hspace{0.1cm}|\hspace{0.1cm}\text{Pa}(X)=u\}=\frac{4}{5}, \hspace{0.1cm} & P_P\{X=x_5\hspace{0.1cm}|\hspace{0.1cm}\text{Pa}(X)=u\}=\frac{3}{5},\\
P_P\{X=x_6\hspace{0.1cm}|\hspace{0.1cm}\text{Pa}(X)=u\}=\frac{2}{5}, \hspace{0.1cm} & P_P\{X=x_7\hspace{0.1cm}|\hspace{0.1cm}\text{Pa}(X)=u\}=\frac{2}{5}, \hspace{0.1cm} & P_P\{X=x_8\hspace{0.1cm}|\hspace{0.1cm}\text{Pa}(X)=u\}=\frac{1}{5}. 
\end{align*}

Notice that this generalised definition of $P_P\{X=x\hspace{0.1cm}|\hspace{0.1cm}\text{Pa}(X)=u\}$ is a value in\break$\{ 1/(n_{X_i}-\ell), 2/(n_{X_i}-\ell),...,(n_{X_i}-\ell-1)/(n_{X_i}-\ell), 1\}$. Further, this can still be interpreted as a factor on the (0,1] scale indicating to what degree the user prefers this choice of value for $X$ (given $\text{Pa}(X)=u$).\\

Now that all of the terms in our previous weight formula are defined in the case of $N$ having indifference statements, we can define outcome ranks for CP-nets with indifference.\\

\textbf{Definition 7: (Generalised) Outcome Rank.} Let $N$ be a CP-net over variables $V$, which may have indifference statements in its CPTs. Let $o$ be an associated outcome. Then, the \emph{(generalised) rank} of $o$, $r_G(o)$, is defined as 
\begin{equation}
\nonumber
r_G(o)=\sum_{X\in V} AF_X(d_X+1)P_P\{X=o[X]\hspace{0.1cm}|\hspace{0.1cm}\text{Pa}(X)=o[\text{Pa}(X)]\},
\end{equation}
where the $P_P$ term uses the more general form given above.\\

For the special case in which there are zero indifference statements, the general form of~$P_P$ clearly simplifies to the original definition, given in \S3.1. Thus, the generalised outcome ranks in this special case, simplify to the outcome ranks given by Definition 4 (for multivalued CP-nets in which we assumed no indifference statements).\\

\textbf{Remark 5.} We have could have used an event tree representation to define generalised outcome ranks by generalising the notion of event tree representation to include indifference. This generalised $T(N)$ would have the same structure as before, but use the $k$ values from the above definition of $P_P$ in order to label the branches. For example, let $N$ be the CP-net in Example 7. At the point where $T(N)$ branches into the possible values of $X$, if $\text{Pa}(X)$ were previously assigned the values in $u$, then these $X$ branches are labelled as follows. The branch corresponding to $x_1$ would be labelled `$1^{st}$'. The $x_2$, $x_3$, and $x_4$ branches would all be labelled `$2^{nd}$'. The $x_5$ branch would be labelled `$3^{rd}$' and so on. We again have that $N$ and $T(N)$ are equivalent by an argument almost identical to that given in Remark 2. The weighted event tree $W(N)$, would be defined in the same way as in \S3.1, now using the new definition of $P_P$, and the generalised outcome rank would be defined analogously to rank (Definition 4). Further, $W(N)$ would also be equivalent to $N$, by similar reasoning to that given in \S3.1.\\ 

All of our applications of the outcome ranks defined in \S3.1 rely solely on the fact that they reflect all entailed relations (Theorem 1). Naturally, we want this property to hold for our generalised outcome ranks and the following theorem shows that it does.\\

\textbf{Theorem 2.} Let $N$ be a CP-net over a set of variables $V$, which may have indifference statements in its CPTs. Let $o,o'$ be associated outcomes. Then,
\begin{equation}
\nonumber
N\vDash o\succ o' \implies r_G(o)>r_G(o') \text{ and }
\end{equation}
\begin{equation}
\nonumber
N\vDash o\sim o' \implies r_G(o)=r_G(o').
\end{equation}

Proof in Appendix A.\\

Thus, the (not necessarily strict) ordering of the outcomes, $\succsim^*$, induced by the ranks $r_G$, is again a consistent ordering. That is, $N\vDash o\succ o' \implies o\succ^* o'$ and $N\vDash o\sim o' \implies o\sim^* o'$. Thus, using the generalised outcome ranks, we can obtain a (not necessarily strict) consistent ordering for any $N$, which may have indifference statements, using exactly the same method as given in \S3.2. Similarly, we can obtain a (not necessarily strict) consistent ordering for any subset of the outcomes or for a CP-net with additional plausibility constraints using the methods given in \S3.2 (ignoring any instruction to arbitrarily order outcomes with equal ranks), now using the generalised ranks $r_G$, given by Definition 7. These orderings can be shown to be consistent in the same way as the corresponding orderings in \S3.2. \citet{boutJAIR2004} claim that their methods for obtaining a consistent ordering of (any subset of) the outcomes also apply to CP-nets with indifference. However, the complexity of ordering queries in this case is unknown (though they conjecture that it is hard) and, therefore, so is the complexity of their method for consistently ordering a subset of the outcomes. In contrast, if one uses our method, the complexity of consistently ordering any subset of the outcomes of size $k$ in the case of indifference is still $O(|V|^4k +k^2)$. This is a result of the fact that we can compute $r_G(o)$ in the same time complexity as $r(o)$ as we show below.\\

In all of the above applications of $r_G$, we have obtained a consistent ordering (of $N$, some subset of the outcomes, or some constrained CP-net $N_C$), $\succsim^*$, which is not necessarily strict. That is, for any entailed relation $o\succ o'$ (or $o\sim o'$) we have $o\succ^* o'$ (or $o\sim^* o'$). The presence of indifference might mean that we do not mind a non-strict ordering, however $o\sim^*o' \nRightarrow$~$o$~and $o'$ are indifferent. Further, it is not trivial to distinguish between when $o\sim^* o'$ is caused by $o$ and $o'$ indifferent and when this is due to $o$ and $o'$ incomparable. Thus, there is no trivial way of obtaining a consistent ordering from $\succsim^*$ that has equality \emph{only} in the case of indifference. If a strict consistent ordering is required (so we are not interested in preserving indifference), then we can obtain a strict ordering of the outcomes, $\succ^*$, from~$\succsim^*$ simply by forcing outcomes of equal rank into an arbitrary order. This strict ordering retains the property that for any entailed relation, $o\succ o'$, we have $o\succ^* o'$ by Theorem 2. Thus, we can obtain a strict ordering that is consistent with all entailed preferences (but not indifferences).\\

Algorithms 1, 2, and 3 can be used to calculate $r_G(o)$ exactly as described for $r(o)$ in~\S4 (with the same time complexity) if we make two small adjustments. First, line \textbf{10} of Algorithm 1 should use $|\text{Dom}(X_i)|- \ell$ in place of $|\text{Dom}(X_i)|$, where $\ell=\#$ indifferences in the $\text{Pa}(X_i)=o[\text{Pa}(X_i)]$ entry of $\text{CPT}(X_i)$. Second, in the case of indifference statements, the preference positions in the input CPTs must be as defined in our definition of the more general form of $P_P$ (these are the $k$ terms). Thus, we can compute $r_G(o)$ in the same time as $r(o)$ and so all complexity results transfer directly to CP-nets with indifferences in their CPTs.\\

Suppose $N$ is a CP-net, which may have indifferences in its CPTs, and let $o$ and $o'$ be associated outcomes. The dominance query $N\vDash o\succ o'$ can be answered using a method very similar to the one described in \S5. First, note that $N\vDash o\succ o'$ if and only if there is an improving flipping sequence $o'\leadsto o$ \citep{boutJAIR2004}. As there may be indifferences we must clarify what we mean by IFS. An IFS is a sequence $o'=o_1,o_2,...,o_m=o$ such that, for all $i$, $o_i$ and $o_{i+1}$ differ on the value taken by exactly one variable and either $N\vDash o_{i+1}\succ o_i$ or  $N\vDash o_{i+1}\sim o_i$ holds; further, for at least one $j$ we have $N\vDash o_{j+1}\succ o_j$. Returning to our dominance query, if $r_G(o')\geq r_G(o)$, then the dominance query is false by Theorem 2. Otherwise, starting from $o'$, we build up the tree as in \S5, only now an outcome branches into all improving flips and all indifferent flips. Only outcomes that are not already in the tree may be added. A branch to outcome $o^*$ is pruned (not explored further) if $r_G(o^*)>r_G(o)$ as, if $o^*$ is on a $o'\leadsto o$ IFS, we must have $N\vDash o\succ o^*$ or $N\vDash o\sim o^*$ and so, by Theorem~2, $r_G(o^*)\leq r_G(o)$. As in \S5, this pruning will improve the efficiency of answering dominance queries and in finitely many steps we will either reach $o$ (dominance query is true) or there will be no more valid branches to explore (dominance query is false). \citet{boutJAIR2004} claim that their pruning methods for dominance queries also transfer to CP-nets with indifference. Additionally, \citet{alleAAC2013} looked at answering dominance queries for CP-nets with indifference utilising a SAT solver. In contrast to our work, he considers `weak dominance', that is, asking whether $N\vDash o\succsim o'$ holds.\\

Note that we can answer indifference queries, $N\vDash o\sim o'$? (if $r_G(o)=r_G(o')$) using the same search technique as above, only now the outcomes branch \emph{only} into indifference flips. However, we cannot use ranks to prune this search as all the outcomes in the tree will have the same rank (the same as that of $o$ and $o'$).\\

Note that least rank improvement ($L(X)$) terms are defined the same way in the case where a CP-net has indifferences. Thus, Lemma 2 still holds in the case where $N$ has indifferences in its CPTs. Further, if $N$ may have indifference statements in its CPTs, we have the following analogous result to Corollary 3. \\

\newpage\textbf{Corollary 4.} Let $N$ be a CP-net over variables $V$, which may have indifference statements within its CPTs. Let $o_1$ and $o_2$ be associated outcomes and $D=\{X\in V\hspace{0.1cm}|\hspace{0.1cm}o_1[X]\neq~o_2[X]\}$. Then,
\begin{equation}
\nonumber
N\vDash o_1\succ o_2 \implies r_G(o_1)-r_G(o_2)\geq min_{X\in D}\{L(X)\} > 0.
\end{equation}

The proof of this result is very similar to that of Corollary 3. One just needs to note that an IFS $o_2\leadsto o_1$ must have at least one improving (not indifferent) flip of at least one $X\in D$. Further, an improving flip of $X\in V$ corresponds to a rank increase of at least $L(X)$ (as we showed for Corollary 3 only now we must allow $N$ to have indifference).\\

\textbf{Definition 8: Minimum (Entailed) Rank Difference.} Let $N$ be a CP-net over variables $V$, which may express indifference. Let $o$ and $o'$ be associated outcomes and let $D=\{X\in V\hspace{0.1cm}|\hspace{0.1cm}o[X]\neq o'[X]\}$. The \emph{minimum (entailed) rank difference} of $o$ and $o'$, denoted~$M_D(o,o')$, is defined to be
\begin{equation}
\nonumber
M_D(o,o')=min_{X\in D}\{L(X)\}.
\end{equation}

By Corollary 4, these terms can be used to prune dominance queries more effectively. Suppose we are answering the dominance query $N\vDash o\succ o'$, such that $r_G(o)\geq r_G(o')+ M_D(o,o')$. Then, starting at $o'$, we build up the search tree as described above. Any branch to an outcome $o^*$ such that $r_G(o^*)>r_G(o)$ can be pruned, as before. Further, we may prune any branch to an outcome $o^*$ such that $r_G(o^*)<r_G(o)$  and $r_G(o^*)+M_D(o^*,o)>r_G(o)$. This is because, if $o^*$ is on an IFS $o'\leadsto o$, then either $N\vDash o\succ o^*$ or $N\vDash o\sim o^*$, but as $r_G(o^*)\neq r_G(o)$ we can't have $N\vDash o\sim o^*$ (by Theorem 2). However, we can't have $N\vDash o\succ o^*$ by Corollary~4, as $r_G(o^*)+M_D(o^*,o)>r_G(o)$. Thus, $o^*$ is not on an IFS $o'\leadsto o$ and so we do not need to explore the branch further and can prune it from the search. As we have an additional pruning condition, this will be more efficient than the above method of answering dominance queries.\\

As with outcome ranks for CP-nets with no indifference, we have also found that reducing a CP-net to these generalised ranks loses no information. That is, from the generalised outcome ranks alone we could reconstruct the original CP-net (which may have indifference statements). We can also answer ordering and dominance queries directly from these generalised outcome ranks, without consulting the CP-net. Further, given the (not necessarily strict) consistent ordering induced from these generalised outcome ranks, $\succsim^*$ , we can determine whether $\alpha\succ^*\beta$ is entailed ($N\vDash \alpha\succ \beta$) or constructed ($\alpha$ and $\beta$  are incomparable, we don't know which is preferred by the user) directly from $\succsim^*$. We can also determine whether $\alpha\sim^*\beta$ is entailed ($N\vDash \alpha\sim \beta$) or constructed, directly from $\succsim^*$. Furthermore, given $\succsim^*$ (or any, not necessarily strict, consistent ordering) we can update this ordering given new (consistent) preference or indifference information without needing to consult the CP-net. However, the details of these claims are not the focus of this paper.\\

In this section, we have shown how our rank definition can be generalised to allow for indifference. Further, we have demonstrated that all of our results now apply to CP-nets with indifference. In particular, we can obtain consistent orderings and improve the efficiency of dominance queries in almost exactly the same way as for CP-nets without indifference. 

\section{Discussion}
In this paper, we have introduced a novel method of quantifying a user's preference over outcomes, given a CP-net representation of their preferences. We have shown these outcome ranks to be an accurate representation of user preference because all entailed (known) user preferences are reflected in the rank values. Thus, our ranks naturally induce a consistent ordering of (any subset of) the outcomes. We have also shown that this is sufficient to find a consistent ordering of any CP-net with additional plausibility constraints. We have also presented algorithms for calculating these outcome ranks in $O(n^4)$ time, where $n$ is the number of variables of interest. Further, we have shown how these outcome ranks can be used to improve the efficiency of answering dominance queries by pruning the search tree. Through experimental comparisons, we have shown that this is more efficient than the existing methods of pruning the search tree. These experiments also evaluated the performance of combinations of pruning methods and found rank pruning to be a crucial component of any effective pruning schema. Finally, we have generalised our outcome rank definition to allow for indifference statements within the CPTs. We have shown that these generalised ranks remain an accurate representation of preference as all entailed preference and indifference is reflected in the rank values. From this result, we have shown that all of our previous results apply also to CP-nets with indifference.\\

As we have mentioned previously, our rank values are more powerful than we have illustrated here. They contain exactly the same information as the original CP-net. Also, we can determine whether $r(o)>r(o')$ is entailed ($N\vDash o\succ o'$) or constructed ($o,o'$ incomparable) without consulting the CP-net. Further, we can iteratively update the induced consistent ordering as we gain new (consistent) information on user preference. These results are explored in a forthcoming paper. In fact, we shall show that these claims hold for any consistent ordering: they are not specific to the consistent ordering induced by our outcome ranks.\\

A key application of our outcome ranks is improving the efficiency of dominance testing. A future direction of our work is to address this problem from a different angle. We intend to improve dominance testing efficiency further by creating a method of preprocessing the CP-net. This process will remove variables that are irrelevant to the dominance query and thus produce a smaller CP-net for which we now answer an easier dominance query. \citet{boutJAIR2004} have introduced a method of preprocessing the CP-net called forward pruning, this removes variable domain values rather than variables themselves. We shall compare the performance of our preprocessing to that of forward pruning, as well as consider how the two methods might combine to form a procedure that is more effective that the sum of its parts.\\

Our outcome ranks reflect all entailed preferences in their relative magnitudes; however, the values themselves have not yet been shown to have a meaningful interpretation. For example, the difference between two ranks is not guaranteed to be an accurate measure of the difference in user preference between the two outcomes. These ranks are sufficient for our current purposes, but we hope to work towards assigning a more meaningful value to the outcomes in our future work. Ideally, this value would approximate the user's utility function (beyond simple pairwise comparisons); however, it seems likely that this would require more information than a basic CP-net provides in most cases. For example, work done by \citet{boutUAI2001} on constructing a global utility function utilises a UCP-net, which has marginal utilities in addition to the basic CP-net.\\

We have mentioned above that we are able to update consistent orderings given new, consistent information on the user's preferences. However, due to natural human inconsistency, we are likely to receive inconsistent (but valid) additional preference information. We would like to look at methods for updating the consistent ordering or CP-net in this case. The correct way to deal with new, inconsistent, information is likely to be very context dependent, and any procedure must be easily adaptable to suit different applications.\\

In the future, it would be interesting to investigate how our work relates to the notion of transparent entailment introduced by \citet{dimoIJCAI2009}. Transparent entailment of a (set of) preferences is a sufficient condition for entailment that is simple to check.

\section*{Acknowledgements}
The work of K. Laing was supported by a University of Leeds Research Scholarship. Part of this work was undertaken on ARC3, part of the High Performance Computing facilities at the University of Leeds, UK.

\appendix
\section*{Appendix A. Proofs}

\noindent\textbf{Proof of Theorem 1.} Before we commence the proof, recall the following. The edge of~$W(N)$ that indicates that $X=x$, given $\text{Pa}(X)=u$ previously, has the following weight. 
\begin{equation}
\label{TreeWeightAbbrev}
AF_X (d_{X}+1) P_P\{X=x\hspace{0.1cm}|\hspace{0.1cm}\text{Pa}(X)=u\}
\end{equation}
Full explanation of the notation is given in \S3.1.\\

If $N\vDash o\succ o'$, then there exists an improving flipping sequence of outcomes $o'=o_1, o_2,..., o_m=o$, such that~$o_{i+1}$ differs from $o_{i}$ on the value taken by exactly one variable and $N\vDash o_{i}\prec o_{i+1}$ \citep{boutJAIR2004}. Thus, proving the theorem for $o$ and~$o'$ that differ on the value of exactly one variable is sufficient, as the more general result follows by transitivity.\\

Suppose $o$ and $o'$ differ only on the value taken by $X$. Let $x$ and $x'$ be the values assigned to $X$ in $o$ and $o'$ respectively (that is, $o[X]=x$ and $o'[X]=x'$). Let $u$ be the set of values assigned to $\text{Pa}(X)$ in both outcomes ($u=o[\text{Pa}(X)]=o'[\text{Pa}(X)]$).\\

Let $x_1\succ x_2\succ\cdots\succ x_m$ be the preference ordering of $\text{Dom}(X)$ given that $\text{Pa}(X)=u$. This is the row of $\text{CPT}(X)$ that corresponds to $\text{Pa}(X)=u$. Suppose $x=x_i$ and $x'=x_j$, we know that $i<j$ as $o'\rightarrow o$ is an improving flip of $X$.\\

Let $o_k$ denote the outcome that has $o_k[X]=x_k$ and for all variables $Y\neq X$ has $o_k[Y]=o[Y](=o'[Y])$.  Then the sequence of outcomes $o_1, o_2,...,o_m$, is a sequence of flips of $X$ through the values $x_1,x_2,...,x_m$. As $\text{Pa}(X)=u$ in each $o_k$, these are improving flips of $X$ so $N\vDash o_1,\succ o_2\succ\cdots\succ o_m$. Notice that $o=o_i$ and $o'=o_j$ with $i<j$ so we have $N\vDash o=o_i\succ o_{i+1}\succ\cdots\succ o_j=o'$. Hence, it is sufficient to prove $r(o)>r(o')$ for the specific case where $x$ and $x'$ are adjacent in the ordering $x_1\succ x_2\succ\cdots\succ x_m$, that is, $j=i+1$. The more general case, where $x$ and $x'$ are not adjacent ($j>i+1$) follows by the fact that $>$ is transitive. \\

To see this explicitly, suppose we have proven the case where $x$ and $x'$ are adjacent\break($j=i+~1$). If we then have the non-adjacent case ($j>i+1$), then we have\break$N\vDash~o=~o_i\succ o_{i+1}\succ\cdots\succ o_j=o'$. From the adjacent case we get that $r(o_i)>r(o_{i+1})$, $r(o_{i+1})>r(o_{i+2})$,...,$r(o_{j-1})>r(o_{j})$. Thus, by the transitivity of $>$, we have $r(o_i)>r(o_j)$, that is, $r(o)>r(o')$. It is therefore sufficient to prove the theorem in the case where $o'\rightarrow o$ is an improving flip of $X$ between adjacent values of $X$ in the ordering $x_1\succ x_2\succ\cdots\succ x_m$. Thus, we can assume that $x$ and $x'$ are adjacent values, that is, $x$ and $x'$ are the~$i^{th}$ and $(i+1)^{th}$ most preferred values of $X$, given $\text{Pa}(X)=u$, for some $i$.\\

We now demonstrate that $r(o)>r(o')$ under the above assumptions. Let $p$ and $p'$ be the root-to-leaf paths of $W(N)$ that correspond to $o$ and $o'$. Recall that $r(o)$ is the sum of the edge weights of $p$. Similarly $r(o')$ is the sum of the edge weights of $p'$. Thus, to evaluate these ranks, we must first determine what these edge weights are.\\

Let $Y\in V$ be a variable such that $Y\neq X$ and $X\not\in \text{Pa}(Y)$. As $o$ and $o'$ differ only on the value of $X$, $Y$ and $\text{Pa}(Y)$ must take the same values in both $o$ and $o'$. Let $y=o[Y]=o'[Y]$ ($y\in \text{Dom}(Y)$) and $w=o[\text{Pa}(Y)]=o'[\text{Pa}(Y)]$ ($w\in \text{Dom}(\text{Pa}(Y))$). From Expression \eqref{TreeWeightAbbrev}, the weight assigned to the edge of $p$ indicating $Y$ takes the value $y$ is $AF_Y(d_Y+1)P_P\{Y=y\hspace{0.1cm}|\hspace{0.1cm}\text{Pa}(Y)=w\}$. The weight assigned by Expression \eqref{TreeWeightAbbrev} to the edge of~$p'$ indicating $Y$ takes the value $y$ is identical. Thus, any such variable (any variable which is neither $X$ itself, nor a child of $X$) contributes exactly the same quantity to both sums,~$r(o)$ and $r(o')$. Let $\alpha$ denote the total contribution to $r(o)$ (and thus $r(o')$ also) by such variables.\\

The weight on the edge of $p$ indicating $X$ takes the value $x$ is\break$AF_X(d_X+1)P_P\{X=x\hspace{0.1cm}|\hspace{0.1cm}\text{Pa}(X)=u\}$. As we have assumed $x$ to be the $i^{th}$ most preferred value of $X$, given $\text{Pa}(X)=u$, $P_P\{X=x\hspace{0.1cm}|\hspace{0.1cm}\text{Pa}(X)=u\}=\frac{n_X-i+1}{n_X}$.\\

The weight on the edge of $p'$ indicating $X$ takes the value $x'$ is\break$AF_X(d_X+1) P_P\{X=x'\hspace{0.1cm}|\hspace{0.1cm}\text{Pa}(X)=u\}$. As we have assumed $x'$ to be the $(i+1)^{th}$ most preferred value of $X$, given $\text{Pa}(X)=u$, $P_P\{X=x'\hspace{0.1cm}|\hspace{0.1cm}\text{Pa}(X)=u\}=\frac{n_X-(i+1)+1}{n_X}$.\\

Let Ch$(X)=\{Y_1,...,Y_\ell\}$ be the set of variables that have $X$ as one of their parent variables in the structure of $N$, these are the \emph{children} of $X$. Let $y_j=o[Y_j]=o'[Y_j]$. Let $v_j=o[\text{Pa}(Y_j)]$ and $v'_j=o'[\text{Pa}(Y_j)]$. The weight on the edge of $p$ which indicates $Y_j=y_j$ is $AF_{Y_j}(d_{Y_j}+1)P_P\{Y_j=y_j\hspace{0.1cm}|\hspace{0.1cm}\text{Pa}(Y_j)=v_j\}$. The weight on the edge of $p'$ which indicates $Y_j=y_j$ is $AF_{Y_j}(d_{Y_j}+1)P_P\{Y_j=y_j\hspace{0.1cm}|\hspace{0.1cm}\text{Pa}(Y_j)=v'_j\}$.\\

Now that we know the weights of all edges in $p$ and $p'$ we can evaluate $r(o)$ and $r(o')$.
\begin{align*}
r(&o)  = \alpha + AF_X(d_X+1)\frac{n_X-i+1}{n_X} + \sum_{j=1}^\ell AF_{Y_j}(d_{Y_j}+1)P_P\{Y_j=y_j\hspace{0.1cm}|\hspace{0.1cm}\text{Pa}(Y_j)=v_j\},\\
r(&o')  = \alpha + AF_X(d_X+1)\frac{n_X-(i+1)+1}{n_X} + \sum_{j=1}^\ell AF_{Y_j}(d_{Y_j}+1)P_P\{Y_j=y_j\hspace{0.1cm}|\hspace{0.1cm}\text{Pa}(Y_j)=v'_j\}.
\end{align*}

Recall that, for any $Z\in V$, $AF_Z={\prod_{W\in \text{Anc}(Z)} \frac{1}{n_W}}$. If $Y_j$ is a child of $X$, then\break$\text{Anc}(X)\cup\{X\}\subseteq \text{Anc}(Y_j)$. Thus, $AF_{Y_j}=AF_X\frac{1}{n_X}\beta_j$, for some $0<\beta_j\leq 1$.\\

Notice that, for any $Z\in V$, $z\in \text{Dom}(Z)$, and $w\in \text{Dom}(\text{Pa}(Z))$, we have\break$\frac{1}{n_Z}\leq P_P\{Z=z\hspace{0.1cm}|\hspace{0.1cm}\text{Pa}(Z)=w\}\leq 1$. Thus, for any $1\leq j\leq \ell$, we have
\begin{equation}
\nonumber
\frac{1}{n_{Y_j}}\leq P_P\{Y_j=y_j\hspace{0.1cm}|\hspace{0.1cm}\text{Pa}(Y_j)=v_j\}\leq 1,
\end{equation}
\begin{equation}
\nonumber
\frac{1}{n_{Y_j}}\leq P_P\{Y_j=y_j\hspace{0.1cm}|\hspace{0.1cm}\text{Pa}(Y_j)=v'_j\}\leq 1.
\end{equation}

Using these results we can rewrite $r(o)$ and $r(o')$ and obtain the following inequalities:
\begin{equation}
\nonumber
\begin{split}
r(o) & = \alpha + AF_X(d_X+1)\frac{n_X-i+1}{n_X} + \sum_{j=1}^\ell AF_X\frac{1}{n_X}\beta_j(d_{Y_j}+1)P_P\{Y_j=y_j\hspace{0.1cm}|\hspace{0.1cm}\text{Pa}(Y_j)=v_j\}\\
& \geq \alpha + AF_X(d_X+1)\frac{n_X-i+1}{n_X} + \sum_{j=1}^\ell AF_X\frac{1}{n_X}\beta_j(d_{Y_j}+1)\frac{1}{n_{Y_j}},\\
\end{split}
\end{equation}
\begin{equation}
\nonumber
\begin{split}
r(o') & = \alpha + AF_X(d_X+1)\frac{n_X-(i+1)+1}{n_X} + \sum_{j=1}^\ell AF_X\frac{1}{n_X}\beta_j(d_{Y_j}+1)P_P\{Y_j=y_j\hspace{0.1cm}|\hspace{0.1cm}\text{Pa}(Y_j)=v'_j\}\\
& \leq \alpha + AF_X(d_X+1)\frac{n_X-(i+1)+1}{n_X} + \sum_{j=1}^\ell AF_X\frac{1}{n_X}\beta_j(d_{Y_j}+1)\cdot 1.
\end{split}
\end{equation}

Next, we show that $d_X=\sum_{j=1}^\ell (d_{Y_j}+1)$. As $|\text{Ch}(X)|=\ell$ there are exactly $\ell$ directed paths of length 1 in the structure of $N$ that originate at $X$. Thus, $d_X-\ell$ is the number of directed paths of length greater than $1$ that originate at $X$ (in the structure of $N$). Every such path can be turned into a distinct directed path that originates at one of $\{Y_1,...,Y_{\ell}\}$ by removing the first edge. Further, any path that originates at some $Y_j\in\{Y_1,...,Y_{\ell}\}$ can be turned into a distinct directed path of length greater than $1$ that originates at $X$ by attaching $X\rightarrow Y_j$ to the beginning. Thus, the number of directed paths of length greater than $1$ that originate at $X$ is equal to the number of directed paths that originate at some $Y_j\in\{Y_1,...,Y_{\ell}\}$. That is, $d_X-\ell=\sum_{j=1}^\ell d_{Y_j}$ or equivalently $d_X = \sum_{j=1}^\ell d_{Y_j} + \ell = \sum_{j=1}^\ell (d_{Y_j}+1)$.\\

Recall that our aim is to show that $r(o)>r(o')$. For the purposes of contradiction, suppose $r(o)\leq r(o')$.
\begin{align*}
\nonumber
\begin{split}
\text{Then } & \alpha + AF_X(d_X+1)\frac{n_X-i+1}{n_X} + \sum_{j=1}^\ell AF_X\frac{1}{n_X}\beta_j(d_{Y_j}+1)\frac{1}{n_{Y_j}}\\
& \leq \alpha + AF_X(d_X+1)\frac{n_X-(i+1)+1}{n_X} + \sum_{j=1}^\ell AF_X\frac{1}{n_X}\beta_j(d_{Y_j}+1)\cdot 1\\
\implies & (d_X+1)(n_X-i+1) + \sum_{j=1}^\ell \beta_j(d_{Y_j}+1)\frac{1}{n_{Y_j}} \leq(d_X+1)(n_X-i) + \sum_{j=1}^\ell \beta_j(d_{Y_j}+1)\cdot 1\\
\implies & d_X+1\leq\sum_{j=1}^\ell\beta_j(d_{Y_j}+1)\bigg(1-\frac{1}{n_{Y_j}}\bigg)\\
\end{split}
\end{align*}
Now let $m=max\{n_{Y_j}\hspace{0.1cm}|\hspace{0.1cm}1\leq j\leq \ell\}$. This implies 
\begin{align*}
d_X+1\leq\sum_{j=1}^\ell\beta_j(d_{Y_j}+1)\bigg(1-\frac{1}{m}\bigg)\\
\end{align*}
Since $0<\beta_j\leq 1$ for all $1\leq j\leq \ell$, it follows that 
\begin{align*}
d_X+1\leq\sum_{j=1}^\ell 1\cdot (d_{Y_j}+1)\bigg(1-\frac{1}{m}\bigg)\\
\end{align*}
Recall that $d_X=\sum_{j=1}^\ell (d_{Y_j}+1)$, this implies
\begin{align*}
(d_X+1)\leq d_X\bigg(1-\frac{1}{m}\bigg)
\end{align*} 

If $X$ has no descendent paths, that is $d_X=0$, then we have shown $r(o)\leq r(o') \implies$\break$1\leq 0$. So we have derived a contradiction.\\

If $d_X>0$, then $r(o)\leq r(o')$ implies that
\begin{align*}
\begin{split}
 & 1 + \frac{1}{d_X} \leq 1-\frac{1}{m} < 1\\
\implies & \frac{1}{d_X} <0\\
\implies & 1< 0.
\end{split}
\end{align*}
Thus, we have again derived a contradiction and so we can conclude $r(o)>r(o')$. $\blacksquare$\\

\noindent \textbf{Proof of Lemma 2.} Let $Y\in\text{Ch}(X)$, then by the reasoning given in the proof of Theorem~1, $AF_Y=AF_X\frac{1}{n_X}\beta_Y$ for some $0<\beta_Y\leq 1$. Also, $\sum_{Y\in \text{Ch}(X)}(d_Y+1) = d_X$, as in the proof of Theorem 1.\\

Suppose for contradiction $L(X)\leq 0$, this implies
\begin{align*}
 & AF_X (d_X+1) \frac{1}{n_X} - \sum_{Y\in \text{Ch}(X)}AF_Y(d_Y+1)\frac{n_Y-1}{n_Y} \leq 0\\
\implies & AF_X (d_X+1) \frac{1}{n_X} \leq \sum_{Y\in \text{Ch}(X)}AF_X\frac{1}{n_X}\beta_Y(d_Y+1)\frac{n_Y-1}{n_Y}\\
\implies & (d_X+1) \leq \sum_{Y\in \text{Ch}(X)}\beta_Y(d_Y+1)\frac{n_Y-1}{n_Y}\\
\end{align*}
As $\beta_Y,\frac{n_Y-1}{n_Y}\leq 1$ for all $Y\in\text{Ch}(X)$, it follows that
\begin{align*}
 (d_X+1) \leq \sum_{Y\in \text{Ch}(X)}(d_Y+1) = d_X\\
\end{align*}
Thus, we have reached a contradiction and so can conclude $L(X)>0$. $\blacksquare$\\

\noindent\textbf{Proof of Corollary 3.} If $N\vDash o_1\succ o_2$, then there exists a sequence of outcomes $o_2=~p_1,p_2,...,p_m=o_1$, such that $N\vDash p_1\prec p_2\prec ...\prec p_m$ and $p_i$ and $p_{i+1}$ differ on the value taken by exactly one variable \citep{boutJAIR2004}. That is, starting at $o_2$, we can reach $o_1$ through $m-1$ improving variable flips. By Theorem 1, we know that $r(p_{i+1})-r(p_{i})>0$. We can rewrite $r(o_1)-r(o_2)$ as the sum of the rank improvements of each flip  as follows
\begin{equation}
\nonumber
r(o_1)-r(o_2)=[r(p_2)-r(p_1)] + [r(p_3)-r(p_2)] + \cdots +[r(p_m)-r(p_{m-1})].
\end{equation}

Suppose $\alpha\rightarrow \beta$ is an improving flip of variable $X$, that is, $\alpha$ and $\beta$ differ only on the value taken by $X$ and $N\vDash \beta\succ \alpha$. Thus, $X$ must be in a more preferred position in $\beta$ than~$\alpha$, given $\text{Pa}(X)=\beta[\text{Pa}(X)](=\alpha[\text{Pa}(X)])$.\\

The only variables whose preference position may differ in $\alpha$ and $\beta$ are $X$ and the children of $X$, Ch$(X)$. Also, recall that for any variable $Y\in V$, and any parental assignment $z\in~\text{Dom}(\text{Pa}(Y))$, we have that $P_P\{Y=y\hspace{0.1cm}|\hspace{0.1cm}\text{Pa}(Y)=z\}\in \{1/n_Y, 2/n_Y,..., 1 \}$ for every $y\in~\text{Dom}(Y)$. Thus, we can deduce the following lower bound on the increase in rank, $r(\beta)-r(\alpha)$.

\begin{align*}
\begin{split}
r(\beta)-r(\alpha) = & \Bigg[AF_X(d_X+1)P_P\{X=\beta[X]\hspace{0.1cm}|\hspace{0.1cm}\text{Pa}(X)=\beta[\text{Pa}(X)]\} \\
& + \sum_{Y\in \text{Ch}(X)} AF_Y(d_Y+1)P_P\{Y=\beta[Y]\hspace{0.1cm}|\hspace{0.1cm}\text{Pa}(Y)=\beta[\text{Pa}(Y)]\}\Bigg] \\
& - \Bigg[AF_X(d_X+1)P_P\{X=\alpha[X]\hspace{0.1cm}|\hspace{0.1cm}\text{Pa}(X)=\alpha[\text{Pa}(X)]\} \\
& + \sum_{Y\in \text{Ch}(X)} AF_Y(d_Y+1)P_P\{Y=\alpha[Y]\hspace{0.1cm}|\hspace{0.1cm}\text{Pa}(Y)=\alpha[\text{Pa}(Y)]\}\Bigg]
\intertext{\hspace{5cm}}
\end{split}
\end{align*}
Recall that $P_P\{Y=y\hspace{0.1cm}|\hspace{0.1cm}\text{Pa}(Y)=z\}\in \bigg{\{}\frac{1}{n_Y},\frac{2}{n_Y},..., 1\bigg{\}} \hspace{0.2cm} \forall Y\in V,y\in \text{Dom}(Y), z\in \text{Dom}(\text{Pa}(Y))$. Thus, we have that

\begin{align*}
r(\beta)-r(\alpha)\geq & AF_X(d_X+1) \bigg[P_P\{X=\beta[X]\hspace{0.1cm}|\hspace{0.1cm}\text{Pa}(X)=\beta[\text{Pa}(X)]\} - \\
& P_P\{X=\alpha[X]\hspace{0.1cm}|\hspace{0.1cm}\text{Pa}(X)=\alpha[\text{Pa}(X)]\}\bigg]+ \sum_{Y\in \text{Ch}(X)} AF_Y(d_Y+1) \bigg[\frac{1}{n_Y} - 1\bigg]
\end{align*}
As $P_P\{X=\beta[X]\hspace{0.1cm}|\hspace{0.1cm}\text{Pa}(X)=\beta[\text{Pa}(X)]\} > P_P\{X=\alpha[X]\hspace{0.1cm}|\hspace{0.1cm}\text{Pa}(X)=\alpha[\text{Pa}(X)]\}$, we have that
\begin{align*}
\begin{split}
r(\beta)-r(\alpha)\geq &  AF_X(d_X+1) \frac{1}{n_X} - \sum_{Y\in \text{Ch}(X)} AF_Y(d_Y+1) \frac{n_Y-1}{n_Y}\\
=& L(X)>0.
\end{split}
\end{align*}

In order to reach $o_1$ from $o_2$, each $X\in D$ must be flipped at least once in the sequence of $m-1$ flips. We know from the above that any improving flip of $X$ corresponds a rank increase of at least $L(X)$. Thus, as $r(o_1)-r(o_2)$ is the sum of the rank increases of each of the $m-1$ flips (each of which has been shown to produce an increase in rank by the above), we have that $r(o_1)-r(o_2)\geq \sum_{X\in D}L(X)$. As $N\vDash o_1\succ o_2$, we cannot have $o_1=o_2$, thus $D\neq\varnothing$ and so $\sum_{X\in D}L(X)>0$ by Lemma 2. $\blacksquare$\\

\noindent\textbf{Proof of Theorem 2.} Note that, for the entirety of this proof, $P_P$ refers to the generalised definition given in \S7.\\

The preference graph for $N$ is defined as before, with the addition of undirected edges for indifference. That is, if $o_1$ and $o_2$  are outcomes that differ only on variable $X$, and $o_1[X]$ is preferred to $o_2[X]$, given the values assigned to $\text{Pa}(X)$ in both $o_1$ and $o_2$, then there is an edge $o_2\rightarrow o_1$ in the preference graph. If the user is indifferent between $o_1[X]$ and $o_2[X]$, given the values assigned to $\text{Pa}(X)$, then there is an undirected edge between $o_1$ and $o_2$ in the preference graph.\\

Thus, $N\vDash o\succ o'$ if and only if there is a path $o'\leadsto o$ in the preference graph which may utilise undirected edges but must utilise at least one directed edge. This means that $N\vDash o\succ o'$ if and only if there exists a sequence of outcomes $o=o_1,o_2,...,o_m=o'$ such that, for all $i$, $o_i$ and $o_{i+1}$ differ on the value of exactly one variable and either $N\vDash o_i\succ o_{i+1}$ or $N\vDash o_i\sim o_{i+1}$ (with $N\vDash o_j\succ o_{j+1}$ for some $j$).\\

Also, $N\vDash o\sim o'$ if and only if there is a path between $o$ and $o'$ that exclusively uses undirected edges. This means that $N\vDash o\sim o'$ if and only if there exists a sequence of outcomes $o=o_1,o_2,...,o_m=o'$ such that, for all $i$, $o_i$ and $o_{i+1}$ differ on the value of exactly one variable and $N\vDash o_i\sim o_{i+1}$.\\

The above results mean that it is sufficient to prove that $N\vDash o\succ o'\implies r_G(o)>r_G(o')$ and $N\vDash o\sim o'\implies r_G(o)=r_G(o')$ in the case where $o$ and $o'$ differ on exactly one variable. The more general results then follow by these specific results and the transitivity of $=$ and $>$.\\

Let us assume that $o$ and $o'$ differ only on the value taken by $X\in V$. Let $X$ take the value $x$ in $o$ ($o[X]=x$) and the value $x'$ in on $o'$ ($o'[X]=x'$). \\

First, we show that $N\vDash o\sim o'\implies r_G(o)=r_G(o')$. Assume that $N\vDash o\sim o'$. Let $u=o[\text{Pa}(X)]=o'[\text{Pa}(X)]$. Recall that
\begin{equation}
\nonumber
r_G(o)=\sum_{Z\in V} AF_Z(d_Z+1)P_P\{Z=o[Z]\hspace{0.1cm}|\hspace{0.1cm}\text{Pa}(Z)=o[\text{Pa}(Z)]\}
\end{equation}
and  similarly for $r_G(o')$. Thus, to evaluate $r_G(o)$ and $r_G(o')$, and subsequently prove that $r_G(o)=r_G(o')$, we must first evaluate these summation terms for all $Z\in V$ for both $r_G(o)$ and $r_G(o')$.\\

Let $Y\in V$ be a variable such that $Y\neq X$ and $X\not\in \text{Pa}(Y)$. Then, as $o$ and $o'$ differ only on the value of $X$, $Y$ and $\text{Pa}(Y)$ must take the same values in both $o$ and $o'$. Let $y=o[Y]=o'[Y]$ and $w=o[\text{Pa}(Y)]=o'[\text{Pa}(Y)]$. Then we have
\begin{equation}
\nonumber
AF_Y(d_Y+1)P_P\{Y=o[Y]\hspace{0.1cm}|\hspace{0.1cm}\text{Pa}(Y)=o[\text{Pa}(Y)]\} = AF_Y(d_Y+1)P_P\{Y=y\hspace{0.1cm}|\hspace{0.1cm}\text{Pa}(Y)=w\},
\end{equation}
\begin{equation}
\nonumber
AF_Y(d_Y+1)P_P\{Y=o'[Y]\hspace{0.1cm}|\hspace{0.1cm}\text{Pa}(Y)=o'[\text{Pa}(Y)]\} = AF_Y(d_Y+1)P_P\{Y=y\hspace{0.1cm}|\hspace{0.1cm}\text{Pa}(Y)=w\}.
\end{equation}
Thus, any such variable (that is, any variable that is neither $X$ itself, nor a child of $X$) contributes exactly the same quantity to both sums, $r_G(o)$ and $r_G(o')$.\\

As $N\vDash o\sim o'$, and $o$ and $o'$ differ only on $X$, we must have that $x\sim x'$ under $\text{Pa}(X)=u$. Therefore $x$ and $x'$ are in the same preference position in the row of $\text{CPT}(X)$ corresponding to $\text{Pa}(X)=u$. Let $x$ and $x'$ be in preference position $i$, given $\text{Pa}(X)=u$.\\

Consider the summation term contributed by $X$. By our assumptions about $o$ and~$o'$ the~$X$ summation terms in $r_G(o)$ and $r_G(o')$ (respectively) are
\begin{equation}
\nonumber
AF_X(d_X+1)P_P\{Y=x\hspace{0.1cm}|\hspace{0.1cm}\text{Pa}(X)=u\} = AF_X(d_X+1)\frac{n_X-\ell-i+1}{n_X-\ell},
\end{equation}
\begin{equation}
\nonumber
AF_X(d_X+1)P_P\{Y=x'\hspace{0.1cm}|\hspace{0.1cm}\text{Pa}(X)=u\} = AF_X(d_X+1)\frac{n_X-\ell-i+1}{n_X-\ell},
\end{equation}

\noindent where $\ell$ is the number of indifferences in the preference ordering over $\text{Dom}(X)$ under\break$\text{Pa}(X)=u$, given in $\text{CPT}(X)$. Thus, $X$ contributes exactly the same quantity to both sums, $r_G(o)$ and $r_G(o')$.\\

Finally, we must consider the weights contributed by  $\text{Ch}(X)=\{Y_1,...,Y_k\}$. Let\break$y_j=o[Y_j]=o'[Y_j]$, $v_j=o[\text{Pa}(Y_j)]$, and $v'_j=o'[\text{Pa}(Y_j)]$. The $Y_j$ summation term in~$r_G(o)$ is $AF_{Y_j}(d_{Y_j}+1)P_P\{Y_j=y_j\hspace{0.1cm}|\hspace{0.1cm}\text{Pa}(Y_j)=v_j\}$. The $Y_j$ summation term in $r_G(o')$ is\break$AF_{Y_j}(d_{Y_j}+1)P_P\{Y_j=y_j\hspace{0.1cm}|\hspace{0.1cm}\text{Pa}(Y_j)=v'_j\}$.\\

Note that $v_j$ and $v'_j$ differ only on the value taken by $X$. We assume, in general, that flipping a variable between values to which the user is indifferent should not be allowed to affect the user's preference over that variable's children \citep{boutJAIR2004}. Here, the user is indifferent between $x$ and $x'$ and the only difference between $v_j$ and $v'_j$ is whether $X=x$ or $X=x'$. By our assumption, the user's preference over $Y_j$ should be identical under $\text{Pa}(Y_j)=v_j$ and $\text{Pa}(Y_j)=v'_j$. This means, under both $\text{Pa}(Y_j)=v_j$ and $\text{Pa}(Y_j)=v'_j$, there are the same number of indifferences in the preference order over Dom$(Y_j)$, and $y_j$ is in the same position of preference in this preference order. By our new definition of $P_P$ (\S7), this implies $P_P\{Y_j=y_j\hspace{0.1cm}|\hspace{0.1cm}\text{Pa}(Y_j)=v_j\}= P_P\{Y_j=y_j\hspace{0.1cm}|\hspace{0.1cm}\text{Pa}(Y_j)=v'_j\}$. Thus, $Y_j$ contributes exactly the same quantity to both sums, $r_G(o)$ and $r_G(o')$.\\

We have shown that all variables $Z\in V$, contribute exactly the same quantity to both sums, $r_G(o)$ and $r_G(o')$. Thus, we must have $r_G(o)=r_G(o')$. We have therefore shown that $N\vDash o\sim o'\implies r_G(o)=r_G(o')$.\\

Next, we show that $N\vDash o\succ o'\implies r_G(o)>r_G(o')$. Suppose $N\vDash o\succ o'$. Let $u=o[\text{Pa}(X)]=o'[\text{Pa}(X)]$ again. Let $x_1\succsim x_2\succsim\cdots\succsim x_m$ be the preference ordering of~$\text{Dom}(X)$ given that $\text{Pa}(X)=u$. This is the row of $\text{CPT}(X)$ that corresponds to $\text{Pa}(X)=u$. Suppose $x=x_i$ and $x'=x_j$, we know that $i<j$ as $o'\rightarrow o$ is an improving flip of $X$.\\

Let $o_k$ denote the outcome that has $o_k[X]=x_k$ and for all variables $Y\neq X$ has $o_k[Y]=o[Y](=o'[Y])$. Then $o_1,...,o_m$ is a sequence of flips of $X$ through the values $x_1,x_2,...,x_m$. As $\text{Pa}(X)=u$ in all $o_k$, this is a sequence of improving or indifferent flips (as $x_1\succsim x_2\succsim\cdots\succsim x_m$ when $\text{Pa}(X)=u$). Thus we have $N\vDash o_1\succsim o_2\succsim\cdots \succsim o_m$. Notice that $o=o_i$ and $o'=o_j$ for $i<j$ so we have $N\vDash o=o_i\succsim o_{i+1}\succsim\cdots \succsim o_j=o'$. This shows that it is sufficient to prove that $r_G(o)>r_G(o')$ for the special case where $x$ and $x'$ are adjacent in $x_1,x_2,...,x_m$, that is $j=i+1$. The more general case, when $x$ and $x'$ are not adjacent, follows from this specific case and that $N\vDash o\sim o'\implies r_G(o)=r_G(o')$ as proven above (this can be seen via similar reasoning to that given in the proof of Theorem 1). Thus, we can assume that $x$ and $x'$ are adjacent values. This implies that $x$ and $x'$ are either (one of) the $i^{th}$ and (one of) the $(i+1)^{th}$ most preferred values of $X$ respectively, given $\text{Pa}(X)=u$, or they are in same preference position. However, $x$ and $x'$ cannot be in the same preference position, as then we must have $x\sim x'$ under $\text{Pa}(X)=u$ and therefore $N\vDash o\sim o'$. This is a contradiction to our assumption that $N\vDash o\succ o'$. So we may assume that $x$ and $x'$ are (one of) the $i^{th}$ and (one of) the $(i+1)^{th}$ most preferred values of $X$ respectively, given $\text{Pa}(X)=u$.\\

We have now assumed that $o$ and $o'$ are outcomes associated with $N$, such that $o$ and~$o'$ differ only on the value taken by $X\in V$ such that $o[X]=x$ and $o'[X]=x'$. Further, under the values assigned to $\text{Pa}(X)$, $u$, by both $o$ and $o'$, we have assumed that $x$ is (one of) the $i^{th}$ most preferred value(s) of $X$ and $x'$ is (one of) the $(i+1)^{th}$ most preferred value(s).\\

In order to evaluate $r_G(o)$ and $r_G(o')$, and subsequently prove that $r_G(o)>r_G(o')$, we must first consider the individual summation terms in $r_G(o)$ and $r_G(o')$, as we did in the indifference case above.\\

Let $Y\in V$ be a variable such that $Y\neq X$ and $X\not\in \text{Pa}(Y)$. Then by the same reasoning as in the indifference case above, $Y$ contributes exactly the same quantity to both sums,~$r_G(o)$ and $r_G(o')$. Let $\alpha$ denote the total contribution to $r_G(o)$ (and thus to $r_G(o')$ also) by such variables.\\

Now, consider the $X$ summation terms.  By our assumptions about $o$ and $o'$ the $X$ summation terms in $r_G(o)$ and $r_G(o')$ (respectively) are
\begin{equation}
\nonumber
AF_X(d_X + 1)P_P\{X=x\hspace{0.1cm}|\hspace{0.1cm}\text{Pa}(X)=u\} = AF_X(d_X + 1) \frac{n_X-\ell-i+1}{n_X-\ell},
\end{equation}
\begin{equation}
\nonumber
AF_X(d_X + 1)P_P\{X=x'\hspace{0.1cm}|\hspace{0.1cm}\text{Pa}(X)=u\} = AF_X(d_X + 1) \frac{n_X-\ell-(i+1)+1}{n_X-\ell},
\end{equation}
where $\ell$ is the number of indifferences in the preference ordering over $\text{Dom}(X)$ under $\text{Pa}(X)=~u$, given in $\text{CPT}(X)$. Note that $0\leq \ell\leq n_X-2$ and $1\leq i\leq n_X-\ell-1$.\\

Finally, we must consider the weights contributed by  $\text{Ch}(X)=\{Y_1,...,Y_k\}$. Let\break$y_j=o[Y_j]=o'[Y_j]$, $v_j=o[\text{Pa}(Y_j)]$, and $v'_j=o'[\text{Pa}(Y_j)]$. The $Y_j$ summation term in~$r_G(o)$ is $AF_{Y_j}(d_{Y_j}+1)P_P\{Y_j=y_j\hspace{0.1cm}|\hspace{0.1cm}\text{Pa}(Y_j)=v_j\}$. The $Y_j$ summation term in $r_G(o')$ is\break$AF_{Y_j}(d_{Y_j}+1)P_P\{Y_j=y_j\hspace{0.1cm}|\hspace{0.1cm}\text{Pa}(Y_j)=v'_j\}$.\\

Now that we know all of the summation terms we can evaluate $r_G(o)$ and $r_G(o')$ as follows.
\begin{align*}
r_G(o) & = \alpha + AF_X(d_X+1)\frac{n_X-\ell-i+1}{n_X-\ell} + \sum_{j=1}^k AF_{Y_j}(d_{Y_j}+1)P_P\{Y_j=y_j\hspace{0.1cm}|\hspace{0.1cm}\text{Pa}(Y_j)=v_j\},\\
r_G(o') & = \alpha + AF_X(d_X+1)\frac{n_X-\ell-(i+1)+1}{n_X-\ell} + \sum_{j=1}^k AF_{Y_j}(d_{Y_j}+1)P_P\{Y_j=y_j\hspace{0.1cm}|\hspace{0.1cm}\text{Pa}(Y_j)=v'_j\}.
\end{align*}

By the same reasoning given in the proof of Theorem 1, $AF_{Y_j}=AF_X\frac{1}{n_X}\beta_j$, for some $0<\beta_j\leq 1$.\\

Let $Z\in V$ be any variable, $z\in \text{Dom}(Z)$, $w\in \text{Dom}(\text{Pa}(Z))$, and let $\ell$ be the number of indifferences in the row of $\text{CPT}(Z)$ that corresponds to $\text{Pa}(Z)=w$. Then, by definition, $\frac{1}{n_Z-\ell}\leq P_P\{Z=z\hspace{0.1cm}|\hspace{0.1cm}\text{Pa}(Z)=w\}\leq 1$. As $0\leq \ell\leq n_Z-1$, this means that\break$\frac{1}{n_Z}\leq P_P\{Z=z\hspace{0.1cm}|\hspace{0.1cm}\text{Pa}(Z)=w\}\leq 1$.  Thus, for any $1\leq j\leq k$, we have that 
\begin{equation}
\nonumber
\frac{1}{n_{Y_j}}\leq P_P\{Y_j=y_j\hspace{0.1cm}|\hspace{0.1cm}\text{Pa}(Y_j)=v_j\}\leq 1,
\end{equation}
\begin{equation}
\nonumber
\frac{1}{n_{Y_j}}\leq P_P\{Y_j=y_j\hspace{0.1cm}|\hspace{0.1cm}\text{Pa}(Y_j)=v'_j\}\leq 1.
\end{equation}

Using these results we can rewrite $r_G(o)$ and $r_G(o')$ and obtain the following inequalities:
\begin{equation}
\nonumber
\begin{split}
r_G(o) & = \alpha + AF_X(d_X+1)\frac{n_X-\ell-i+1}{n_X-\ell} + \sum_{j=1}^k AF_X\frac{1}{n_X}\beta_j(d_{Y_j}+1)P_P\{Y_j=y_j\hspace{0.1cm}|\hspace{0.1cm}\text{Pa}(Y_j)=v_j\}\\
& \geq \alpha + AF_X(d_X+1)\frac{n_X-\ell-i+1}{n_X-\ell} + \sum_{j=1}^k AF_X\frac{1}{n_X}\beta_j(d_{Y_j}+1)\frac{1}{n_{Y_j}},\\
\end{split}
\end{equation}
\begin{equation}
\nonumber
\begin{split}
r_G(o') = & \alpha + AF_X(d_X+1)\frac{n_X-\ell-(i+1)+1}{n_X-\ell}\\
& + \sum_{j=1}^k AF_X\frac{1}{n_X}\beta_j(d_{Y_j}+1)P_P\{Y_j=y_j\hspace{0.1cm}|\hspace{0.1cm}\text{Pa}(Y_j)=v'_j\}\\
\leq & \alpha + AF_X(d_X+1)\frac{n_X-\ell-(i+1)+1}{n_X-\ell} + \sum_{j=1}^k AF_X\frac{1}{n_X}\beta_j(d_{Y_j}+1)\cdot 1.
\end{split}
\end{equation}

Recall that our aim is to show that $r_G(o)>r_G(o')$. For the purposes of contradiction, suppose $r_G(o)\leq r_G(o')$. This implies 
\begin{align*}
\begin{split}
 & \alpha + AF_X(d_X+1)\frac{n_X-\ell-i+1}{n_X-\ell} + \sum_{j=1}^k AF_X\frac{1}{n_X}\beta_j(d_{Y_j}+1)\frac{1}{n_{Y_j}}\\
& \leq \alpha + AF_X(d_X+1)\frac{n_X-\ell-(i+1)+1}{n_X-\ell} + \sum_{j=1}^k AF_X\frac{1}{n_X}\beta_j(d_{Y_j}+1)\cdot 1\\
\implies & (d_X+1)\frac{n_X-\ell-i+1}{n_X-\ell} + \sum_{j=1}^k \frac{1}{n_X}\beta_j(d_{Y_j}+1)\frac{1}{n_{Y_j}}\\
& \leq (d_X+1)\frac{n_X-\ell-i}{n_X-\ell} + \sum_{j=1}^k \frac{1}{n_X}\beta_j(d_{Y_j}+1)\cdot 1\\
\implies & (d_X+1)\frac{n_X}{n_X-\ell}\leq \sum_{j=1}^k \beta_j(d_{Y_j}+1)\bigg(1- \frac{1}{n_{Y_j}}\bigg)\\
\end{split}
\end{align*}
As $0\leq \ell\leq n_X-2$, and thus $1\leq\frac{n_X}{n_X-\ell}$, it follows that
\begin{align*}
d_X +1 \leq \sum_{j=1}^k \beta_j(d_{Y_j}+1)\bigg(1- \frac{1}{n_{Y_j}}\bigg)
\end{align*}
From this point, we derive a contradiction in an identical manner to the proof of Theorem~1. Thus, we have shown $N\vDash o\succ o'\implies r_G(o)>r_G(o')$. $\blacksquare$

\section*{Appendix B. Algorithm Details}

In this section, we describe how CP-nets and outcomes should be formatted as inputs to Algorithm 1, given in \S4. We also explain how Algorithm 1 works and why it is correct.
\vspace*{-0.2cm}
\subsection*{B.1 Input Formats for Algorithm 1} 
For this section, suppose we have a CP-net $N$, over a set of variables $V=\{X_1,...,X_n\}$, which are in  a topological ordering with respect to the structure of $N$. Further, suppose $\text{Dom}(X_i)=\{x_i^1,...,x_i^{n_i}\}$.\\

CP-nets are input to Algorithm 1 as a pair, $N=(A,CPT)$. The first entry $A$, is the adjacency matrix for the structure of $N$ as described in \S4.\\

\textbf{Example 8.} For the CP-net given in Example 1, the adjacency matrix would be:
\[
\begin{blockarray}{ccccc}
& A & B & C & D \\
\begin{block}{c(cccc)}
  A & 0 & 0 & 1 & 0 \\
  B & 0 & 0 & 1 & 0 \\
  C & 0 & 0 & 0 & 1 \\
  D & 0 & 0 & 0 & 0 \\
\end{block}
\end{blockarray}
 \] 
\indent The second entry in the pair is the set of CPTs associated with $N$. We input $CPT$ as  a list of the CPTs, so for any $1\leq i\leq n$, $CPT[i]=\text{CPT}(X_i)$.\\

Let Pa$(X_i)=\{X_{\beta_1},...,X_{\beta_\ell}\}$ ($\beta_1<\beta_2\cdots <\beta_\ell$).\\

Let $u$ be an assignment of values to Pa$(X_i)$, $u=x_{\beta_1}^{\alpha_1}\cdots x_{\beta_\ell}^{\alpha_\ell}$, so $u$ is a $|\text{Pa}(X_i)|$-tuple in~$\text{Dom}(\text{Pa}(X_i))$.\\

Then $\text{CPT}(X_i)$ is input as an array such that $\text{CPT}(X_i)[\alpha_1,...,\alpha_\ell]$ is an $|\text{Dom}(X_i)|$-tuple,~$\sigma$.\\

For all $1\leq k\leq |\text{Dom}(X_i)|$, $\sigma[k]$ is the position of preference of $x_i^k$ according to the CPTs, given that Pa$(X_i)=u$ ($\sigma[k]=1$ if $x_i^k$ is the most preferred and so on).\\

\newpage\textbf{Example 9.} For the CP-net given in Example 1, recall that $\text{CPT}(C)$ is as follows:
\begin{center}
\begin{tabular}{|c|c|}
\hline
$ab$ & $c\succ \bar{c}\succ \bar{\bar{c}}$\\
$a\bar{b}$ & $\bar{c}\succ \bar{\bar{c}}\succ c$\\
$\bar{a}\bar{b}$ & $\bar{\bar{c}}\succ \bar{c}\succ c$\\
$\bar{a}b$ & $\bar{\bar{c}}\succ c\succ\bar{c}$\\
\hline
\end{tabular}
\end{center}

For this example, $V=\{A,B,C,D\}$ (note that $B,A,C,D$ is also a valid topological ordering, we use $A,B,C,D$ for ease) and $CPT=[\text{CPT}(A),\text{CPT}(B),\text{CPT}(C),\text{CPT}(D)]$. We have $X_1=A, X_2=B, X_3=C$, and $\text{Dom}(A)=\{a,\bar{a}\}$, $\text{Dom}(B)=\{b,\bar{b}\}$, $\text{Dom}(C)=\{c,\bar{c},\bar{\bar{c}}\}$. Thus, $x_1^1=a, x_1^2=\bar{a}$, and $x_2^1=b, x_2^2=\bar{b}$, and $x_3^1=c, x_3^2=\bar{c}, x_3^3=\bar{\bar{c}}$. Also, Pa$(C)=\{A,B\}$, so we would input $\text{CPT}(C)$ ($CPT[3]$) as follows:
\begin{center}
\begin{tabular}{c|c|c|}
& $[\cdot,1]$ & $[\cdot,2]$\\
\hline
$[1,\cdot]$ & $(1,2,3)$ & $(3,1,2)$\\
\hline
$[2,\cdot]$ & $(2,3,1)$ & $(3,2,1)$\\
\hline
\end{tabular}
\end{center}
This says, for example, that $\text{CPT}(C)[2,1]=CPT[3][2,1]=(2,3,1)$. Here we are inputting the user's preference over $\text{Dom}(C)$ under $X_1=x_1^2$ and $X_2=x_2^1$, that is, $A=\bar{a}$ and $B=b$. We know that in this case we have $\bar{\bar{c}}\succ c\succ\bar{c}$, so $x_3^1=c$ is in preference position 2, $x_3^2=\bar{c}$ is in preference position 3, and $x_3^3=\bar{\bar{c}}$ is in preference position 1. Hence $\text{CPT}(C)[2,1]=(2,3,1)$.\\

Note that from this input $CPT[3]$, we can clearly extract $|\text{Dom}(C)|$ by looking at the length of the tuples in the array. To keep Algorithm 1 in \S4 as readable as possible, we assume that, given $1\leq i\leq n=|V|$, we can extract $|\text{Dom}(X_i)|$ from the CPTs input rather than putting the details of how this is achieved.\\

An outcome $o$, should be input as a $|V|$-tuple in $\{1,...,|\text{Dom}(X_1)|\}\times\cdots\times\{1,...,|\text{Dom}(X_n)|\}$. If $X_i$ takes value $x^k_i$ and $X_j$ takes value $x^\ell_j$ in outcome $o$, then $o[i]=k$ and $o[\{i,j\}]=(k,\ell)$. For our running example, consider the outcome $o=\bar{a}b\bar{\bar{c}}d$, we can rewrite this as $o=x_1^2x_2^1x_3^3x_4^1$ and we would input it as the tuple $(2,1,3,1)$. So, in $o$, $B$ takes value $b$, that is, $X_2$ takes value $x_2^1$, and so $o[2]=1$. Similarly $X_3$ takes value $x_3^3$ ($C$ takes value $\bar{\bar{c}}$) so $o[3]=3$.
\vspace*{-0.2cm}
\subsection*{B.2 Correctness of Algorithm 1}

In this section we give the details of how Algorithm 1 works and why it is correct. \\

Algorithm 1 takes the CP-net $N$, and some outcome $o$, and outputs the rank of this outcome $r(o)$. It calculates $r(o)$ by setting the value of $r(o)$ to 0 (step \textbf{1}) and successively adding the edge weights of the root to leaf path in $W$ that corresponds to $o$ (steps \textbf{2-11}). The weight attached to the edge indicating the value taken by $X_i$ in $o$ is given by Expression~\eqref{TreeWeight} in \S3.1.\\

The algorithm calculates the weight given by \eqref{TreeWeight} for each $X_i$ in several steps, and then adds it to the $r(o)$ term. The leftmost product term in \eqref{TreeWeight} is calculated in two steps (\textbf{3-4}). First, calling Algorithm 2, \emph{ancestor}, to obtain $\text{Anc}(X_i)$, and then forming the product of the inverses of the domains of all $Y\in \text{Anc}(X_i)$. We then call Algorithm 3, \emph{DP}, to obtain the number of descendent paths of $X_i$, $d_{X_i}$, (step \textbf{5}) in order to calculate the central product term in \eqref{TreeWeight}.\\

Extracting the rightmost term in \eqref{TreeWeight} from $N$ and $o$ is slightly more convoluted. The parent set of $X_i$, Pa$(X_i)$, is the set of variables $Y$, such that there is an edge $Y\rightarrow X_i$ in the structure of $N$. We can extract this set directly from the adjacency matrix (step~\textbf{6}). We can then find the values taken by Pa$(X_i)$ in $o$ by extracting the appropriate entries of $o$, we call this assignment to the parent variables $u$ (step \textbf{7}). So $u$ is a $|\text{Pa}(X_i)|$-tuple in $\text{Dom}(\text{Pa}(X_i))$. Next, we can find the user's order of preference over $\text{Dom}(X_i)$ under $\text{Pa}(X_i)=u$ by extracting the appropriate entry of the $\text{CPT}(X_i)$ array input, $\text{CPT}(X_i)[u]$ (step \textbf{8}). \\

\enlargethispage{\baselineskip}
The $k$ in the rightmost product of \eqref{TreeWeight} is the position of preference of the value taken by~$X_i$ in $o$ in the preference order we have just obtained. Thus, we can find $k$ by extracting the element of the order that indicates the position of preference of the value taken by $X_i$ in~$o$ (this is $o[i]$) (step \textbf{9}). Now that we have $k$ we can calculate the rightmost term in \eqref{TreeWeight} with $n_{X_i}=|\text{Dom}(X_i)|$ (\textbf{10}). Finally, we form the whole term given in \eqref{TreeWeight} and add it to the~$r(o)$ term (step \textbf{11}). Repeating this for every $X_i\in V$ gives us the rank of $o$ by definition. At this point, Algorithm 1 exits its ``for" loop and outputs $r(o)$ (step \textbf{12}).

\subsection*{B.3 Algorithms 2 and 3}
\hrule
\vspace*{0.1cm}
\textbf{Algorithm 2: Ancestor Algorithm}
\vspace*{0.1cm}
\hrule
\vspace*{0.1cm}
\textbf{Inputs:} $1\leq i\leq |V|$

\hspace{1.55cm}$A$ - Adjacency matrix of $N$
\vspace*{0.1cm}
\hrule
\vspace*{0.1cm}
\textbf{1} $\textbf{Anc}:= \textbf{0}_{|V|} $ \hspace{2cm}\#$\textbf{0}_{|V|}$ is the zero $|V|$-tuple

\textbf{2} $\textbf{a}:= A_{\cdot,i}$ \hspace{2cm}\#$A_{\cdot,i}$ is the $i^{th}$ column of $A$

\textbf{3} \textbf{while} $sum(\textbf{a})>0$

\textbf{4} \hspace*{0.5cm} $\textbf{Anc}= \textbf{Anc} + \textbf{a}$

\textbf{5} \hspace*{0.5cm} $\textbf{a}= A \textbf{a}$

\textbf{6} $Anc=\{X_j\hspace{0.1cm}|\hspace{0.1cm} \textbf{Anc}[j]\neq 0\}$ \hspace{1.5cm}\#The set of variables with a non-zero entry in $\textbf{Anc}$

\textbf{7} \textbf{return} $Anc$
\vspace*{0.1cm}
\hrule
\vspace*{0.5cm}

Algorithm 2, \emph{ancestor}, takes $i$ (an integer, $1\leq i\leq |V|$, indicating which variable's ancestor set we are interested in) and the adjacency matrix $A$ and outputs $\text{Anc}(X_i)$. For any $X\in V$, the following statements are equivalent.
\begin{equation}
\nonumber
Y\in \text{Anc}(X) \iff \exists \text{ directed } Y\leadsto X \text{ path} \iff (A^k)_{Y,X}\neq 0 \text{ for some } 1\leq k \leq |V|-1
\end{equation}
because $(A^k)_{i,j}=\#$ directed paths of length $k$ in $N$ originating at $X_i$ and terminating at~$X_j$. Also, no path in $N$ can be of length greater than $|V|-1$ as there are $|V|$ variables (vertices) in the (acyclic) structure. Thus, Algorithm 2 calculates Anc$(X_i)$ by summing, component-wise, the $i^{th}$ columns of $A^k$ for $k=1,...,|V|-1$. Then Anc$(X_i)$ are the variables whose corresponding element is non-zero.\\

\newpage\hrule
\vspace*{0.1cm}
\textbf{Algorithm 3: Descendent Path, DP, Algorithm}
\vspace*{0.1cm}
\hrule
\vspace*{0.1cm}
\textbf{Inputs:} $1\leq i\leq |V|$

\hspace{1.55cm} $A$ - Adjacency matrix of $N$
\vspace*{0.1cm}
\hrule
\vspace*{0.1cm}
\textbf{1} $\textbf{a}:=A_{i,\cdot}$ \hspace{2cm}\#$A_{i,\cdot}$ is the $i^{th}$ row of $A$

\textbf{2} $d:=0$

\textbf{3} \textbf{while} $sum(\textbf{a})>0$

\textbf{4} \hspace*{0.5cm} $d= d+ sum(\textbf{a})$

\textbf{5} \hspace*{0.5cm} $\textbf{a}=\textbf{a} A$

\textbf{6} \textbf{return} $d$
\vspace*{0.1cm}
\hrule
\vspace*{0.5cm}

Algorithm 3, \emph{DP}, takes $i$ (an integer, $1\leq i\leq |V|$, indicating which variable's descendent paths we are interested in) and the adjacency matrix $A$, and outputs $d_{X_i}$. As $(A^k)_{ij}=\break\#$ directed paths of length $k$ in $N$ originating at $X_i$ and terminating at $X_j$, then for any variable $X_i$, we have
\begin{equation}
\nonumber
\begin{split}
d_{X_i} & =\sum_{j=1}^n \#\text{directed paths } X_i\leadsto X_j\\
& = \sum_{j=1}^n \sum_{k=1}^{|V|-1} \#\text{directed paths } X_i\leadsto X_j \text{ of length } k\\
& = \sum_{j=1}^n \sum_{k=1}^{|V|-1} (A^k)_{i,j} = \sum_{k=1}^{|V|-1}\sum_{j=1}^n (A^k)_{i,j}
\end{split}
\end{equation}
Therefore, Algorithm 3 calculates $d_{X_i}$ by summing the entries of the $i^{th}$ rows of $A^k$ for $k=1,...,|V|-1$.

\section*{Appendix C. Zero Outcomes Traversed Cases}
In \S6, we experimentally evaluated the performance of seven different dominance testing functions by applying them to the same sets of dominance queries and recording outcomes traversed and time elapsed. These seven functions were all possible combinations of suffix fixing \citep{boutJAIR2004}, penalty pruning \citep{liAAMS2011}, and rank pruning (\S5). Each combination has certain conditions that would result in a dominance query being immediately found false, and the outcomes traversed would be recorded as zero, as discussed in~\S6.1. We shall refer to these as the \emph{initial conditions} of the pruning combinations. Suppose we wish to answer the dominance query $N\vDash o\succ o'$, a summary of these initial conditions is given below.\\

\begin{center}
\begin{tabular}{c|c}
Initial Condition & Pruning Combination\\
\hline
$o=o'$ & All\\
$f(o')<0$ & All combinations including penalty pruning\\
$r(o)-r(o')<L_D(o,o')$ & All combinations including rank pruning\\
\end{tabular}
\end{center}

In this section, we look at the proportion of queries from our \S6 experiment that resulted in zero outcomes traversed (that is, that met one of the initial conditions) for each function (pruning combination). This proportion shows us how often a dominance testing function's initial conditions are strong enough to immediately answer the query. Further, by comparing these proportions to the proportion of queries that were false, we can evaluate how well a function's initial conditions can predict the outcome of a dominance query.\\

First note that adding suffix fixing to a combination does not add any initial conditions. Thus, it is sufficient to evaluate these proportions only for the four functions that used rank pruning, penalty pruning, suffix fixing, and the combination of rank pruning and penalty pruning.\\

In the case of binary CP-nets, for each of $3\leq n\leq 10$, we tested all seven functions on a set of 1000 dominance queries in the \S6 experiment. Thus, each function answered the same set of 8000 dominance queries. Out of these queries, 5870 (0.73375) of them were false. Note that, despite the random generation of queries, this is not close to 0.5. This is because there are three possibilities, either $N\vDash o\succ o'$, $N\vDash o'\succ o$, or $o$ and $o'$ are incomparable. For the dominance query `$N\vDash o\succ o'$?', only the first case makes the query true, the other two cases imply that the query is false. In Table 1, for each function, we give the proportion of the 8000 queries that were determined to be false by the initial conditions (that is, were answered with zero outcomes traversed), $Z_P$, and the proportion of false queries that were identified as false by the initial conditions, $Z_P/0.73375$.\\

\begin{table}[h]
\begin{center}
\begin{tabular}{|c|c|c|c|c|}
\hline
& Rank & Penalty & Suffix Fixing & Rank + Penalty\\
\hline
$Z_P$ & 0.70463 & 0.63300 & 0.03288 &0.70513\\
\hline
$Z_P/0.73375$ &0.96031 & 0.86269 & 0.04480 & 0.96099\\
\hline
\end{tabular}
\end{center}
\caption{Zero Outcomes Traversed Proportions - Binary Case}
\end{table}

The $Z_P$ value for suffix fixing shows us the proportion of $o=o'$ cases. Thus, the initial rank condition determines $0.67175=0.70463-0.03288$ of the 8000 queries to be false immediately, and similarly for the initial penalty condition. Clearly, the rank condition is stronger than the penalty condition as it determines a greater number of queries to be false. Further, by looking at the $Z_P$ value for rank and penalty pruning combined, we can see that utilising both conditions is only a slight improvement upon the rank condition alone. The $Z_P/0.73375$ value shows us how many of the false dominance queries were detected by the initial conditions. Using rank pruning alone, over $96\%$ of the false dominance queries were determined to be false by the initial conditions. This suggests that our initial conditions could be used as fairly accurate predictor for the outcome of a dominance query. Any dominance query determined to be false by the rank pruning initial conditions is false. Of those that do not meet any of these initial conditions (that is, those we would `predict' to be true), only $(0.73375-Z_P)/(1-Z_P)\times 100= 9.860\%$ would actually be false (in these cases, $o$ and~$o'$ are incomparable). This percentage would be slightly smaller if both the rank and penalty pruning initial conditions were used. Thus, we could use these initial conditions, which are quick to check, to predict dominance query outcomes to a reasonable level of accuracy.\\

For the multivalued case, we tested all seven functions on a set of 1000 queries for each $3\leq n\leq 8$. So, in this case, we have 6000 dominance queries and 4520 (0.75333) of these were false. Table 2 gives the proportion of queries with zero outcomes traversed for each function, and the proportion of false queries identified by initial conditions for each function.\\

\begin{table}[h]
\begin{center}
\begin{tabular}{|c|c|c|c|c|}
\hline
& Rank & Penalty & Suffix Fixing & Rank + Penalty\\
\hline
$Z_P$ & 0.66250 & 0.55700 & 0.009 & 0.66583\\
\hline
$Z_P/0.75333$ & 0.87942 & 0.73938 & 0.01195 & 0.88385\\
\hline
\end{tabular}
\end{center}
\caption{Zero Outcomes Traversed Proportions - Multivalued Case}
\end{table}

These proportions show similar patterns to the binary case. The initial rank condition remains the strongest condition, determining the largest number of queries to be false. Again, adding the initial penalty condition makes little improvement to this number. However, the proportions are smaller in general in this case; only $88\%$ of false queries are identified by the initial conditions of rank pruning in this case. Thus, these initial conditions would be less accurate at predicting dominance query outcomes. If we used the initial conditions of rank pruning as a predictor here, then $(0.75333-Z_P)/(1-Z_P)\times 100= 27.914\%$ of queries predicted to be true would in fact be incomparable cases (false queries).\\

From the above proportions, we can see that including rank pruning in a dominance testing function allows a large proportion of dominance queries to be answered immediately (by initial conditions). The number answered immediately by rank pruning initial conditions is greater than that answered by the initial conditions of penalty pruning, showing rank pruning initial conditions to be stronger. Further, the number of queries answered by the combination of the two is only slightly more than those answered by rank pruning initial conditions alone. This suggests that there are few queries answered by the initial penalty condition that are not already answered by the initial rank condition. We have also seen that rank pruning initial conditions identify the majority of false queries. Thus, at least in the binary case, the initial condition of rank pruning, which is quick to check, could be used as a reasonably accurate predictor of dominance query outcomes.

\bibliography{ref}

\end{document}